%% file: arxiv_ver.tex
\documentclass[final]{cvpr}

\usepackage{times}
\usepackage{epsfig}
\usepackage{xcolor}
\usepackage{graphicx}
\usepackage{float}
\usepackage{amsmath}
\usepackage{amssymb,bm}
\usepackage[utf8]{inputenc}

\usepackage{caption}
\usepackage{subcaption}
\usepackage{array}

\usepackage[pagebackref=true,breaklinks=true,colorlinks,bookmarks=false]{hyperref}
\usepackage{cleveref}

\usepackage[numbers,sort,compress]{natbib}

\def\cvprPaperID{2691} 
\def\confYear{CVPR 2021}

\DeclareMathOperator*{\argmax}{arg\,max}
\DeclareMathOperator*{\argmin}{arg\,min}

\newcommand{\boldparagraph}[1]{\vspace{0.2cm}\noindent{\bf #1:}}
\def\wrt{wrt\onedot}

\begin{document}

\title{Locally Aware Piecewise Transformation Fields\\ for 3D Human Mesh Registration}

\author{
  Shaofei Wang$^{1}$,\; Andreas Geiger$^{2,3}$,\; Siyu Tang$^{1}$\\
  $^1$ETH Z\"{u}rich \quad
  $^2$Max Planck Institute for Intelligent Systems, T\"{u}bingen \quad
  $^3$University of T\"{u}bingen
}


\twocolumn[{%
\renewcommand\twocolumn[1][]{#1}%
\maketitle
\centering
\begin{minipage}{1\linewidth}
\includegraphics[trim=1.4cm 9.5cm 2.35cm 10cm, width=1\linewidth]{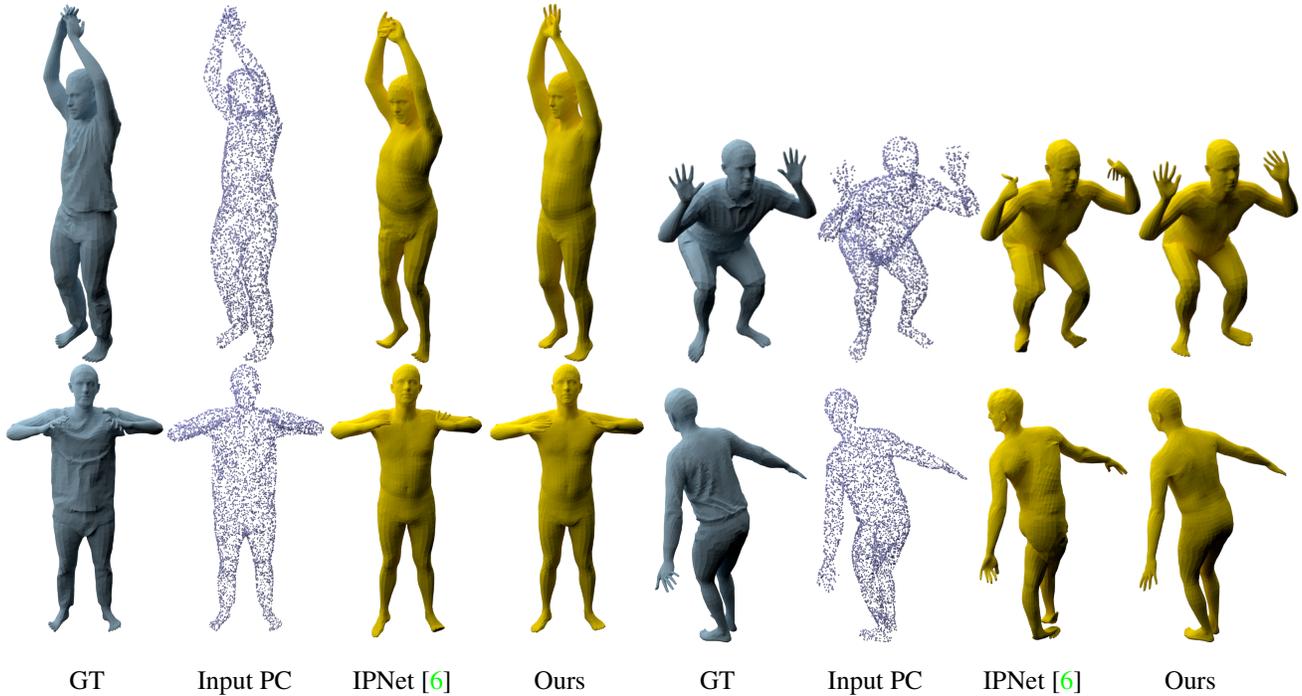}
\centering
\begin{tabular}{>{\centering\arraybackslash}p{\dimexpr 0.12\linewidth-2\tabcolsep} >{\centering\arraybackslash}p{\dimexpr 0.12\linewidth-2\tabcolsep} >{\centering\arraybackslash}p{\dimexpr 0.12\linewidth-2\tabcolsep} >{\centering\arraybackslash}p{\dimexpr 0.12\linewidth-2\tabcolsep} >{\centering\arraybackslash}p{\dimexpr 0.12\linewidth-2\tabcolsep} >{\centering\arraybackslash}p{\dimexpr 0.12\linewidth-2\tabcolsep} >{\centering\arraybackslash}p{\dimexpr 0.12\linewidth-2\tabcolsep} >{\centering\arraybackslash}p{\dimexpr 0.12\linewidth-2\tabcolsep}}
GT & Input PC & IPNet~\cite{Bhatnagar_ECCV2020} & Ours & GT & Input PC & IPNet~\cite{Bhatnagar_ECCV2020} & Ours
\end{tabular}
\captionof{figure}{Our proposed approach aims at providing accurate pose initialization which facilitates the subsequent optimization-based registration. We show here ground-truth clothed meshes \cite{CAPE:CVPR:20}, input point clouds, under-cloth SMPL meshes registered by IPNet~\cite{Bhatnagar_ECCV2020}, and under-cloth SMPL meshes registered with our model.\label{fig:teaser}}
\vspace{1cm}
\end{minipage}
}]

\begin{abstract}
Registering point clouds of dressed humans to parametric human models is a challenging task in computer vision. Traditional approaches often rely on heavily engineered pipelines that require accurate manual initialization of human poses and tedious post-processing. More recently, learning-based methods are proposed in hope to automate this process. We observe that pose initialization is key to accurate registration but existing methods often fail to provide accurate pose initialization. One major obstacle is that, despite recent effort on rotation representation learning in neural networks, regressing joint rotations from point clouds or images of humans is still very challenging. To this end, we propose novel piecewise transformation fields (PTF), a set of functions that learn 3D translation vectors to map any query point in posed space to its correspond position in rest-pose space. We combine PTF with multi-class occupancy networks, obtaining a novel learning-based framework that learns to simultaneously predict shape and per-point correspondences between the posed space and the canonical space for clothed human. Our key insight is that the translation vector for each query point can be effectively estimated using the point-aligned local features; consequently, rigid per bone transformations and joint rotations can be obtained efficiently via a least-square fitting given the estimated point correspondences, circumventing the challenging task of directly regressing joint rotations from neural networks. Furthermore, the proposed PTF facilitate canonicalized occupancy estimation, which greatly improves generalization capability and results in more accurate surface reconstruction with only half of the parameters compared with the state-of-the-art. Both qualitative and quantitative studies show that fitting parametric models with poses initialized by our network results in much better registration quality, especially for extreme poses.
\end{abstract}

\input{intro}

\input{related}
\input{fundamentals}
\input{PTF}
\input{experiments}

\section{Conclusion}
\label{sec:conclude}
In this paper, we introduce Piecewise Transformation Fields (PTFs), a set of piecewise transformation fields that learns to canonicalize query points. Combining PTFs and occupancy networks results in more parameter-efficient models that not only improve point cloud reconstruction quality, but also enable us to estimate joint rotations accurately through efficient optimization. With better reconstructed surfaces and accurately initialized poses, we achieve state-of-the-art results on automatic point cloud registration of dressed people. In the future, we plan to extend the proposed PTFs to the image domain, as well as combine it with other neural-representations for human bodies~\cite{SCALE:CVPR:21,SCANimate:CVPR:21,LEAP:CVPR:21} that require poses as inputs.

\boldparagraph{Acknowledgement} Andreas Geiger was supported by the ERC Starting Grant LEGO-3D (850533) and the DFG EXC number 2064/1 - project number 390727645. We thank Qianli Ma for helping us setting up the CAPE dataset. 

{\small
\bibliographystyle{ieee_fullname}
\bibliography{egbib}
}

\appendix
\onecolumn
\numberwithin{equation}{section}
\setcounter{equation}{0}
\numberwithin{figure}{section}
\setcounter{figure}{0}
\numberwithin{table}{section}
\setcounter{table}{0}

\input{arxiv_diff_nasa_ours}
\input{arxiv_model_archs}
\input{arxiv_quantitative_cape}
\input{arxiv_qualitative_cape}
\input{arxiv_qualitative_buff}
\input{arxiv_limitations}

\end{document}

%% file: intro.tex
\section{Introduction}
\label{sec:intro}
Human pose and shape registration from sensor inputs is a long-standing problem in computer vision. While unstructured 3D point clouds are becoming increasingly available, accurately registering these point clouds to parametric human shape models~\cite{SMPL:2015,SMPL-X:2019,Xu_2020_CVPR,Joo_2018_CVPR,SCAPE,Hasler2009CGF} still remains challenging, especially when considering clothed human with arbitrary poses. Traditional approaches often assume access to either temporal sequences of dense scans~\cite{dfaust:CVPR:2017,BUFF,Pons-Moll:Siggraph2017} or manually enforced constraints~\cite{Allen:SIGGRAPH:03,pishchulin17pr,SCAPE,Hasler2009CGF,Bogo:CVPR:2014,dfaust:CVPR:2017,Dyna:SIGGRAPH:2015}. These requirements limit the applicability of traditional approaches to static, arbitrarily obtained point clouds.

With the advent of neural implicit functions~\cite{Occupancy_Networks,DeepSDF,Chen_Implicit_2019_CVPR,Michalkiewicz_Implicit_2019_ICCV}, learning-based methods that reconstruct human shapes from point clouds are becoming increasingly accurate~\cite{IFNet}. However, most existing neural implicit models treat reconstructed human shapes as static objects and do not provide a way to register such reconstructions to parametric body models. Note that in the context of this paper, we refer to reconstruction as implicit surface reconstruction from point clouds, while registration refers to finding the shape and poses of a parametric model which best explains the input point cloud or the reconstructed surface. Some works distinguish registration (registering a point cloud to a template)~\cite{Bogo:CVPR:2014,groueix2018b,Allen:SIGGRAPH:03,SCAPE,Hasler2009CGF,Wei:CVPR:2016} and model-fitting (estimating parameters of a parametric model)~\cite{Bhatnagar_ECCV2020,bhatnagar2020loopreg,Zuffi:CVPR:2015}. In this context, our approach falls into the model-fitting category. However the definitions of registration and model-fitting are not consistent across papers, and for the ease of understanding we follow the convention of our major baseline~\cite{Bhatnagar_ECCV2020} and use registration and model-fitting interchangeably.

More recently, IPNet~\cite{Bhatnagar_ECCV2020} has been proposed for automatic point cloud registration of clothed humans. It predicts two sets of implicit surfaces, one for the clothed human and one for the human body under-cloth. 
IPNet then exploits optimization-based registration to fit a parametric model to the implicit surfaces from a fixed initial pose for all the subjects, with additional information about body part labels for each point in space. 
However, such semantic information is still very coarse and registration may fail when the target pose deviates too much from the initial pose (Fig.~\ref{fig:teaser}). 

We observe that, although the local point cloud features~\cite{IFNet} lead to reliable surface reconstruction, which facilitates the registration of parametric human models (e.g.~SMPL~\cite{SMPL:2015}), an accurate pose initialization is still key to reliable registration, as the underlying optimization problem is non-convex, thus reliable estimates are only obtained when initialized close to the solution.
However, it is difficult to estimate parametric poses from point clouds directly, because regressing pose parameters from neural networks is hard and unintuitive (as investigated in~\cite{Bhatnagar_ECCV2020}, as well as demonstrated in our experiments in Sec.~\ref{sec:eval_rot}). 

In this paper, we introduce a novel approach to estimating the SMPL~\cite{SMPL:2015} pose parameters, based on implicit representations and local point cloud features. Instead of regressing pose parameters directly from global features, we introduce a set of transformation functions, which take a query point and a local point cloud feature as input, and transform the query point to the rest-pose space (Fig.~\ref{fig:overview}). We assign one such transformation function per body-part, and name these functions \textit{Piecewise Transformation Fields} (PTF). The general idea of PTF is inspired by the observation that the rigid bone-transformations in SMPL can be calculated if we know point correspondences between the posed space and the unposed space. Another motivation for PTF is that by transforming query points into a canonical space before the occupancy query, we make the occupancy learning/inference task easier~\cite{Deng_ECCV2020,Huang_CVPR2020}. With PTF, our novel occupancy networks can estimate for each point: 1) its double-layer occupancy value (\ie\ inside body, in-between body and cloth, outside cloth) like IPNet does, 2) which body-part it belongs to, and 3) its corresponding position in rest-pose space. With this information, we can then extract a mesh-surface of human body, along with the semantic part label and the corresponding position in the rest-pose for each surface vertex. As a side-benefit, rigid transformations for each body-part can be estimated directly via least-square fitting. In terms of registering implicit surfaces to parametric models, our approach also employs optimization-based registration as IPNet. But unlike IPNet, our approach exploits point correspondences which allow us to more accurately initialize the pose parameters. This makes the registration process much more stable and accurate, especially for extreme poses; this will be evidenced in Section~\ref{sec:exp}. 

In summary, the contribution of this work is three-fold: (1) We propose Piecewise Transformation Fields (PTF) that learn to transform arbitrary points from posed space to rest-pose space. (2) We combine our PTF modules with occupancy networks, and achieve state-of-the-art results on clothed human reconstruction from point clouds on the CAPE dataset~\cite{CAPE:CVPR:20} while reducing the number of parameters by half. (3) We propose an alternative learning-based method for estimating joint rotations of the parametric SMPL model from point clouds. Our learning-based method takes advantage of local point-aligned features and produces more accurate and robust estimations than direct regression from global features. When fitting parametric models to implicit reconstructions using our estimated poses as initialization, we achieve $18\%$ reduction in registration error on average. Code is available at \href{https://taconite.github.io/PTF/website/PTF.html}{https://taconite.github.io/PTF/website/PTF.html}.

The remainder of this paper is structured as follows: in Section~\ref{sec:related_works} we give an overview of existing works that are related to our approach. In Section~\ref{sec:fundamentals} we review the fundamentals, SMPL~\cite{SMPL:2015} and NASA~\cite{Deng_ECCV2020}, upon which we build our proposed approach. In Section~\ref{sec:ptfs} we introduce our proposed PTF and its application to fast joint rotation estimation. In Section~\ref{sec:exp} we benchmark our PTF model, showing its advantages for registration and reconstruction. In Section~\ref{sec:conclude} we conclude and discuss possible future works.

%% file: related.tex
\section{Related Works}
\label{sec:related_works}
\boldparagraph{Optimization-based Registration} Optimization-based methods are most commonly used in traditional registration pipelines~\cite{pishchulin17pr,SCAPE,Allen:SIGGRAPH:03,Hasler2009CGF,Pons-Moll:Siggraph2017}. Traditional approaches rely on use of physical markers~\cite{Allen:SIGGRAPH:03,pishchulin17pr}, manually selected landmark-correspondences~\cite{SCAPE,Hasler2009CGF} or colored patterns on skin~\cite{Bogo:CVPR:2014,dfaust:CVPR:2017,Dyna:SIGGRAPH:2015}, and are thus not practical for automatic registration of arbitrary point clouds. Most recently proposed automatic registration algorithms aim at fitting a parametric model to sparse keypoints detected from images~\cite{SMPLify,SMPL-X:2019,song2020lgd}, they usually require the design of a complicated non-linear, non-convex loss function, optimize this loss \wrt the body parameters (body shapes, rotations, global translation, etc.), and are often prone to local optima. There are also approaches that conduct optimization-based registration using dynamic 3D inputs, \eg~\cite{dfaust:CVPR:2017,BUFF,Pons-Moll:Siggraph2017}, to name a few. Since we focus on static point cloud registration here, a complete review on this aspect is beyond the scope of this paper.

\boldparagraph{Learning-based Registration} Learning-based methods have been proposed to enable automatic registration of clothed humans from sparse point clouds~\cite{Bhatnagar_ECCV2020} or images~\cite{lazova2019360,alldieck2019learning,tex2shape_ICCV_2019,BodyNet:ECCV:2018}. Earlier works focus on one-shot estimation of parametric models from images~\cite{Dibra_2017_CVPR,Rong_2019_ICCV, Popa_2017_CVPR,Lassner_CVPR2017,Kanazawa_CVPR2018,Xiang_CVPR2019,Xu_ICCV2019,Kolotouros_ICCV2019, Kocabas_CVPR2020,kolotouros2019cmr,NBF:3DV:2018} using feed-forward neural networks. However, estimates from these methods are often noisy and most of them do not handle clothing. Recently, learning-based methods have been used to initialize important variables for optimization-based registration of clothed humans from point clouds~\cite{Bhatnagar_ECCV2020} or images~\cite{alldieck2019learning,tex2shape_ICCV_2019,BodyNet:ECCV:2018}. Our approach also uses a learning-based method to initialize optimization-based registration from sparse point clouds. Compared to the most recent learning-based registration approach for point clouds~\cite{Bhatnagar_ECCV2020}, our approach differs critically in that we extract additional point correspondence information, which produces more accurate pose initialization and thus improves subsequent optimization as evidenced by our experiments. Note that our approach directly estimates pose/joint-rotations as initialization, which is orthogonal to landmark-constrained optimization presented in~\cite{BodyNet:ECCV:2018,SCAPE,Hasler2009CGF}. In a concurrent work~\cite{bhatnagar2020loopreg}, the authors also propose to conduct $\mathbb{R}^3$ to $\mathbb{R}^3$ correspondence prediction using neural networks. Our approach differs from theirs in several aspects: (1)~\cite{bhatnagar2020loopreg} predicts correspondence from posed-space to unposed-unshaped-space and leaves pose estimation entirely to non-linear optimization, while our approach predicts correspondence from posed-space to unposed-shaped-space, enabling us to directly estimate joint rotations via fast linear optimization. (2)~\cite{bhatnagar2020loopreg} does not handle surface reconstruction. (3)~\cite{bhatnagar2020loopreg} devises a self-supervised training scheme, which, with moderate modifications, should be applicable to our framework as well, however we will leave such improvements for future studies.

\boldparagraph{Implicit Representations} With the introduction of neural implicit functions~\cite{Occupancy_Networks,Chen_Implicit_2019_CVPR,DeepSDF,Michalkiewicz_Implicit_2019_ICCV}, a number of implicit function-based methods have been proposed to reconstruct human bodies from either images~\cite{Saito_ICCV2019,Saito_CVPR2020,Huang_CVPR2020} or point clouds~\cite{IFNet}. However, all aforementioned methods treat human body as a rigid object and do not handle registration.~\cite{occupancy_flow, jiang2020shapeflow} model the human body as a non-rigid, deformable occupancy field and are able to transform human shape representations across poses via neural ordinary differential equations (ODEs)~\cite{neuralODEs}. Our approach is partially inspired by~\cite{occupancy_flow} as both our approach and~\cite{occupancy_flow} learn a $\mathbb{R}^3$ to $\mathbb{R}^3$ transformation. However, our approach aims to transform points from posed space to rest-pose space and uses a traditional neural network to do that, while~\cite{occupancy_flow,jiang2020shapeflow} learn transformations from one posed-space to another posed-space. Furthermore, our approach utilizes local point cloud features to learn the transformation while~\cite{occupancy_flow,jiang2020shapeflow} are based on global features. 

\boldparagraph{Correspondence Prediction for Model Fitting} Dense correspondence prediction has been employed for human pose estimation~\cite{vitruvian_manifold,Pons-Moll_BMVC_2013,Pons-Moll_IJCV-2015}, hand pose tracking~\cite{taylor2016efficient}, non-rigid object alignment~\cite{zollhoefer2014deformable,Bernard_2020_CVPR} and undressed human mesh registration~\cite{groueix2018b}.~\cite{vitruvian_manifold,Pons-Moll_BMVC_2013,Pons-Moll_IJCV-2015} regresses 3D correspondences from the input to a single canonical model after which they infer the human pose via non-linear optimization.~\cite{zollhoefer2014deformable} focuses on non-rigid object reconstruction and tracking and is not able to produce controllable models.~\cite{Bernard_2020_CVPR} optimizes over a sparse set of points for general non-rigid object alignment but does not handle model-fitting of parametric models.~\cite{groueix2018b} focuses on the registration of undressed people and does not handle dressed people registration or under-cloth body registration. Contrary to all these works, our approach addresses registrations of dressed humans and their under-cloth bodies, as well as SMPL parameter estimation simultaneously using a single neural-network.

%% file: fundamentals.tex
\section{Fundamentals}
\label{sec:fundamentals}
We start by briefly reviewing the fundamentals upon which we derive our approach: the classical SMPL model~\cite{SMPL:2015} and a recently proposed neural implicit approximation of the SMPL model, NASA~\cite{Deng_ECCV2020}.

\subsection{SMPL Body Model}
\label{sec:SMPL}
SMPL is a classical parametric human body model. In SMPL, shapes are controlled by a low-dimensional vector $\bm{\beta} \in \mathbb{R}^{10}$ which weights a set of different vertex-offsets that correspond to the principal components of PCA on training scans; poses for 24 joints (23 body-joints plus pelvis as root) are denoted as $\bm{\theta} \in \mathbb{R}^{72}$, where the pose for each joint is defined as the rotation (in axis-angle format) relative to its parent in the kinematic tree. To build a posed and shaped mesh of the human body, SMPL starts with an artist created mean-shape template $\bar{\mathbf{T}} \in \mathbb{R}^{3 \times 6890}$ with blend skinning weights $\mathcal{W} \in \mathbb{R}^{6890 \times 24}$. SMPL first applies a blend-shape function $B_S(\bm{\beta}): \mathbb{R}^{|\bm{\beta}|} \mapsto \mathbb{R}^{3 \times 6890}$ which maps $\bm{\beta}$ to a set of additive per-vertex displacements that describes the person's body shape (shape-blend-shapes). Then, SMPL applies a pose-dependent blend shape function $B_P(\bm{\theta}): \mathbb{R}^{|\bm{\theta}|} \mapsto \mathbb{R}^{3 \times 6890}$ which maps $\bm{\theta}$ to another set of additive per-vertex displacements that accounts for dynamic soft-tissue deformations caused by the pose deviation from the rest-pose (pose-blend-shapes). Adding the aforementioned displacements to the mean template results in a rest-pose mesh $\mathbf{T}_P(\bm{\beta}, \bm{\theta}) = \bar{\mathbf{T}} + B_S(\bm{\beta}) + B_P(\bm{\theta})$, which includes the body shape of a person as well as dynamic soft-tissue deformations. Simultaneously, SMPL also regresses joint locations at the rest-pose from shape parameters via a linear joint regressor $J(\bm{\beta}): \mathbb{R}^{|\bm{\beta}|} \mapsto \mathbb{R}^{72}$. A set of $4 \times 4$ rigid bone transformations $\{ \mathbf{B}_b \} = \{ \mathbf{B}_1, \cdots, \mathbf{B}_{24} \}$ can be constructed from rest-pose joint locations and relative joint rotations $\bm{\theta}$. SMPL applies a standard linear-blend-skinning (LBS) function $W(\cdot)$ with bone transformations $\{ \mathbf{B}_b \}$ and skinning weights $\mathcal{W}$, transforming $\mathbf{T}_P(\bm{\beta}, \bm{\theta})$ into the posed and shaped mesh. Finally, global translation $\mathbf{t} \in \mathbb{R}^3$ is added to the posed and shaped mesh. In summary,
SMPL models a function $M(\bm{\beta}, \bm{\theta}, \mathbf{t})$ that takes $\bm{\beta}$, $\bm{\theta}$ and $\mathbf{t}$ as inputs and outputs a human mesh:
\begin{align}
\label{eqn:SMPL_func}
    M(\bm{\beta}, \bm{\theta}, \mathbf{t}) = W(\bar{\mathbf{T}} + B_S(\bm{\beta}) + B_P(\bm{\theta}); \mathcal{W}, \{ \mathbf{B}_b \}) + \mathbf{t}
\end{align}
We denote the $3 \times 3$ rotation matrix representation of joint $b$'s rotation as $\mathbf{R}_b (\bm{\theta})$, with $\mathbf{R}_b (\mathbf{0}) = \mathbf{I}, \forall b \in \{1, \cdots, 24 \}$. Let $\mathbf{t}_b \in \mathbb{R}^3$ be the location of rest-pose joint $b$ obtained via $\mathcal{J} (\bm{\beta})$. Let $\mathbf{A}(b)$ denote the ordered set of $b$ and its ancestors on the kinematic tree. Let $\bar{b}$ denote the parent of joint $b$. The bone-transformations $\{ \mathbf{B}_b \}$ are defined as:
\begin{align}
    \mathbf{B}_b &= G_b (\bm{\beta}, \bm{\theta}) G_b (\bm{\beta}, \mathbf{0})^{-1} \label{eqn:rel_bone_transformations} \\  
    G_b (\bm{\beta}, \bm{\theta}) &= \prod_{b^{\prime} \in \mathbf{A}(b)} \begin{bmatrix}   \mathbf{R}_{b^{\prime}} (\bm{\theta}) & \mathbf{t}_{b^{\prime}} - \mathbf{t}_{\bar{b^{\prime}}} \\ \mathbf{0} & 1 \end{bmatrix}\label{eqn:abs_bone_transformations} \\
    \{ \mathbf{t}_b \} &= \mathcal{J}(\bm{\beta}) \label{eqn:joint_locations}
\end{align}
Note that adding the global translation $\mathbf{t}$ to the translation parts of $\{ \mathbf{B}_b \}$ does not change the results of $M(\bm{\beta}, \bm{\theta}, \mathbf{t})$. Thus in the remainder of the paper we re-define $\{ \mathbf{B}_b \}$ by adding $\mathbf{t}$ to its translation parts.

\paragraph{Registration:} The SMPL model in Eq.~\eqref{eqn:SMPL_func} is end-to-end differentiable \wrt to $\bm{\beta}, \bm{\theta}, \mathbf{t}$, thus given a point cloud, it is straightforward to construct an energy function on the distances between the point cloud and the SMPL mesh and minimize it \wrt $\bm{\beta}, \bm{\theta}, \mathbf{t}$. However this optimization is non-convex and non-linear and thus prone to local minima. Therefore, manual initialization of pose and manual post-processing to fix erroneous fits are usually required.

IPNet~\cite{Bhatnagar_ECCV2020} proposes to reconstruct an inner surface (for under-cloth body) and an outer surface from point cloud of dressed humans. For registration, it first fits a SMPL mesh $M(\bm{\beta}, \bm{\theta}, \mathbf{t})$ to the inner surface, then it fits a SMPL+D~\cite{alldieck2019learning,lazova2019360} mesh $M^{\prime}(\bm{\beta}, \bm{\theta}, \mathbf{t}, \mathbf{D})$ to the outer surface, where $\mathbf{D} $ represents per-vertex displacements that capture clothing deformations in rest-pose space. Formally,~\cite{Bhatnagar_ECCV2020} aims at minimizing the following energy:
\begin{align}
\label{eqn:SMPL_optim}
E(\bm{\theta}, \bm{\beta}, \mathbf{t}, \mathbf{D}) = w_{\text{data}}E_{\text{data}} + w_{\text{part}}E_{\text{part}} + w_{\text{reg}}E_{\text{reg}}
\end{align}
in which $E_{\text{data}}$ measures the distances between reconstructed surfaces and the SMPL/SMPL+D meshes. $E_{\text{part}}$ enforce part correspondences. $E_{\text{reg}}$ is the regularization term, which includes a Laplacian regularization term for clothed surface, a pose prior from~\cite{Pons-Moll:Siggraph2017} and a shape prior that encourages $\bm{\beta}$ to be small. The $E_{\text{part}}$ term makes the optimization more stable and thus~\cite{Bhatnagar_ECCV2020} leads to improved registrations. However, the part correspondences are very coarse and the registration may fail for extreme poses as illustrated in Fig.~\ref{fig:teaser}.

In the next subsection we will review NASA~\cite{Deng_ECCV2020}, a recent work that relates neural implicit functions to Eq.~\eqref{eqn:SMPL_func}.

\subsection{Neural Articulated Shape Approximation}
\label{sec:nasa}
NASA~\cite{Deng_ECCV2020} proposes to learn an implicit version of Eq~\eqref{eqn:SMPL_func}. Towards this goal, NASA learns a set of occupancy functions parameterized by $\omega$, $\{ \bar{\mathcal{O}}^b_{\omega} \} = \{ \bar{\mathcal{O}}^1_{\omega}, \cdots, \bar{\mathcal{O}}^{24}_{\omega} \}$, producing one occupancy value for each bone, given a 3D query point $\mathbf{x} \in \mathbb{R}^3$ and $\bm{\beta}$, $\bm{\theta}, \mathbf{t}$\footnote{With some abuse of notation, we also use $\mathbf{x}$ to denote query points in homogeneous representation in $\mathbb{R}^4$ when applicable.}. The key idea of NASA is to learn the occupancy functions in the local-coordinate space of each bone with shape-blend-shapes and pose-blend-shapes. However these local-coordinates are piecewise \textit{discontinuous}. We extend NASA's idea by learning occupancy functions in the \textit{continuous} rest-pose space as this facilitates the correspondence learning that will be presented in Section~\ref{sec:ptfs}. We refer readers to Appendix~\ref{appx:NASA_diff} for an elaboration on the difference between our modified NASA the original NASA~\cite{Deng_ECCV2020}. For our modified NASA, $\{ \bar{\mathcal{O}}^b_{\omega} \}$ is an implicit approximation to the $\bar{\mathbf{T}} + B_S(\bm{\beta}) + B_P(\bm{\theta})$ term in Eq.~\eqref{eqn:SMPL_func}. The occupancy value of an arbitrary point $\mathbf{x}$ in posed-space can be queried as:
\begin{align}
\label{eqn:nasa}
\mathcal{O} (\mathbf{x} | \{ \mathbf{B}_b \}, \mathbf{t}_1) = \max_b \{ \bar{\mathcal{O}}^b_{\omega} \big(\mathbf{B}_b^{-1} \mathbf{x}, \bm{\Pi}^b_{\omega} [\{ \mathbf{B}_b^{-1} \mathbf{t}_1 \}] \big) \}
\end{align}
Where $\mathbf{t}_1 \in \mathbb{R}^4$ is the root-joint location in homogeneous representation. $\{ \mathbf{B}_b^{-1} \mathbf{t}_1 \} = \{ \mathbf{B}_1^{-1} \mathbf{t}_1, \cdots, \mathbf{B}_{24}^{-1} \mathbf{t}_1 \} \in \mathbb{R}^{72}$ is a compact representation of pose. $\bm{\Pi}^b_{\omega} \in \mathbb{R}^{72 \times 4}$ is a learned matrix that applies a sub-space linear projection on $\{ \mathbf{B}_b^{-1} \mathbf{t}_1 \}$, one per $b \in \{ 1, 2, \cdots, 24 \}$. Since \{$\mathbf{B}_b\}$ depends on $\bm{\beta}$, $\bm{\theta}$ and $\mathbf{t}$ (Eq.~\eqref{eqn:rel_bone_transformations}-\eqref{eqn:joint_locations}), the term $\bm{\Pi}^b_{\omega} [\{ \mathbf{B}_b^{-1} \mathbf{t}_1 \}]$ serves to provide shape-blend-shapes and pose-blend-shapes information to the occupancy functions $\{ \bar{\mathcal{O}}^b_{\omega} \}$.

Essentially, Eq.~\eqref{eqn:nasa} \textit{canonicalizes} the query point $\mathbf{x}$ \textit{before} doing the occupancy query. Thus the neural networks  $\{ \bar{\mathcal{O}}^b_{\omega} \}$ are learning occupancy in rest-pose, which reduces the difficulty for training occupancy functions compared to a posed representation. As we will show in Section~\ref{sec:exp}, although NASA does not handle reconstruction, the general argument that canonicalizing query points helps occupancy learning still holds for our reconstruction networks.

%% file: PTF.tex
\section{Piecewise Transformation Fields}
\label{sec:ptfs}
\begin{figure*}[t]
\centering
\includegraphics [trim=1cm 0.5cm 1cm 1cm, width=0.95\textwidth]{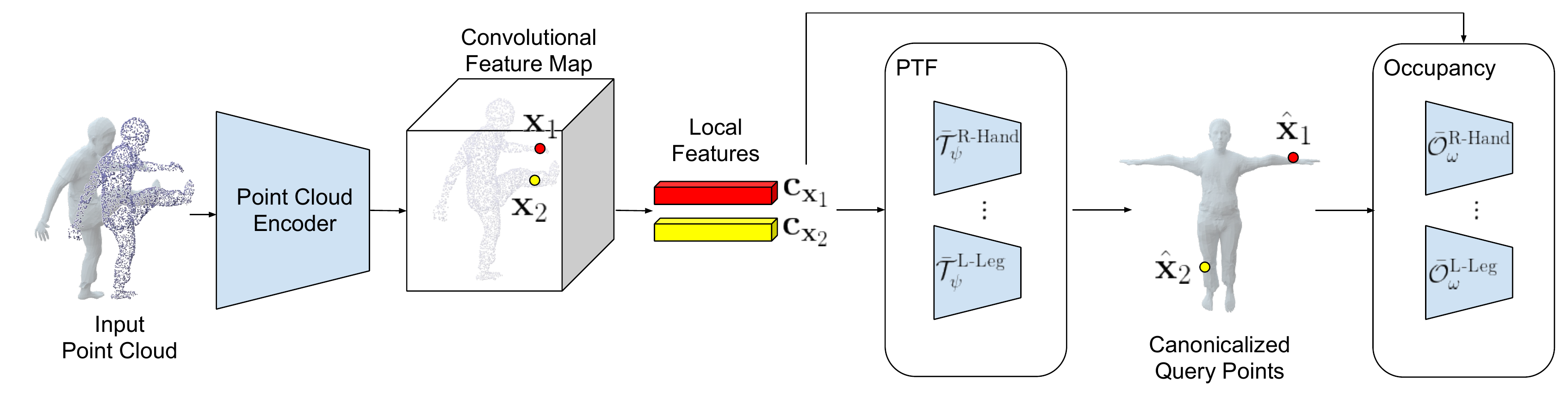}
\caption{Overview of the proposed method. The key idea is to utilize the learned PTF module to transform query points $\mathbf{x}_1, \mathbf{x}_2$ in posed space into corresponding points $\hat{\mathbf{x}}_1, \hat{\mathbf{x}}_2$ in unposed-space, and conduct occupancy classification in the unposed-space. This construction makes occupancy learning easier, while also providing dense $\mathbb{R}^3$ to $\mathbb{R}^3$ correspondences, which enables us to estimate joint-rotations via efficient linear optimization.}
\label{fig:overview}
\end{figure*}
It is important to remark that NASA requires ground-truth SMPL paramaters $\bm{\beta}$, $\bm{\theta}$, $\mathbf{t}$ as inputs and does not solve the sensor-input-to-shape inference problem. In other words, NASA is a neural approximation to the SMPL model. In contrast, in this work, we assume $\bm{\theta}$, $\bm{\beta}$, $\mathbf{t}$ and consequently $\{ \mathbf{B}_b \}$ unknown and propose to learn a set 
of 3D point transformations from input space to canonical space. We parameterize these transformation functions by $\psi$, and formally define them as $\{ \bar{\mathcal{T}}^b_{\psi} (\mathbf{x}, \mathbf{c}_{\mathbf{x}}) \}: (\mathbb{R}^3, \mathbb{R}^{|\mathbf{c}_{\mathbf{x}}|}) \rightarrow \mathbb{R}^{3 \times 24}$. These functions approximate the \textit{effect of bone transformations} on the query points instead of bone transformations themselves, \ie\ $\bar{\mathcal{T}}^b_{\psi} (\mathbf{x}, \mathbf{c}_{\mathbf{x}}) \approx \mathbf{B}_b^{-1} \mathbf{x}, \forall b \in \{ 1, 2, \cdots, 24 \}$, where $\mathbf{c}_{\mathbf{x}}$ is a point-aligned feature taken from a convolutional feature map~\cite{IFNet,ConvONet} at the location of $\mathbf{x}$. This is advantageous in that bone transformations are local for each bone, thus local features should be more suitable for predicting local bone transformations. This will be experimentally demonstrated in Section~\ref{sec:eval_rot}. Since we have one transformation field per bone, we formally name the set of transformation functions $\{ \bar{\mathcal{T}}^b_{\psi} (\cdot, \cdot) \}$ \textit{Piecewise Transformation Fields} (PTF). With our PTF, the occupancy functions are now only dependent on point cloud features $\mathbf{c}_{\mathbf{x}}$ and thus we can handle reconstruction from point cloud. Also, with the learned feature $\mathbf{c}_{\mathbf{x}}$, we found that our networks generalize well even without the projection matrix $\bm{\Pi}^b_{\omega}$, thus we discard $\bm{\Pi}^b_{\omega}$ in our models.

With PTF, we can rewrite Eq.~\eqref{eqn:nasa} as follows:
\begin{align}
\label{eqn:ptfs+nasa}
\mathcal{O} (\mathbf{x} | \mathbf{c}_{\mathbf{x}} ) = \max_b \{ \bar{\mathcal{O}}^b_{\omega} \big(\bar{\mathcal{T}}^b_{\psi} (\mathbf{x}, \mathbf{c}_{\mathbf{x}}), \mathbf{c}_{\mathbf{x}} \big) \}
\end{align}
Note that this model differs critically from IPNet~\cite{Bhatnagar_ECCV2020} in that our approach canonicalizes query points via PTF and performs occupancy queries in rest-pose space, while IPNet does occupancy queries in posed-space. We denote the model defined in Eq.~\eqref{eqn:ptfs+nasa} as PTF-Piecewise as it is fully piecewise, \ie\ the individual modules are independent of each other. We use fewer channels (80) per function as compared to IPNet (256), which, combined with the piecewise design, reduces the total number of parameters from 2.2M to 1.2M. As we will show in Section~\ref{sec:exp}, although using fewer parameters, our PTF-Piecewise model retains a comparable reconstruction accuracy. 

\boldparagraph{Loss functions} To define loss functions to Eq.~\eqref{eqn:ptfs+nasa}, we first define the outputs from Eq.~\eqref{eqn:ptfs+nasa}. We employ the multi-class occupancy outputs in IPNet which enable the occupancy networks to predict occupancy of dressed humans and their under-cloth bodies simultaneously, thus the set of outputs from $\{ \bar{\mathcal{O}}^b_{\omega} \}$ can be denoted as $\{ \hat{\mathbf{o}}_b \} \in [0, 1]^{3 \times 24}$.  We further denote the outputs from $\{ \bar{\mathcal{T}}^b_{\psi} \}$ as $\{ \hat{\mathbf{x}}_b \} \in \mathbb{R}^{3 \times 24}$. Thus for each query point $\mathbf{x}$ we will predict 24 sets of occupancy value and correspondence. To obtain the final occupancy value $\mathcal{O}(\mathbf{x} | \mathbf{c}_{\mathbf{x}}) = \hat{\mathbf{o}} \in [0, 1]^3$ and the correspondence $\mathcal{T}(\mathbf{x} | \mathbf{c}_{\mathbf{x}}) = \hat{\mathbf{x}} \in \mathbb{R}^3$, we also need to define a probability simplex $\hat{\mathbf{p}} \in [ 0, 1 ]^{24}, \text{s.t.} \sum_{b=1}^{24} \hat{\mathbf{p}}_b = 1$ which describes part probabilities for the query point $\mathbf{x}$. This $\hat{\mathbf{p}}$ can be predicted via $\hat{\mathbf{p}}_b = \frac{\max \bm{\sigma}_b}{\sum_{b^{\prime}} \max \bm{\sigma}_{b^{\prime}}}$ where $\bm{\sigma}_b \in \mathbb{R}^3$ is the pre-activation occupancy logits of $\hat{\mathbf{o}}_b$. Note that we drop the dependencies of $\{ \hat{\mathbf{o}}_b \}$, $\{ \bm{\sigma}_b \}$, $\hat{\mathbf{o}}$ and $\hat{\mathbf{p}}$ on $\mathbf{x}$ for notation simplicity.

With the aforementioned notations, we can rewrite the outputs of Eq.~\eqref{eqn:ptfs+nasa} as:
\begin{align}
\label{eqn:ptfs-occ-corr}
    \hat{\mathbf{o}} = \sum_{b=1}^{24} \hat{\mathbf{p}}_b \hat{\mathbf{o}}_b, \quad \hat{\mathbf{x}} = \sum_{b=1}^{24} \hat{\mathbf{p}}_b \hat{\mathbf{x}}_b
\end{align}
%
Similar to IPNet, we can also add an additional part classifier to predict $\hat{\mathbf{p}}$ directly from $\mathbf{c}_{\mathbf{x}}$. This results in slightly more parameters than PTF-Piecewise but also slightly more accurate reconstruction than PTF-Piecewise. To keep the memory usage roughly the same as IPNet, we reduces the number of channels in the occupancy classifier from 256 to 128, and use 128 channels as well for the PTF. This reduces the total number of parameters in the decoder from 2.2M to 1.4M. For a fair comparison, we follow IPNet and use 14 parts, \ie\ we use 14 piecewise occupancy functions for the occupancy classifier. But the part classifier outputs 24 parts probabilities. This is necessary because SMPL defines 24 parts and in order to compute SMPL-compatible joint rotations, the labels for query points has to be consistent with SMPL. We merge the 24 parts probabilities into 14 parts in order to make Eq.~\eqref{eqn:ptfs-occ-corr} possible. 
We denote this model as the PTF-FC model. Please refer to Appendix~\ref{appx:archs} for detailed network architectures of PTF-Piecewise and PTF-FC.


To train our PTF models, we define three loss terms: an occupancy loss $\mathcal{L}_{occ}$, a classification loss $\mathcal{L}_{cls}$ and a correspondence loss $\mathcal{L}_{corr}$. 
\begin{align}
    \mathcal{L}_{occ} &= \text{CE}(\hat{\mathbf{o}}, \mathbf{o}_{\mathbf{x}}) \\
    \mathcal{L}_{cls} &= \text{CE}(\hat{\mathbf{p}}, \mathbf{p}_{\mathbf{x}}) \\
    \mathcal{L}_{corr} &= \left\Vert \hat{\mathbf{x}} - W^{-1}(\mathbf{x}; \mathcal{W}, \{ \mathbf{B}_b \}) \right\Vert_2
\end{align}
where CE stands for multi-class cross-entropy loss. $\mathbf{o}_{\mathbf{x}}$ is the ground-truth occupancy value for query point $\mathbf{x}$, similarly $\mathbf{p}_{\mathbf{x}}$ is the one-hot vector of the ground-truth part-label. $W^{-1}(\mathbf{x}; \mathcal{W}, \{ \mathbf{B}_b \}): \mathbb{R}^3 \mapsto \mathbb{R}^3$ is the inverse LBS function that maps an arbitrary point from the posed-space to the rest-pose using ground-truth bone-transformations $\{ \mathbf{B}_b \}$ and skinning weights $\mathcal{W}$. To obtain the ground-truth labels $\mathbf{p}_{\mathbf{x}}$ and for calculating $W^{-1} (\mathbf{x}; \mathcal{W}, \{ \mathbf{B}_b \})$ we need to associate the query point $\mathbf{x}$ to one of the 6890 SMPL vertices; this is done by assigning $\mathbf{x}$ its the nearest vertex on the SMPL mesh. Finally, we weight $\mathcal{L}_{cls}$ and $\mathcal{L}_{corr}$ with $w_1$ and $w_2$, the total loss is then:
\begin{align}
    \mathcal{L} = \mathcal{L}_{occ} + w_1 \mathcal{L}_{cls} + w_2 \mathcal{L}_{corr}
\end{align}
we set $w_1=0.1$ and $w_2=1.0$ throughout our experiments.

\boldparagraph{Model Fitting with PTF} An advantage of our model is that it allows for efficient and accurate prediction of joint rotations. Let $\mathbf{X} = \{ \mathbf{x}^{(1)}, \mathbf{x}^{(2)}, \cdots, \mathbf{x}^{(D)} \}$ be a set of $D$ points that lie on a human-mesh surface, and let $\mathbf{l} \in \{ 1, 2, \cdots, 24 \}^D$ be the vector of semantic labels for these $D$ points. We estimate bone transformation $\mathbf{B}_b^*$ for $b$ by:
\begin{align}
\mathbf{B}_b^* &= \argmin_{\mathbf{B}_b} \sum_{i=1}^{D} [ \mathbf{l} (i) = b ] {\lVert \mathbf{x}^{(i)} - \mathbf{B}_b \bar{\mathcal{T}}^b_{\psi} (\mathbf{x}^{(i)}, \mathbf{c}_{\mathbf{x}^{(i)}} ) \rVert}_{2} \label{eqn:lsq_obj} \\
\mathbf{l} (i) &= \argmax_b \{ \bar{\mathcal{O}}^b_{\omega} \big(\bar{\mathcal{T}}^b_{\psi} (\mathbf{x}^{(i)}, \mathbf{c}_{\mathbf{x}^{(i)}} ), \mathbf{c}_{\mathbf{x}^{(i)}} \big) \} \label{eqn:part_labels}
\end{align}
where $[\cdot]$ is an indicator function. Eq.~\eqref{eqn:lsq_obj} can be solved efficiently for each part via least-square fitting~\cite{rigid_lsq}. Joint rotation estimations $\bm{\theta}^*$ can then be factorized from $\mathbf{B}_b^*$ via Eq.~\eqref{eqn:rel_bone_transformations}-\eqref{eqn:joint_locations}. To make the estimation more robust to outliers, in practice we use least-square fitting Eq.~\eqref{eqn:lsq_obj} within RANSAC~\cite{RANSAC_1981}. With the estimated $\bm{\theta}^*$ as initialization, we then fit SMPL/SMPL+D meshes to our reconstructed body and cloth surfaces according to Eq.~\eqref{eqn:SMPL_optim} using exactly the same weights and learning rate schedules as in IPNet~\cite{Bhatnagar_ECCV2020}.


%% file: experiments.tex
\section{Experiments}
\label{sec:exp}
Since IPNet~\cite{Bhatnagar_ECCV2020} is the only existing automatic point cloud registration work for clothed humans, we primarily compare our approach against IPNet in terms of both registration quality and reconstruction quality. Furthermore, we provide ablation studies on encoder choices and the necessity of using our PTF architecture. We also compare joint rotation estimation of our approach to a naive baseline that regresses rotation matrices from global point cloud features.

\boldparagraph{Datasets}
Because the official model of IPNet was trained on private data which is not publicly available, we instead use the CAPE dataset~\cite{CAPE:CVPR:20} to train and test IPNet and our models for a fair comparison. The CAPE dataset consists of 148584 pairs of registered clothed/minimally-clothed meshes of 15 subjects with different genders, the registered meshes all have SMPL topology and thus are compatible with SMPL. We use 12 subjects for training and 3 subjects for testing. We also subsample the recordings by a factor of 5. The final training set consists of 26004 frames while the validation set consists of 3965 frames. We generate input point clouds by sampling 5K points on the surfaces of the clohted meshes with Gaussian noise of zero mean and 1mm standard-deviation. The sampling strategy of query points is as follows: we sample 2K points uniformly in the point cloud's bounding box, 4K points near clothed-meshes' surface and 4K points near the minimally-clothed-meshes' surface.

\boldparagraph{Training Details}
We train all IPNet models and our models using the Adam optimizer with a learning rate of $1e-4$. The batch size is set to 12. We run 200K iterations for training, using a recently proposed convolutional point cloud encoder~\cite{ConvONet}. The training takes roughly two days on a single NVIDIA-V100 GPU.
\begin{figure*}[t]
\centering
 \begin{subfigure}[b]{0.15\textwidth}
    \includegraphics [trim=9cm 7cm 9cm 8.5cm, width=0.95\textwidth]{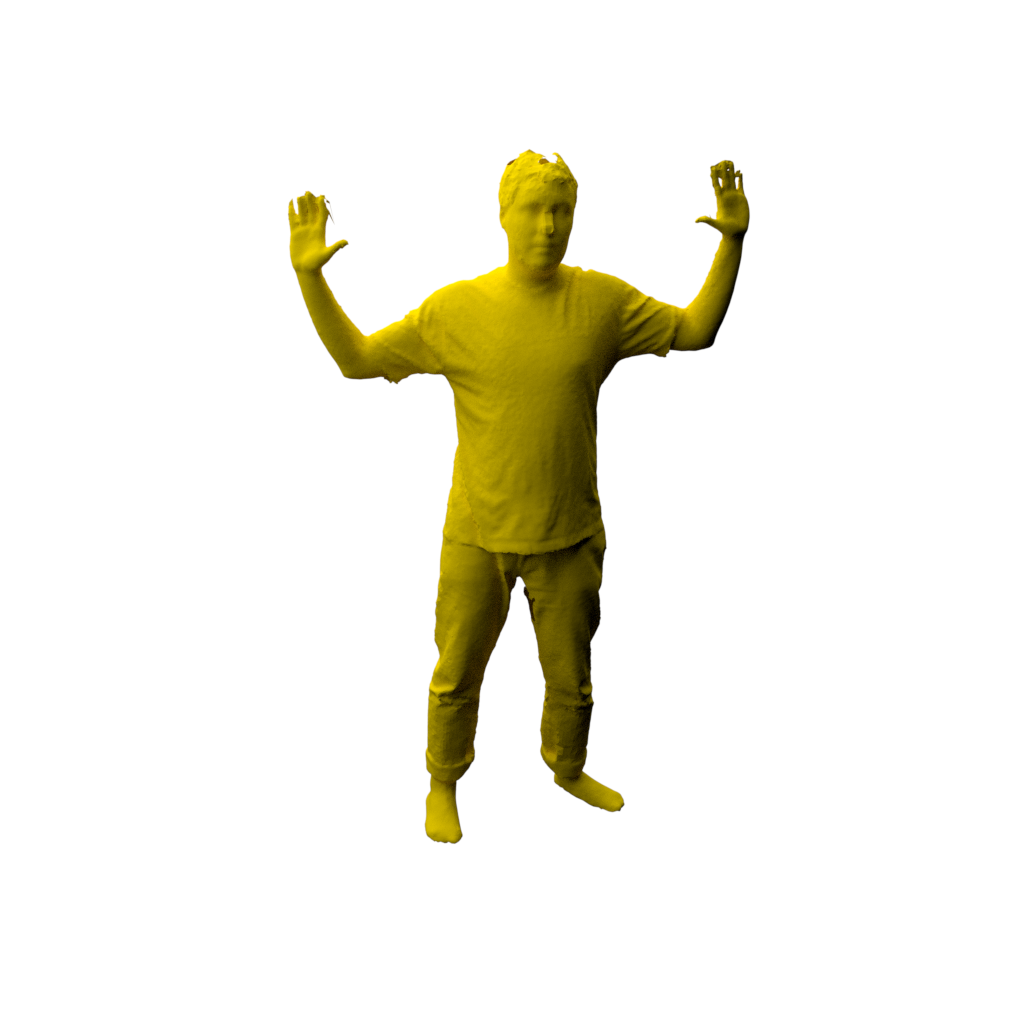}
\end{subfigure}
 \begin{subfigure}[b]{0.15\textwidth}
    \includegraphics [trim=9cm 7cm 9cm 8.5cm, width=0.95\textwidth]{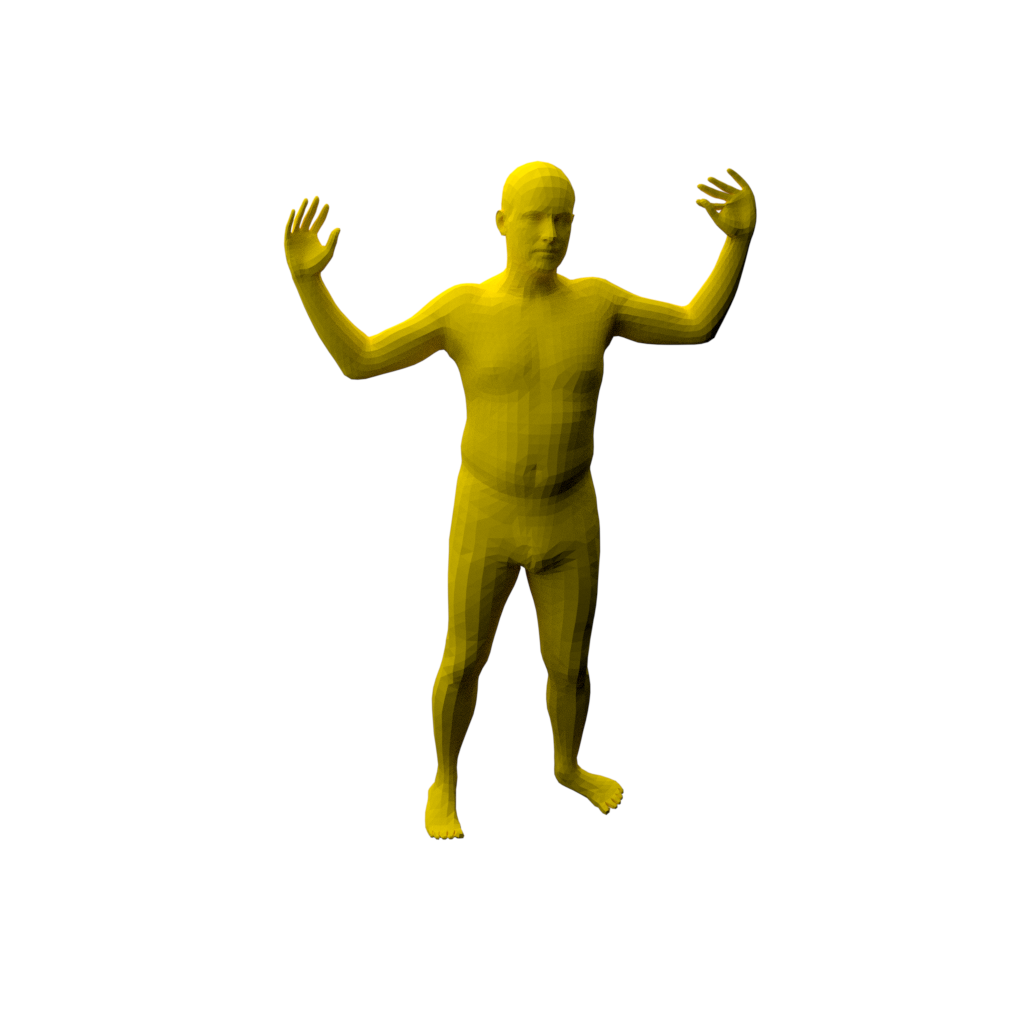}
 \end{subfigure}
 \begin{subfigure}[b]{0.15\textwidth}
    \includegraphics [trim=9cm 7cm 9cm 8.5cm, width=0.95\textwidth]{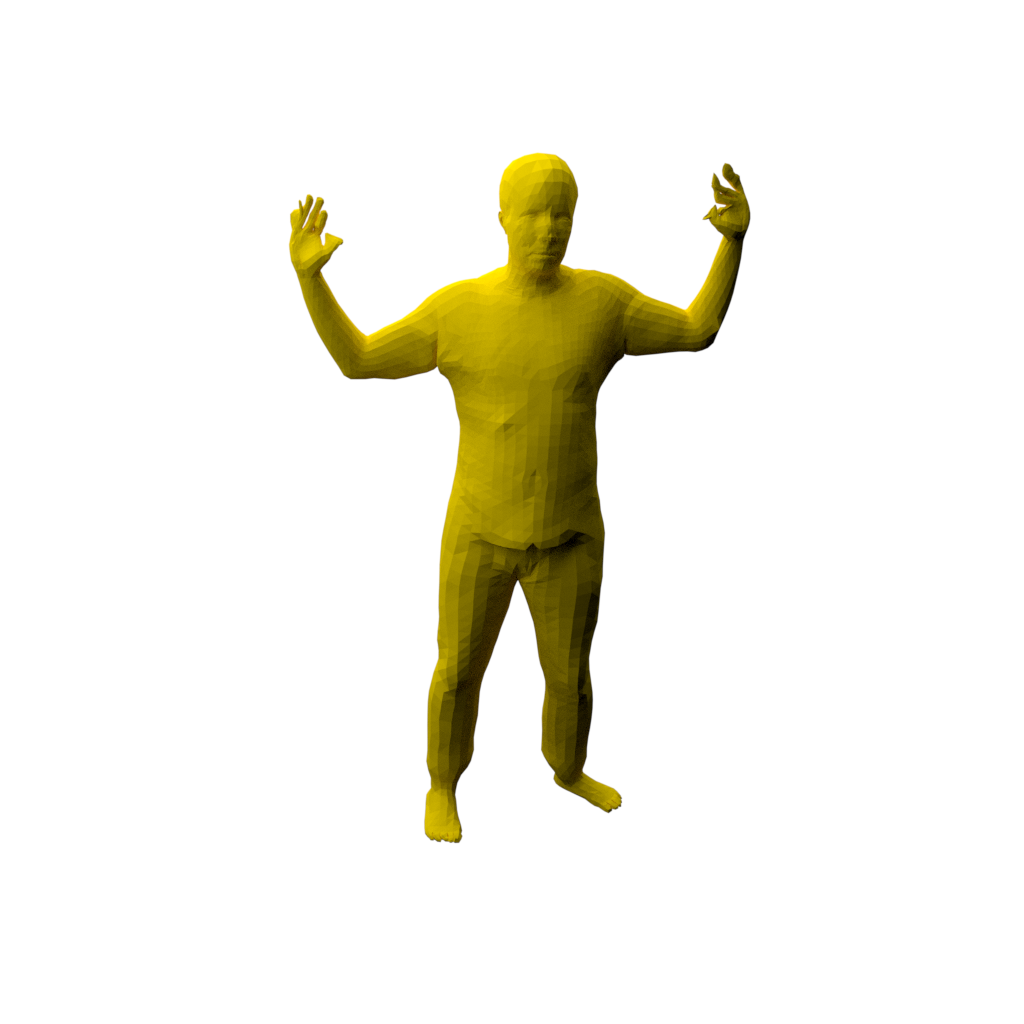}
 \end{subfigure}
 \begin{subfigure}[b]{0.15\textwidth}
    \includegraphics [trim=9cm 7cm 9cm 8.5cm, width=0.95\textwidth]{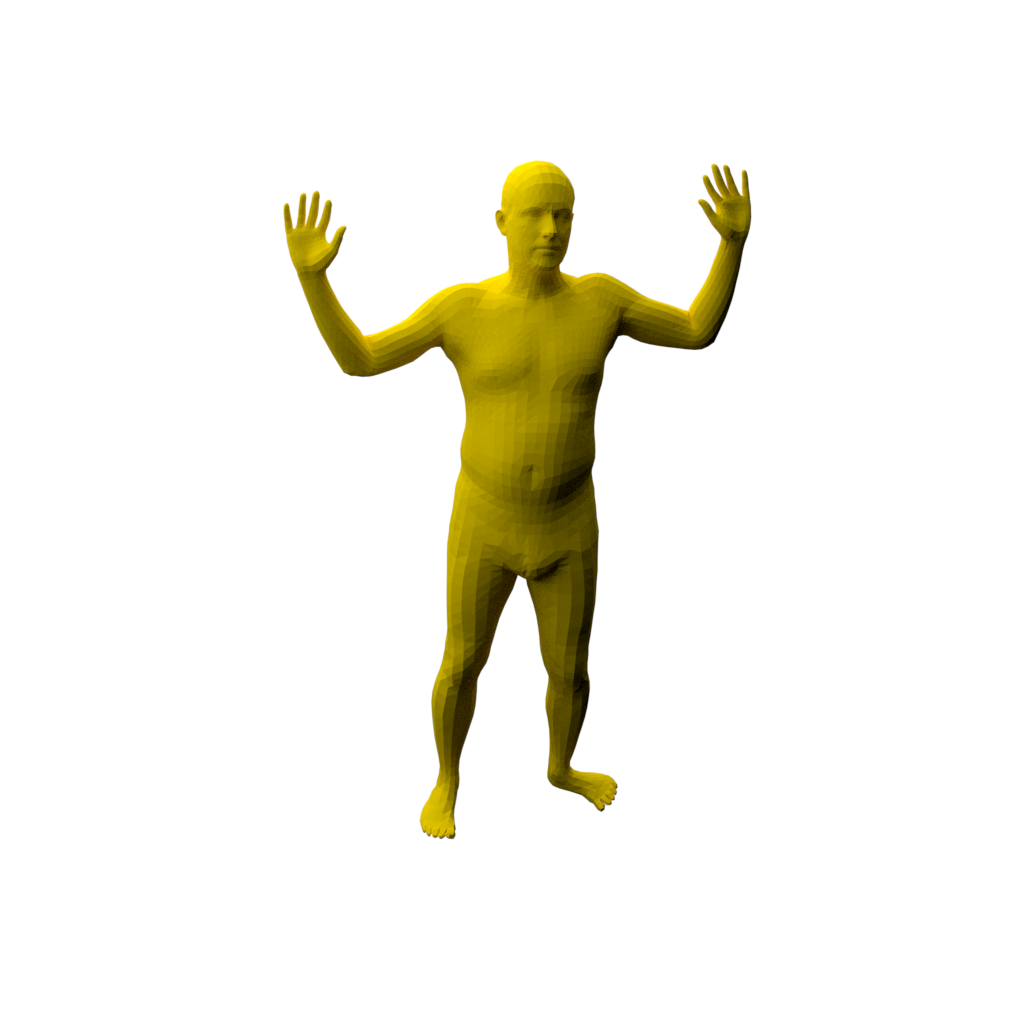}
 \end{subfigure}
 \begin{subfigure}[b]{0.15\textwidth}
    \includegraphics [trim=9cm 7cm 9cm 8.5cm, width=0.95\textwidth]{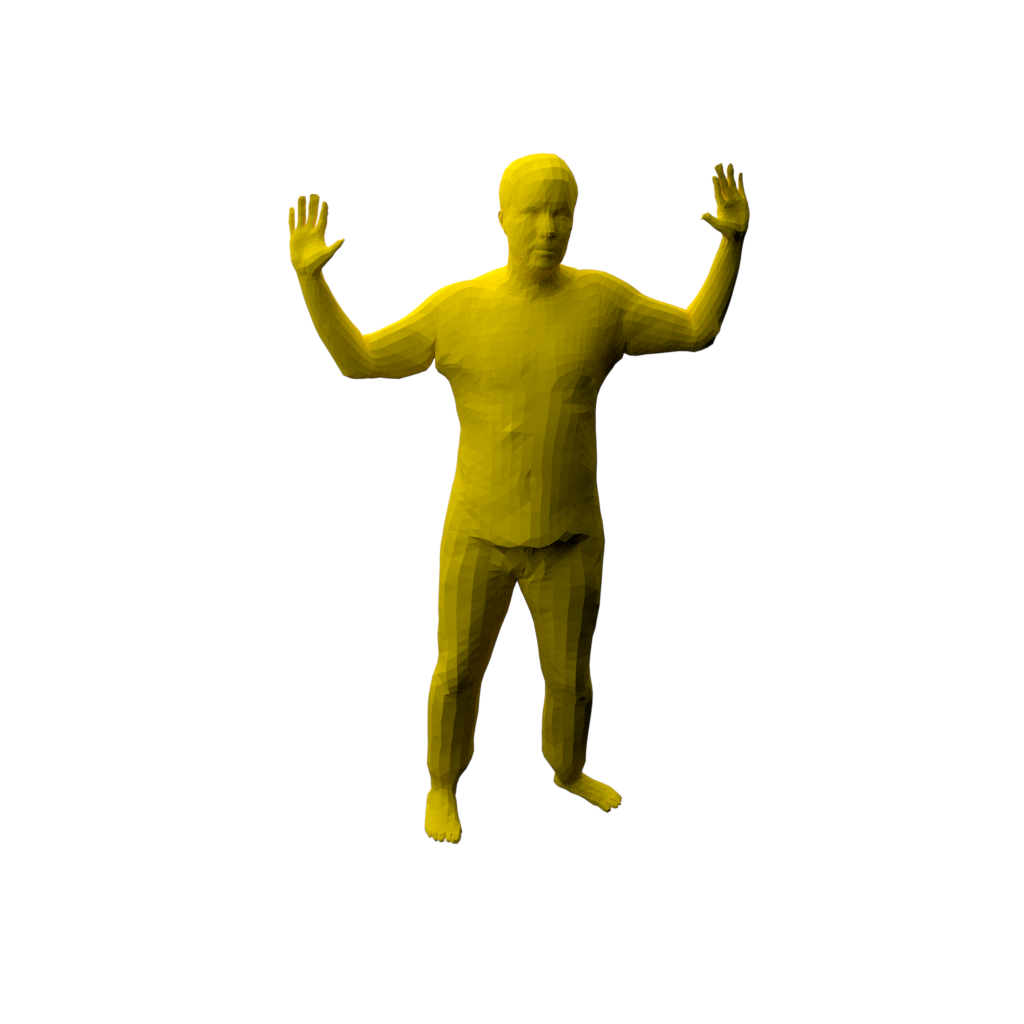}
 \end{subfigure}
 \begin{subfigure}[b]{0.15\textwidth}
    \includegraphics [trim=9cm 7cm 9cm 8.5cm, width=0.95\textwidth]{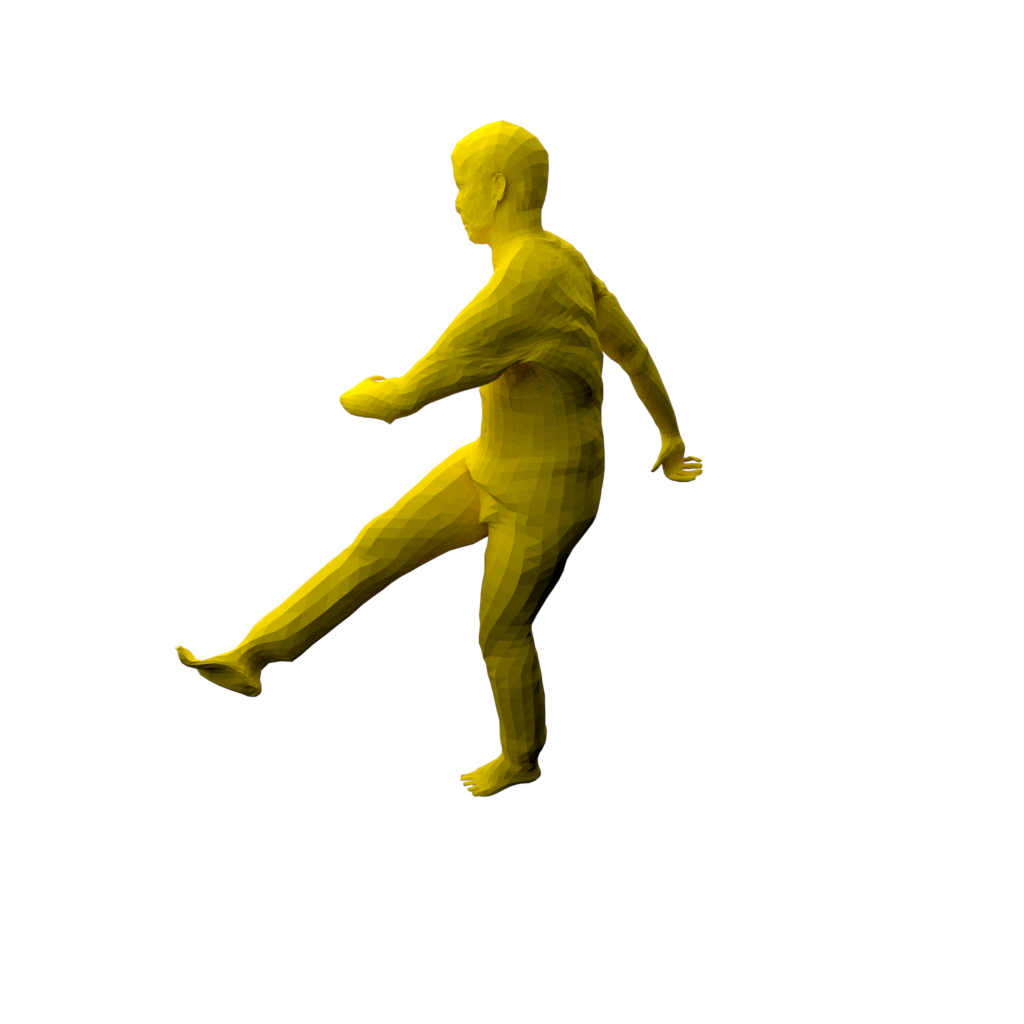}
 \end{subfigure} \\
 \begin{subfigure}[b]{0.15\textwidth}
    \includegraphics [trim=7cm 3cm 7cm 4.5cm, width=0.95\textwidth]{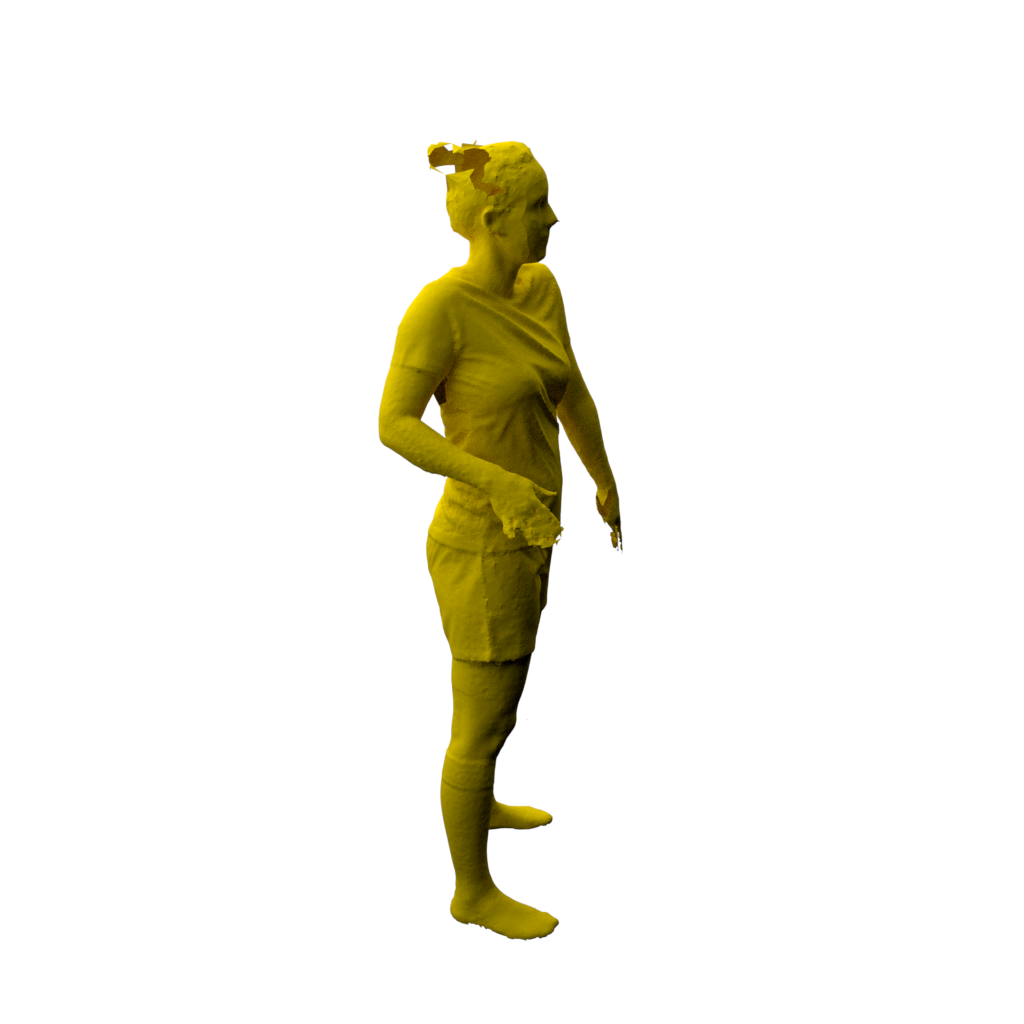}
    \caption{Raw scan}
\end{subfigure}
 \begin{subfigure}[b]{0.15\textwidth}
    \includegraphics [trim=7cm 3cm 7cm 4.5cm, width=0.95\textwidth]{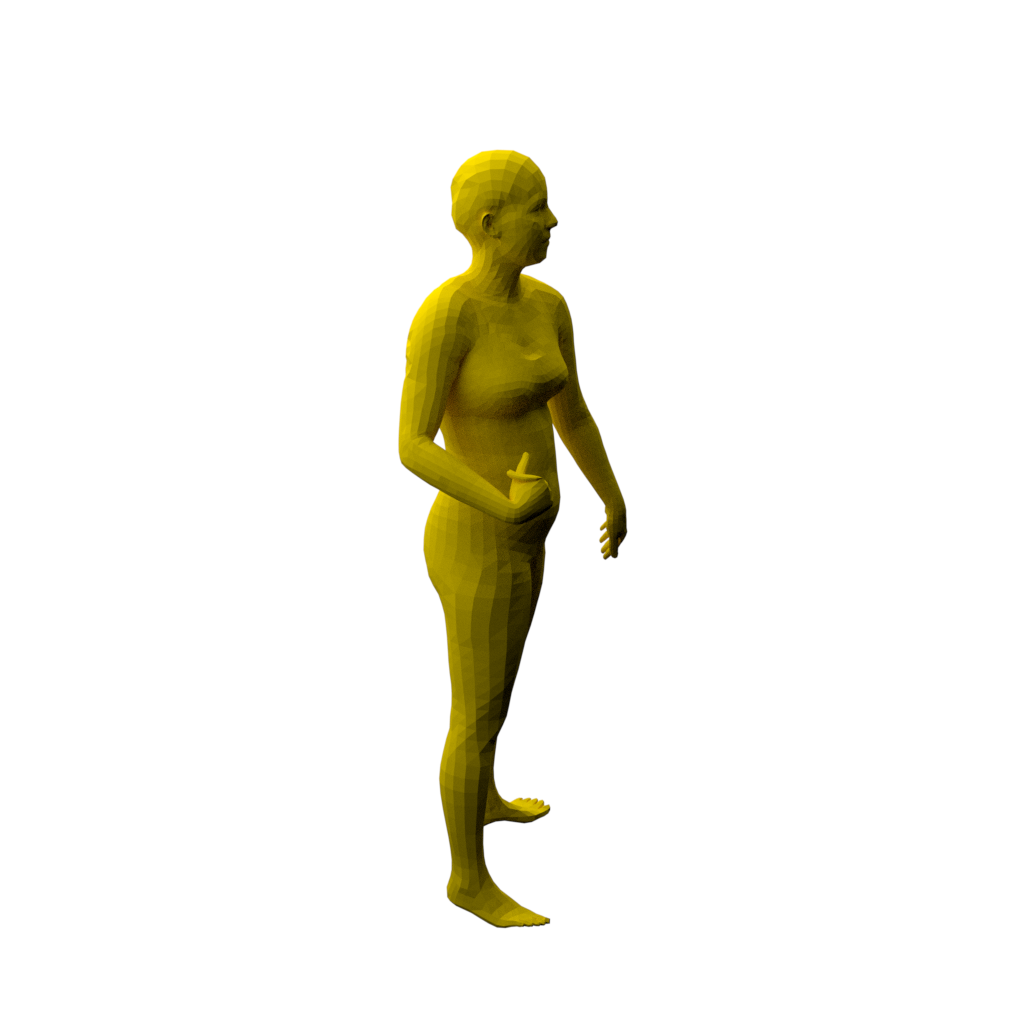}
    \caption{IPNet SMPL}
 \end{subfigure}
 \begin{subfigure}[b]{0.15\textwidth}
    \includegraphics [trim=7cm 3cm 7cm 4.5cm, width=0.95\textwidth]{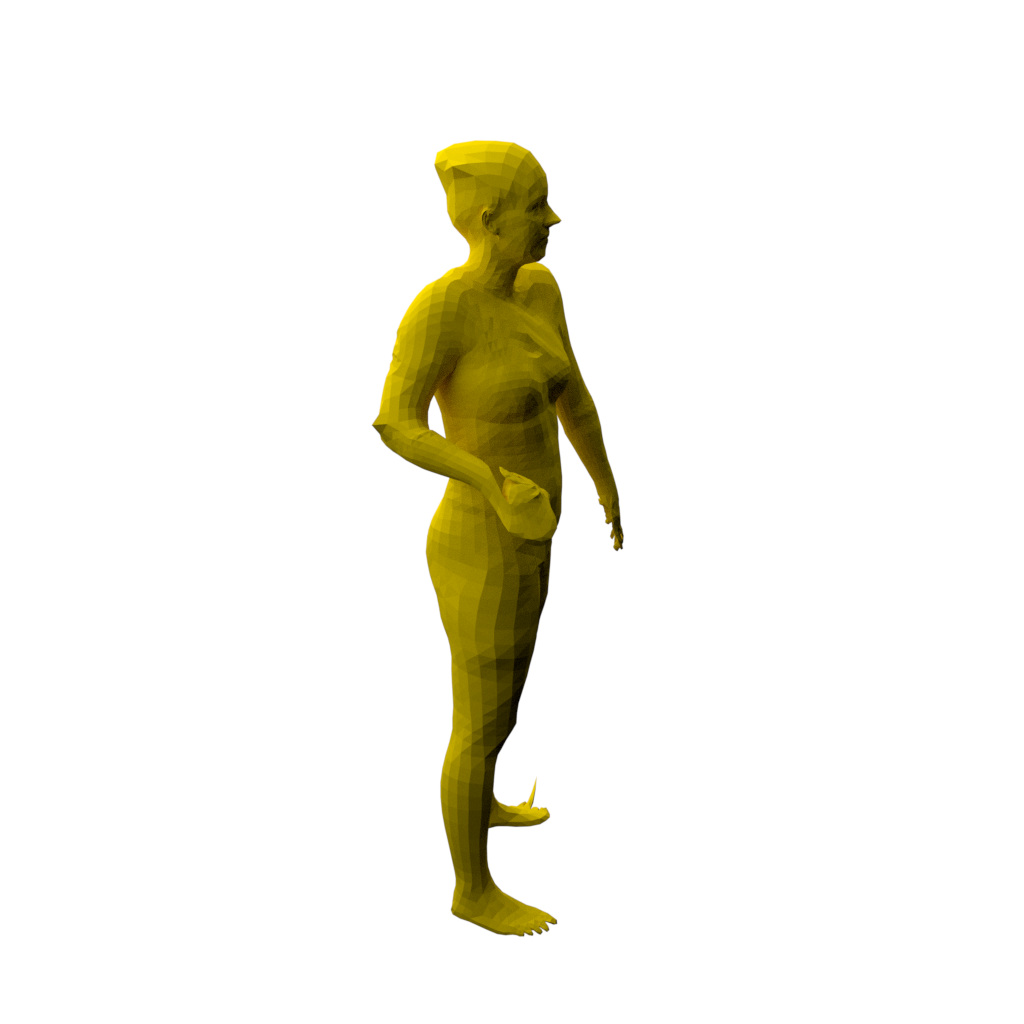}
    \caption{IPNet SMPL+D}
 \end{subfigure}
 \begin{subfigure}[b]{0.15\textwidth}
    \includegraphics [trim=7cm 3cm 7cm 4.5cm, width=0.95\textwidth]{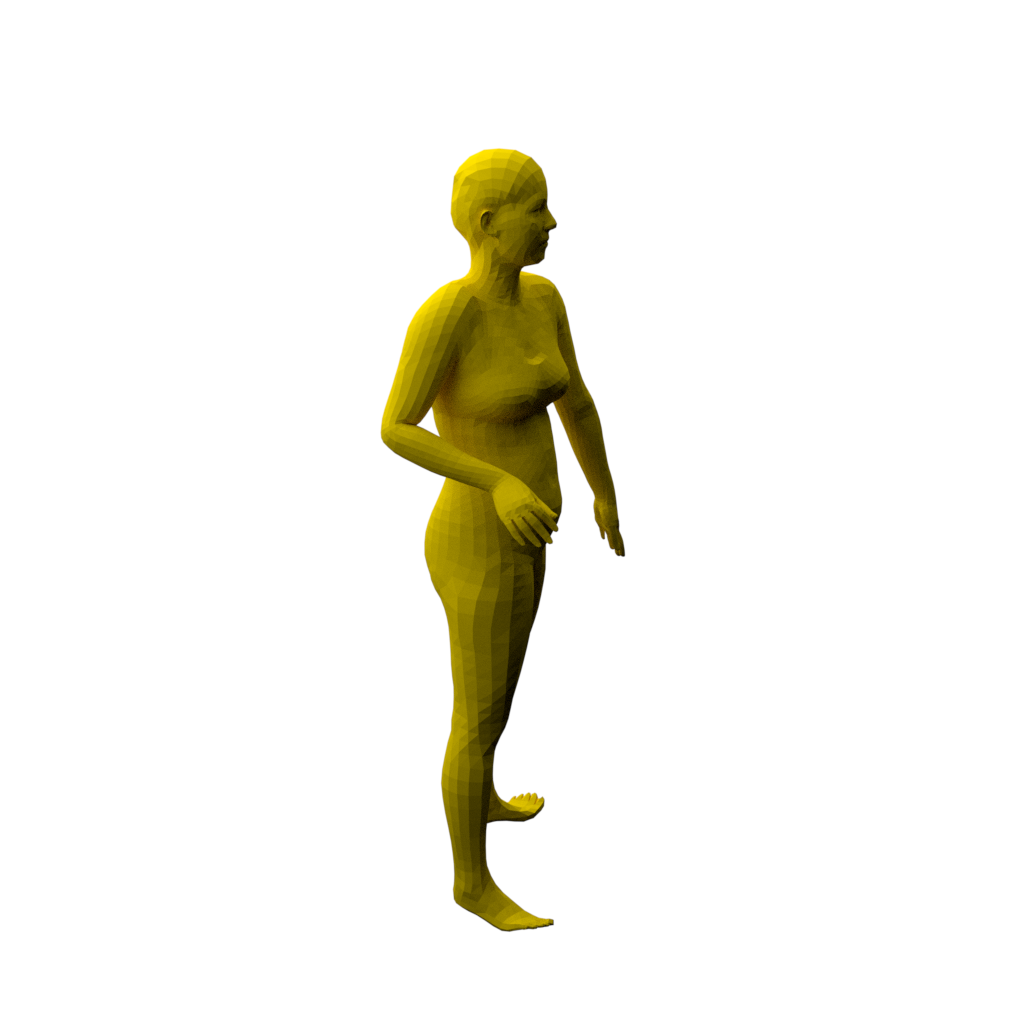}
    \caption{Ours SMPL}
 \end{subfigure}
 \begin{subfigure}[b]{0.15\textwidth}
    \includegraphics [trim=7cm 3cm 7cm 4.5cm, width=0.95\textwidth]{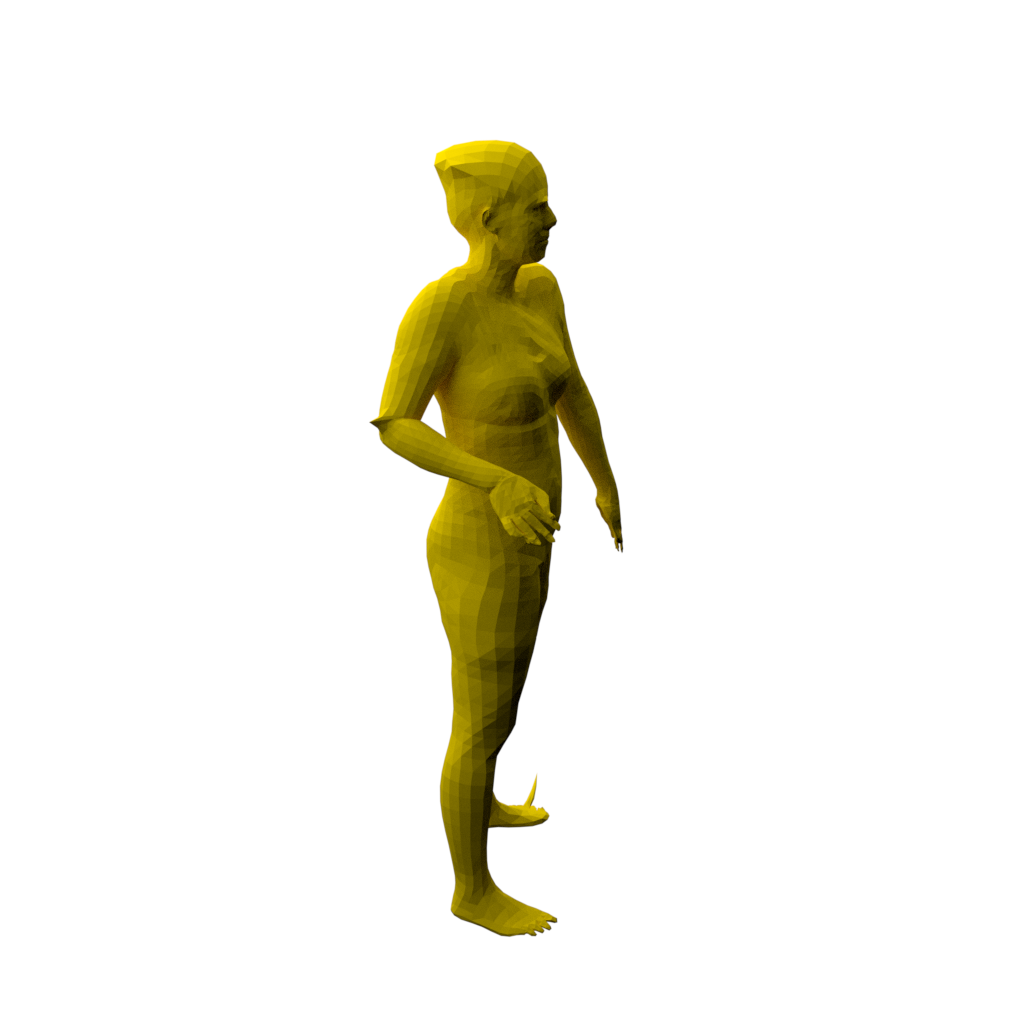}
    \caption{Ours SMPL+D}
 \end{subfigure}
 \begin{subfigure}[b]{0.15\textwidth}
    \includegraphics [trim=3cm 6cm 13cm 7cm, width=0.95\textwidth]{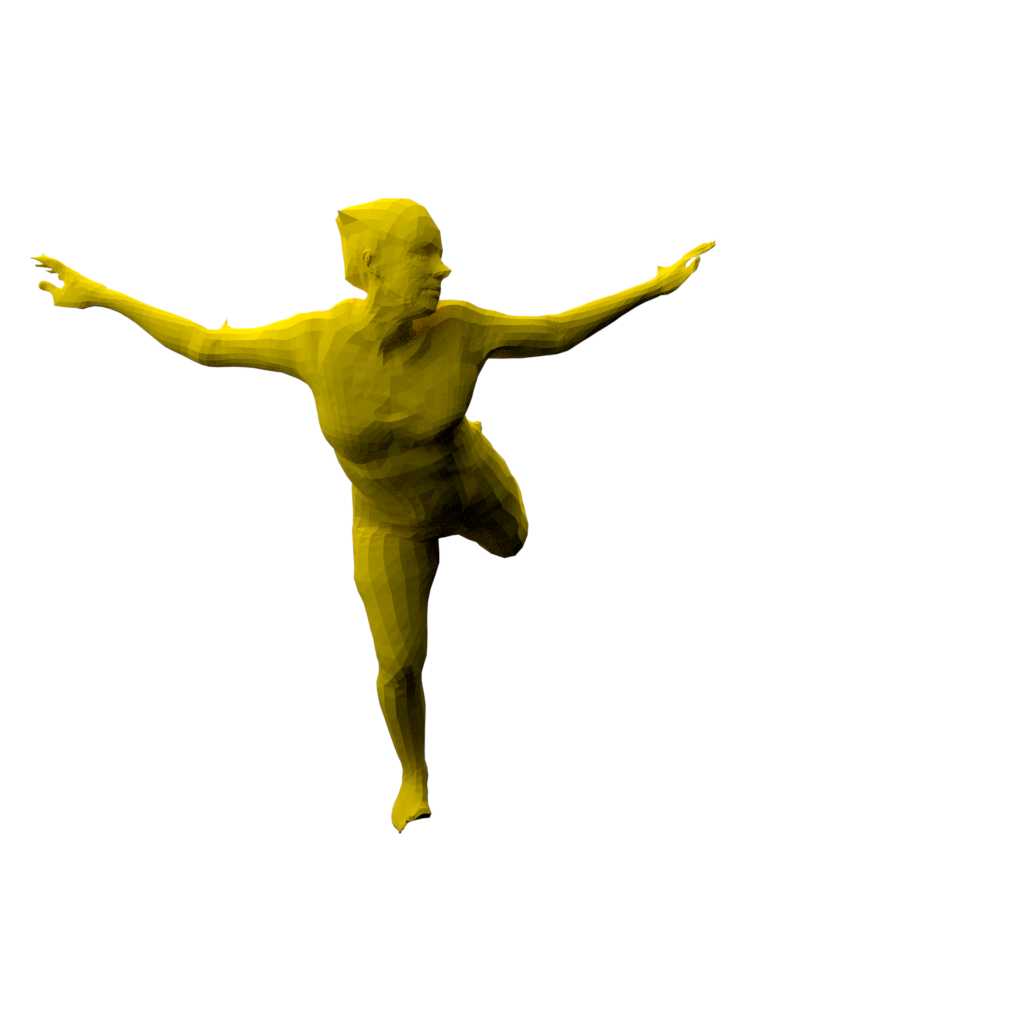}
    \caption{Ours re-posed}
 \end{subfigure}
 \vspace{-0.1in}
\caption{Although our model is trained on point clouds sampled from registered meshes, it generalizes well to raw scans. We show qualitative results on the BUFF dataset~\cite{BUFF}. Please see Appendix~\ref{appx:CAPE_qualitative}, \ref{appx:BUFF_qualitative} for more results.}
\label{fig:qualitative_results}
\end{figure*}

\boldparagraph{Evaluations}
In the remainder of this section, we will evaluate our approach along three dimensions. In Sec.~\ref{sec:eval_registration} we evaluate results on registrations from point clouds to SMPL/SMPL+D meshes. In Sec.~\ref{sec:eval_reconstruction}  we show how decoders with our PTF modules improve reconstruction quality while using less parameters than the IPNet decoder does. Lastly, in Sec.~\ref{sec:eval_rot} we demonstrate the superiority of our local-feature-based pose estimation framework, which reduces error by a factor of two, compared to a baseline that regresses poses directly from global point cloud features.

\subsection{Evaluation of Registration\protect\footnote{For more results, please see Appendix~\ref{appx:CAPE_quantitative}}}
\label{sec:eval_registration}
We follow the registration routine described in~\cite{Bhatnagar_ECCV2020} to register from point clouds to SMPL and SMPL+D models for under-cloth bodies (inner) and clothed bodies (outer), respectively. For our PTF models, $\bm{\theta}$ is initialized by solving Eq.~\eqref{eqn:lsq_obj} and factorizing out the rotation part of the results.
Registration error is measured as average per-vertex euclidean distance from registered SMPL/SMPL+D mesh to the ground-truth SMPL/SMPL+D mesh. Errors are formally reported in Table~\ref{tab:eval_registration}.

As we can see in Table~\ref{tab:eval_registration}, on the entire CAPE validation set, our PTF achieve \textit{18\%} improvements on registration error compared to~\cite{Bhatnagar_ECCV2020}. 

\begin{table}
 \begin{center}
 \renewcommand{\tabcolsep}{3.0pt}
 \begin{tabular}{| l | c | c |}
 \hline
 Method & Outer Err. & Inner Err. \\ \hline
 IPNet~\cite{Bhatnagar_ECCV2020} & 28.2 mm & 28.3 mm \\
 PTF-Piecewise & 23.9 mm & 23.6 mm  \\
 PTF-FC-w/o-pose & 27.5 mm & 27.7 mm \\
 PTF-FC & \textbf{23.1 mm} & \textbf{23.1 mm} \\ \hline
 \end{tabular}
 \end{center}
 \vspace{-0.1in}
 \caption{Registration error on the full CAPE validation set (3965 frames). PTF-FC-w/o-pose indicates fitting SMPL/SMPL+D to our PTF-FC outputs without our pose initialization.}
 \label{tab:eval_registration}
\end{table}


\subsection{Evaluation of Reconstruction}
\label{sec:eval_reconstruction}
For evaluating reconstruction quality on the validation set, we report average intersection over union (mIoU $\uparrow$) of 100K random query points (50K sampled uniformly, 25K sampled near the clothed mesh, 25K sampled near the minimally-clothed mesh) and L1 Chamfer distance (CD) for both clothed surfaces (outer CD $\downarrow$) and under-cloth surfaces (inner CD $\downarrow$). 

\subsubsection{Ablation on Encoders}
\label{sec:ablat_encoders}
IPNet uses IFNet~\cite{IFNet} as encoder that takes a voxel grid of size $128^3$ as input, which we find very memory hungry and slow to run. With the aforementioned dataset setups, we can only use a batch size of 4. Given the number of iterations required, it would take weeks to train IPNet with $128^3$ voxel inputs on the CAPE dataset. Reducing the voxel input to $64^3$ helps with the training time, but it still takes about a week to train on CAPE. Instead, we choose to use a recently proposed convolutional feature encoder~\cite{ConvONet} with 3-plane-features, which we denote as ConvONet, as a more efficient and more accurate alternative to IFNet $64^3$. A brief comparison of IFNet and ConvONet can be found in Table~\ref{tab:ablation_encoder}. Clearly, ConvONet is more practical on the CAPE dataset. We further apply data-augmentation that rotates the input point clouds randomly, which gives another boost in accuracy (ConvONet+aug in Table~\ref{tab:ablation_encoder}). We will thus use ConvONet as encoder and apply the random rotation to inputs during training throughout all remaining experiments.

\begin{table}
 \begin{center}
 \renewcommand{\tabcolsep}{3.0pt}
 \begin{tabular}{| l | c | c | c | c |}
 \hline
 Encoder & VRAM & mIoU & Outer CD & Inner CD \\ \hline
 IFNet~\cite{IFNet} & 1.75G & 86.7\% & 0.0191 & 0.0196 \\
 ConvONet~\cite{ConvONet} & 1.25G & 87.8\% & 0.0156 & 0.0162 \\
 ConvONet~\cite{ConvONet}+aug & \textbf{1.25G} & \textbf{88.6\%} & \textbf{0.0151} & \textbf{0.0157} \\ \hline
\end{tabular}
 \end{center}
 \vspace{-0.1in}
 \caption{We fix the decoder to be the IPNet baseline and ablate on different encoders. VRAM indicates per-sample GPU memory usage during training. ConvONet achieves higher reconstruction quality with less memory usage.}
 \label{tab:ablation_encoder}
\end{table}

\subsubsection{Ablation on Decoders}
\label{sec:ablat_heads}
\begin{table}
 \begin{center}
 \renewcommand{\tabcolsep}{3.0pt}
 \begin{tabular}{| l | c | c | c | c |}
 \hline
 Method & \#params & mIoU & Outer CD & Inner CD \\ \hline
 IPNet~\cite{Bhatnagar_ECCV2020} & 2.2M & 88.6\% & 0.0151 & \textbf{0.0157} \\ \hline
 PTF-Piecewise & \textbf{1.2M} & 89.4\% & 0.0152 & 0.0162 \\
 PTF-FC & 1.4M & \textbf{89.7}\% & \textbf{0.0148} & 0.0158 \\ \hline
 \multicolumn{5}{| c |}{Ablation on Point Correspondence Prediction} \\ \hline
 Method & \#params & mIoU & Outer CD & Inner CD \\ \hline
 IPNet+Corr & 2.3M & 88.9\% & 0.0153 & 0.0159 \\
 TF-FC & 7.3M & 88.1\% & 0.0154 & 0.0161 \\ \hline
 \end{tabular}
 \end{center}
 \vspace{-0.1in}
 \caption{Ablation study on decoder architectures. Our PTF models achieve comparable or better reconstruction quality while using less parameters than the baseline.}
 \label{tab:ablation_heads}
\end{table}

We report reconstruction accuracy of IPNet and our PTF models in Table~\ref{tab:ablation_heads}. PTF-Piecewise achieves comparable accuracy to IPNet while using only 54\% parameters of IPNet. PTF-FC uses 64\% parameters of IPNet, but achieves a noticeable improvement in clothed surface reconstruction.

We also ablate on different strategies of predicting point correspondences to demonstrate the necessity of using our PTF module. We first construct IPNet-Corr, which extends the final layer of IPNet's occupancy classifier to also predict the point correspondences $\hat{\mathbf{x}}$ directly. This variant applies occupancy inference in the posed-space. IPNet+Corr gives slightly better mIoU but worse Chamfer distance than the IPNet baseline. 

Next, we construct a variant, TF-FC. TF-FC replaces the PTF module with a 4-layer MLP with $128 \times 14$ channels. This simplification results in degraded performance.

\subsection{Evaluation of Rotation Estimation}
\label{sec:eval_rot}
\begin{table}
 \begin{center}
 \renewcommand{\tabcolsep}{3.0pt}
 \begin{tabular}{| l | c | c | c | c |}
 \hline
 Method & Per-vetex error \\ \hline
 Baseline & 74.4 mm  \\ \hline
 PTF-Piecewise & 34.8 mm \\
 PTF-FC & \textbf{34.1 mm} \\ \hline
 \end{tabular}
 \end{center}
 \vspace{-0.1in}
 \caption{Comparison of our pose estimation approach against the baseline that regresses a continuous 6D representation~\cite{Zhou_2019_CVPR} of rotation matrices directly from global point cloud features.}
 \label{tab:pose_estimation}
\end{table}

In this experiment, we demonstrate superiority of our local-feature-based pose estimation pipeline versus a direct regression baseline. The baseline takes 5K points as input, encodes the point cloud with PointNet~\cite{qi2017pointnet}, and then regresses rotation matrices in 6D continuous representation~\cite{Zhou_2019_CVPR}. We evaluate the per-vertex error on the SMPL mesh, assuming ground-truth shape and global-translation are known. For our method, $\bm{\theta}^*$ is obtained by solving Eq.~\eqref{eqn:lsq_obj}-\eqref{eqn:part_labels} without optimizing Eq.~\eqref{eqn:SMPL_optim}. 
Results are shown in Table~\ref{tab:pose_estimation}. We reduce the per-vertex error significantly, from 74.4mm to 34.1mm.

%% file: arxiv_diff_nasa_ours.tex
\section{Difference Between the Original NASA~\cite{Deng_ECCV2020} and Our Modified NASA}
\label{appx:NASA_diff}
In the original NASA~\cite{Deng_ECCV2020} paper, the bone transformations $\{ \mathbf{B}_b^{\prime} \}$ are defined as:
\begin{align}
\label{eqn:bone_transformations_nasa}
    \mathbf{B}_b^{\prime} &= G_b (\bm{\beta}, \bm{\theta})  \\  
    G_b (\bm{\beta}, \bm{\theta}) &= \prod_{b^{\prime} \in \mathbf{A}(b)} \begin{bmatrix}   \mathbf{R}_{b^{\prime}} (\bm{\theta}) & \mathbf{t}_{b^{\prime}} - \mathbf{t}_{\bar{b^{\prime}}} \\ \mathbf{0} & 1 \end{bmatrix} \\
    \{ \mathbf{t}_b \} &= \mathcal{J}(\bm{\beta})
\end{align}
The $\{ \mathbf{B}_b^{\prime} \}$ are relative to \textit{the origin}, whereas in our formulation we have: 
\begin{align}
    \mathbf{B}_b &= G_b (\bm{\beta}, \bm{\theta}) G_b (\bm{\beta}, \mathbf{0})^{-1} \label{eqn:rel_bone_transformations_appx}\\  
    G_b (\bm{\beta}, \bm{\theta}) &= \prod_{b^{\prime} \in \mathbf{A}(b)} \begin{bmatrix}   \mathbf{R}_{b^{\prime}} (\bm{\theta}) & \mathbf{t}_{b^{\prime}} - \mathbf{t}_{\bar{b^{\prime}}} \\ \mathbf{0} & 1 \end{bmatrix} \label{eqn:abs_bone_transformations_appx}\\
    \{ \mathbf{t}_b \} &= \mathcal{J}(\bm{\beta}) \label{eqn:joint_locations_appx}
\end{align}
Eq.~\eqref{eqn:rel_bone_transformations_appx}-\eqref{eqn:joint_locations_appx} are the same as Eq.~\eqref{eqn:rel_bone_transformations}-\eqref{eqn:joint_locations}.
The $\{ \mathbf{B}_b \}$ are relative to \textit{the rest-pose joints}.


%% file: arxiv_model_archs.tex
\section{Model Architectures}
\label{appx:archs}
\begin{figure*}[t]
\begin{center}
\begin{subfigure}[b]{0.45\textwidth}
    \includegraphics[clip, trim=5.80cm 13.5cm 5.15cm 2cm, width=0.95\textwidth]{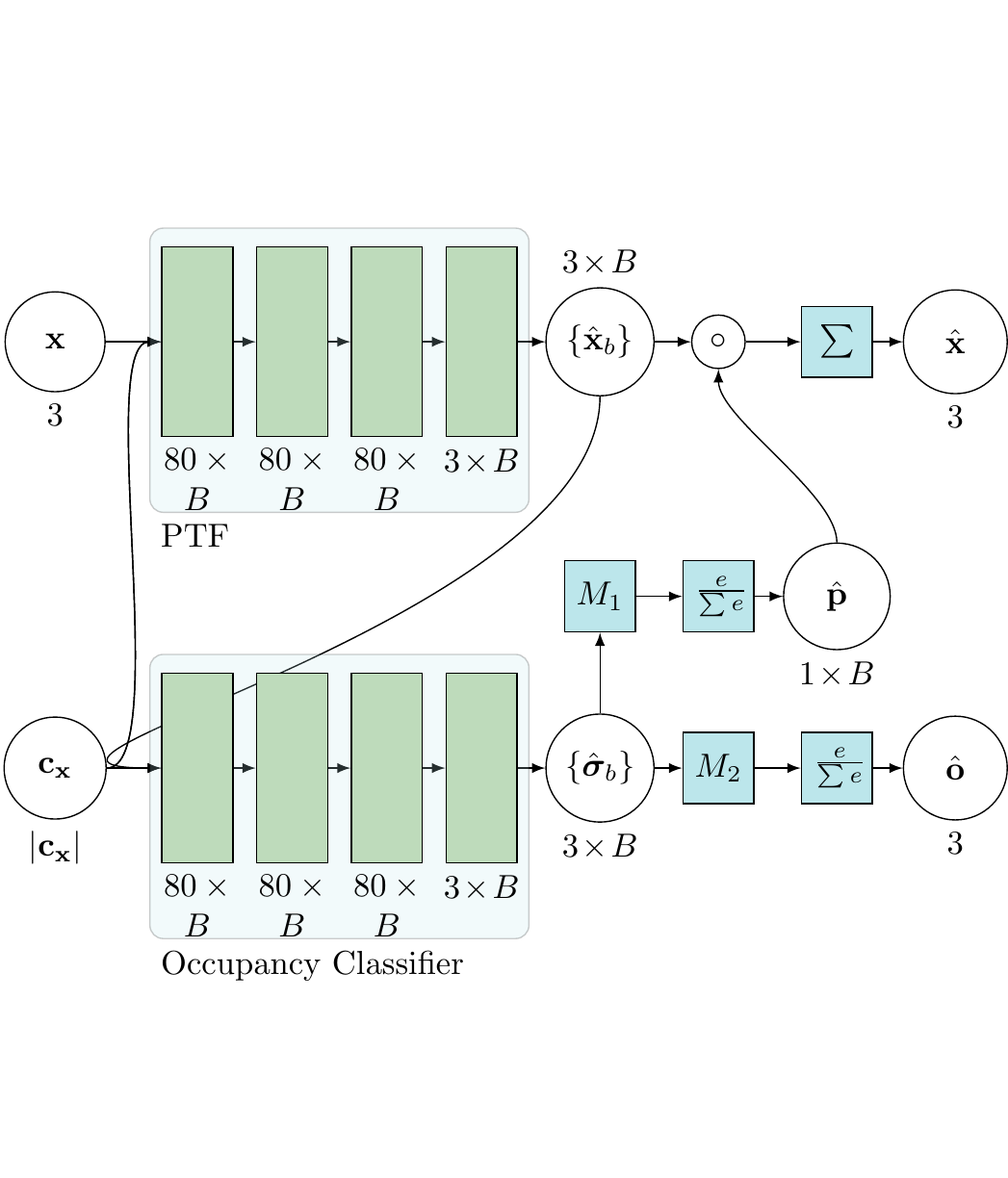}
    \caption{PTF-Piecewise}
    \label{fig:PTFs-Piecewise_arch}
\end{subfigure}
\begin{subfigure}[b]{0.45\textwidth}
    \includegraphics[clip, trim=6cm 11.65cm 5cm 4.25cm, width=0.95\textwidth]{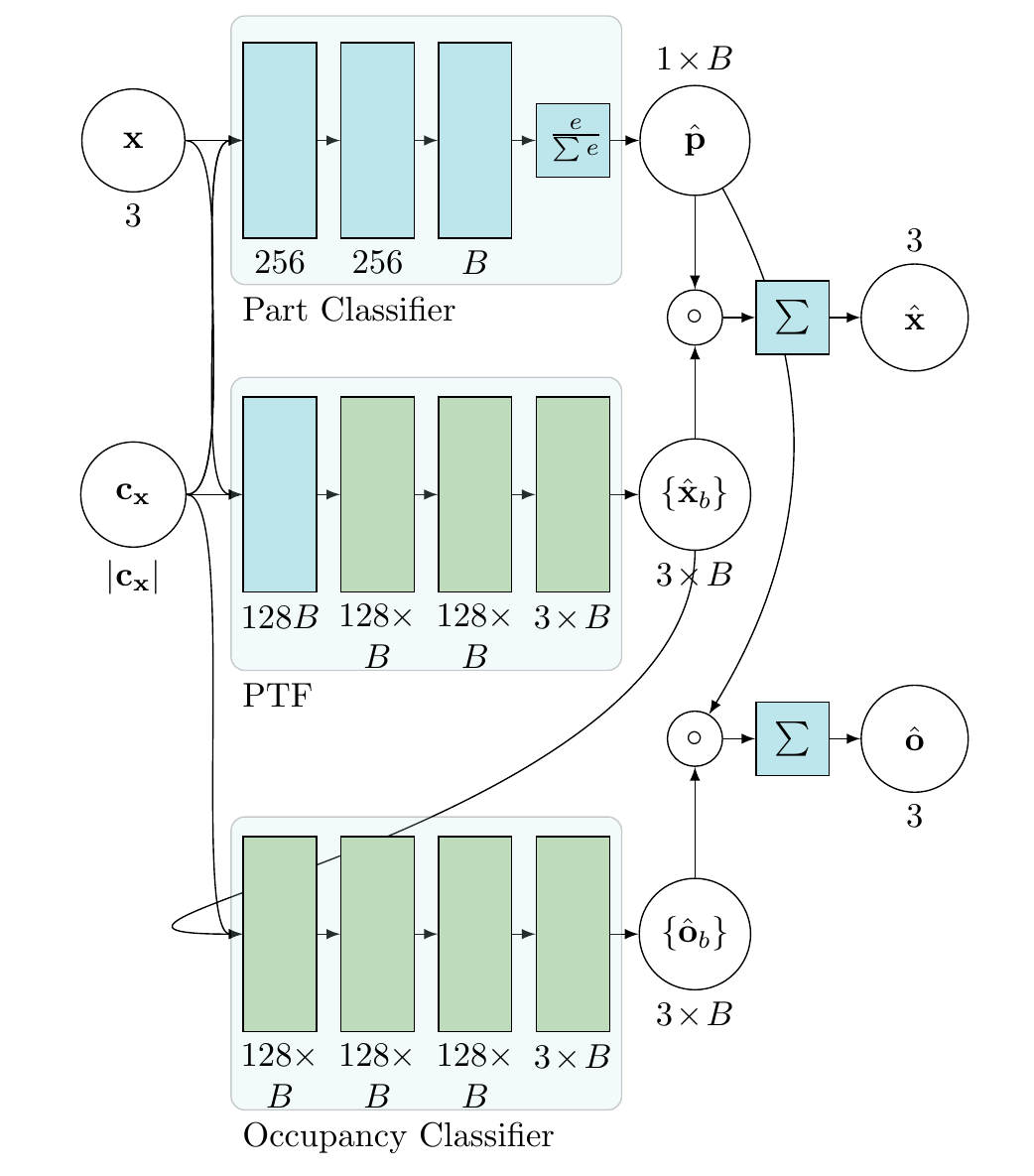}
    \caption{PTF-FC}
    \label{fig:PTFs-FC_arch}
\end{subfigure}
\end{center}
\caption{Illustration for model architectures. Blue rectangles indicate 1D convolution layers, green rectangles indicate 1D group-convolutions with $B$ groups, $\frac{e}{\sum e}$ squares indicate the softmax function, $\circ$ indicates the Hadamard product with broadcasting, $\sum$ squares indicate summation over part dimension, $M_1$ and $M_2$ squares indicate $\max$ over the occupancy dimension and part dimension, respectively. Both PTF-Piecewise and PTF-FC take a query point $\mathbf{x}$ and the corresponding local point feature $\mathbf{c}_{\mathbf{x}}$ as input, and output a multi-class occupancy probability $\hat{\mathbf{o}}$ and the rest-pose correspondence $\hat{\mathbf{x}}$.}
\label{fig:archs}
\end{figure*}
We provide illustrations for our model architectures in Fig.~\ref{fig:archs}. Note that throughout our implementations, we use 1D convolutions with kernel size 1 to substitute the fully-connected layers used in typical occupancy networks. This enables us to use the off-the-shelf implementation of grouped convolutions~\cite{alexnet} in PyTorch~\cite{pytorch} to represent piecewise occupancy classifiers $\{ \bar{\mathcal{O}}^b_{\omega} \}$ and PTF $\{ \bar{\mathcal{T}}^b_{\psi} \}$.

Fig.~\ref{fig:PTFs-Piecewise_arch} shows our fully-piecewise model, PTF-Piecewise. The key observation we make here is that, the pre-activation occupancy logits $\{ \bm{\sigma}_b \} \in \mathbb{R}^{3 \times B}$ can be used to compute the softmax probability over parts. To construct inputs to the occupancy classifier module, we concatenate the local point cloud feature $\mathbf{c}_{\mathbf{x}}$ to each $\hat{\mathbf{x}}_b, \forall b \in \{ 1, \cdots, B \}$, thus a piecewise occupancy function $\bar{\mathcal{O}}^b_{\omega}$ of bone $b$ takes the concatenation of $\mathbf{c}_{\mathbf{x}}$ and $\hat{\mathbf{x}}_b$ as the input and outputs the occupancy logits $\hat{\bm{\sigma}}_b$. We set $B=24$ for PTF-Piecewise.

Fig.~\ref{fig:PTFs-FC_arch} shows our PTF-FC model. Here we use a similar structure as in IPNet~\cite{Bhatnagar_ECCV2020} which predicts part probabilities with a separate part classifier. The difference between IPNet and our PTF-FC is that we apply PTF to query point $\mathbf{x}$ before feeding it to the occupancy classifier. The first layer (in blue) of the PTF module takes the concatenation of $\mathbf{c}_{\mathbf{x}}$ and $\mathbf{x}$ as the input and outputs a $128B$ dimensional feature for later stages. We set $B=14$ for PTF-FC.

%% file: arxiv_quantitative_cape.tex
\section{Additional Quantitative Results on the CAPE Dataset}
\label{appx:CAPE_quantitative}
\begin{table}
 \begin{center}
 \renewcommand{\tabcolsep}{3.0pt}
 \footnotesize
 \begin{tabular}{| l | c | c | c | c |}
 \hline
 Method & Outer Err. & Max Outer Err. & Inner Err. & Max Inner Err. \\ \hline
 IPNet & 28.2 mm & 562.7 mm & 28.3 mm & 544.2 mm \\
 IPNet-128 & 26.4 mm & 564.4 mm & 26.9 mm & 546.5 mm \\
 Stitched Puppet~\cite{Zuffi:CVPR:2015} & 36.1 mm & 454.9 mm & NA & NA \\ 
 3D-CODED~\cite{groueix2018b} & 23.7 mm & 614.8 mm & NA & NA \\ \hline
 PTF-FC & 23.1 mm & \textbf{76.6 mm} & 23.1 mm & 88.9 mm \\
 PTF-FC-128 & \textbf{21.2 mm} & 83.6 mm & \textbf{21.4 mm} & \textbf{82.3 mm} \\ \hline
 \end{tabular}
 \end{center}
 \caption{Additional registration evaluation on the CAPE dataset. IPNet-128 and PTF-FC-128 indicate IPNet model and our model trained with encoder resolution of $128 \times 3$, where size of the input point cloud remains 5K.}
 \label{tab:eval_registration_ex}
 \vspace{-0.15in}
\end{table}

In this section we compare our approach to other related baselines, namely 3D-CODED~\cite{groueix2018b} and the Stitched Puppet~\cite{Zuffi:CVPR:2015}. Note that to our best knowledge, IPNet is the only published method that addresses automatic {\bf model-fitting} of {\bf both} under-cloth body and clothed body, given unoriented, sparse point-clouds of {\bf clothed} humans. Both IPNet and ours can produce controllable parameters and clothed avatars from input point clouds.

3D-CODED~\cite{groueix2018b} only focuses on correspondence prediction. Their method can not produce controllable parameters for under-cloth or clothed body models. Furthermore, 3D-CODED uses exhaustive initial rotation search and is allowed to deform the template freely, while we can only optimize for $\bm{\theta}$, $\bm{\beta}$, $\mathbf{t}$ and $\mathbf{D}$, where $\bm{\theta}$, $\bm{\beta}$ are limited to SMPL parameter space, and $\mathbf{D}$ is in the A-pose and regularized to be small.

The Stitched Puppet~\cite{Zuffi:CVPR:2015} tackles model-fitting as a pure optimization problem via belief propagation. Since~\cite{Zuffi:CVPR:2015} is based on SCAPE~\cite{SCAPE} topology, we fit it to FAUST~\cite{Bogo:CVPR:2014} scans with the neutral pose, and then obtain correspondence from SCAPE vertices to SMPL vertices via barycentric interpolation.

We report the numbers in Tab.~\ref{tab:eval_registration_ex}. For clothing surface registration, 3D-CODED produces slightly inferior results than ours although it is less constrained. The results from the Stitched Puppet are inferior to all others probably due to 1) the sensitivity of belief propagation to noise and 2) the fact that clothed surfaces vary more than SCAPE can capture. Most importantly, our approach {\bf completely avoids} catastrophic failures ($> 100$ mm error) that happen for the baselines, demonstrating strong generalization.

%% file: arxiv_qualitative_cape.tex
\section{Additional Qualitative Results on the CAPE Dataset}
\label{appx:CAPE_qualitative}
\begin{figure*}[t]
\centering
 \begin{subfigure}[b]{0.15\textwidth}
    \includegraphics [trim=10cm 7cm 10cm 4cm, width=0.95\textwidth]{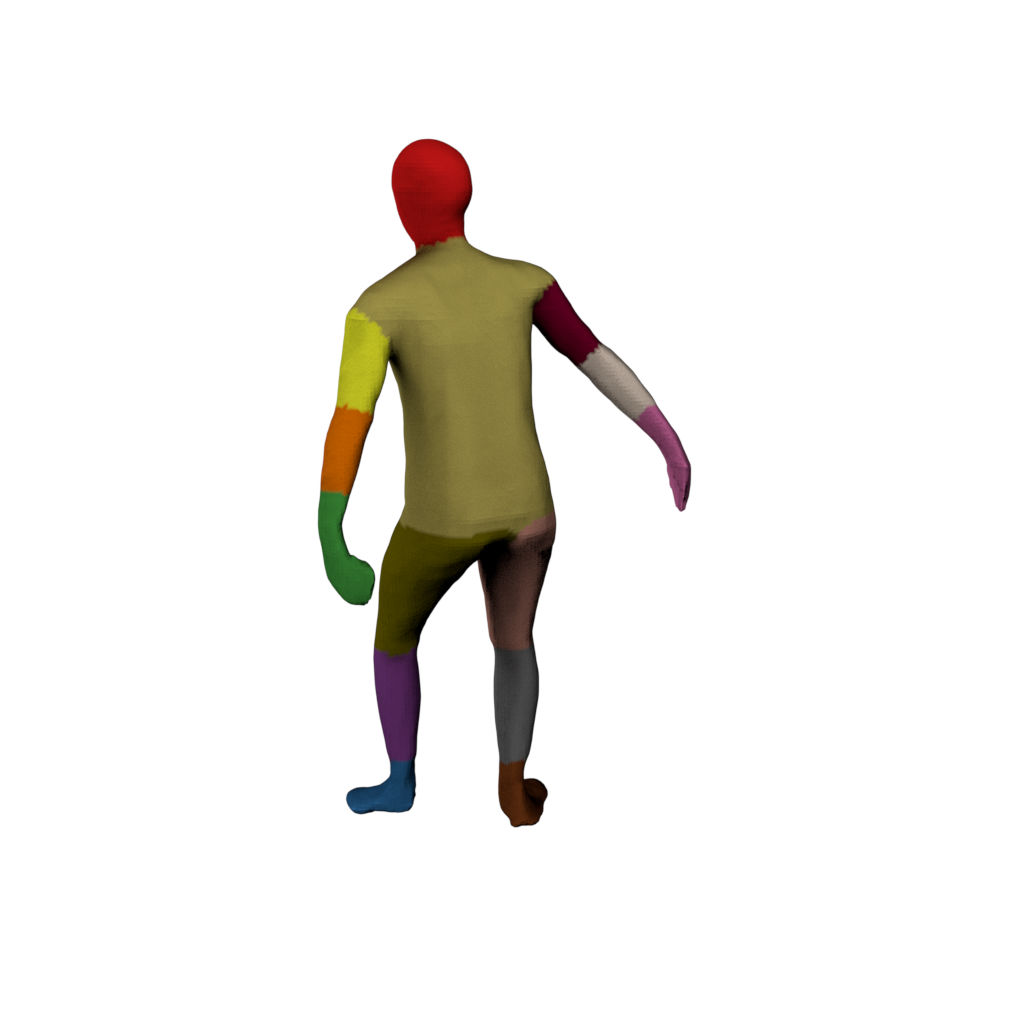}
    \caption{IPNet surface}
\end{subfigure}
 \begin{subfigure}[b]{0.15\textwidth}
    \includegraphics [trim=10cm 7cm 10cm 4cm, width=0.95\textwidth]{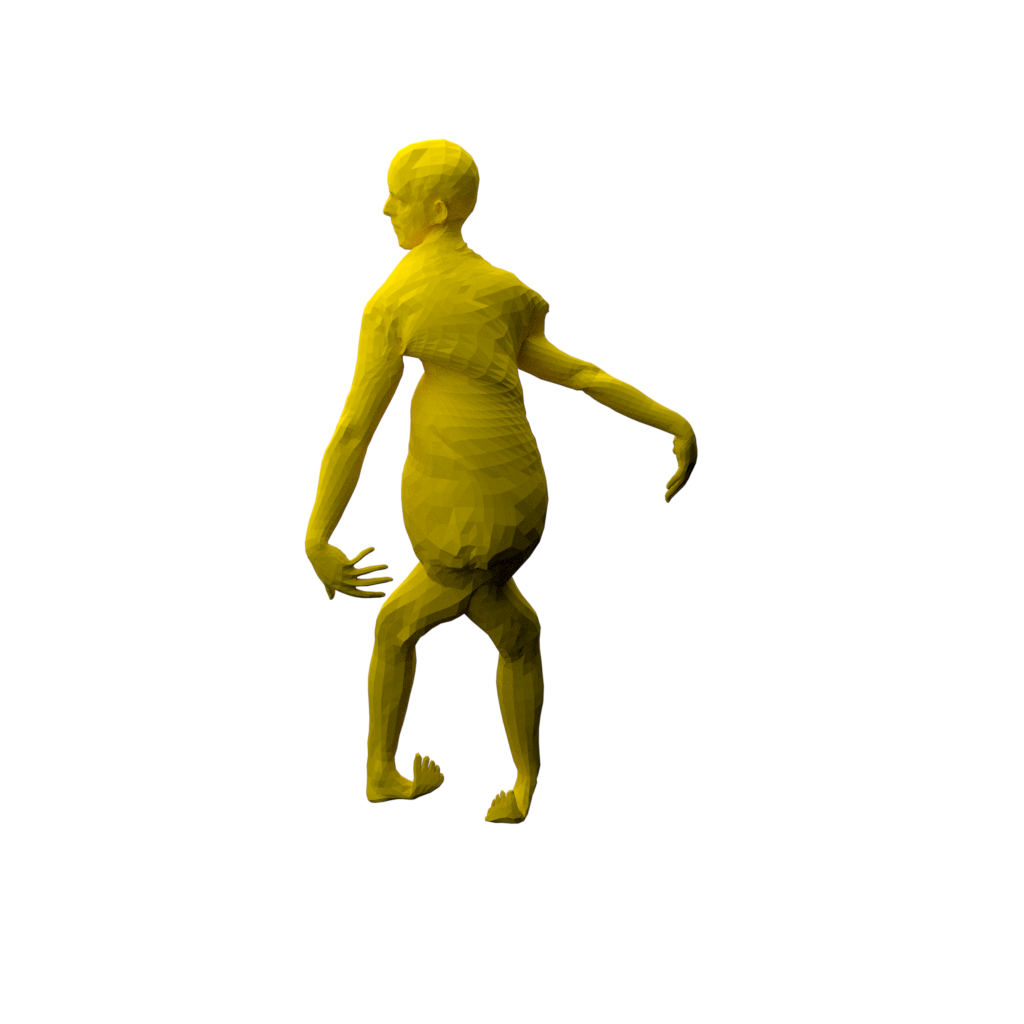}
    \caption{IPNet SMPL}
 \end{subfigure}
 \begin{subfigure}[b]{0.15\textwidth}
    \includegraphics [trim=10cm 7cm 10cm 4cm, width=0.95\textwidth]{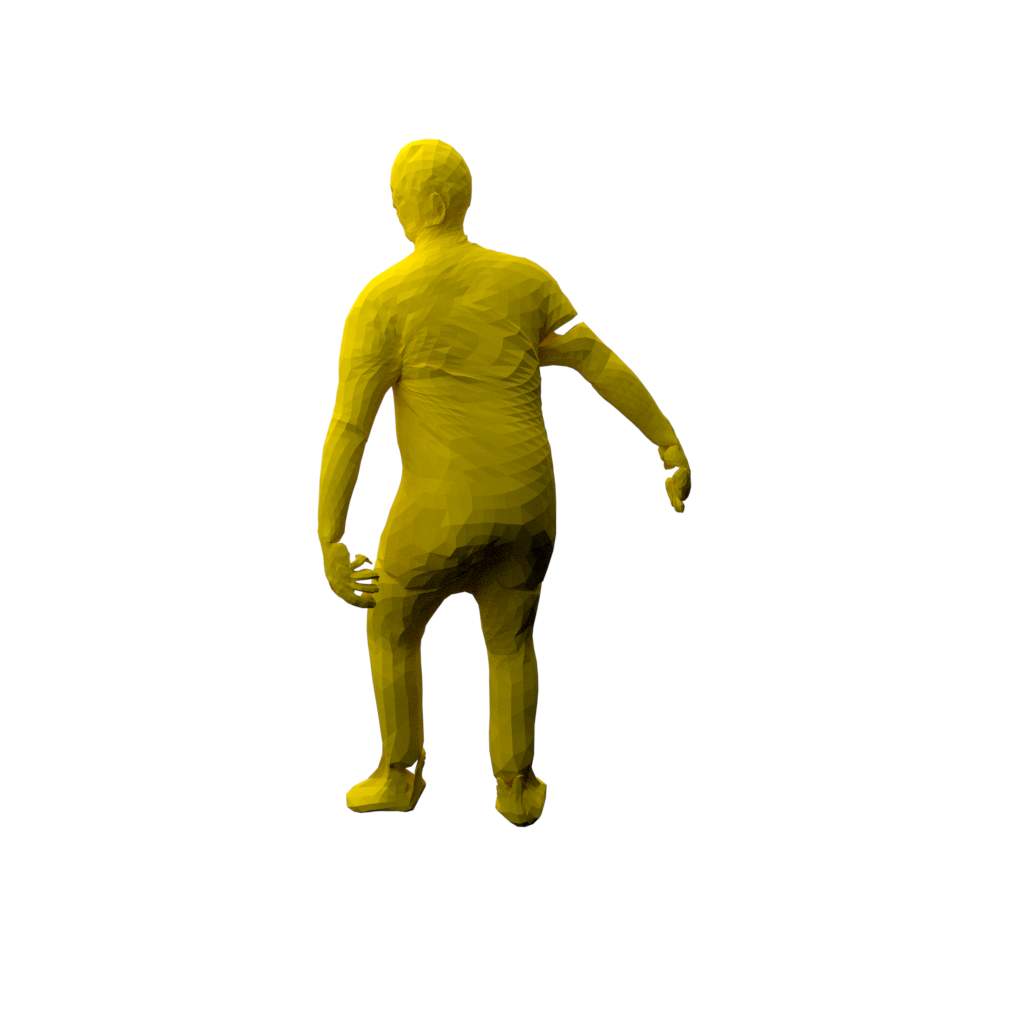}
    \caption{IPNet SMPL+D}
 \end{subfigure}
 \begin{subfigure}[b]{0.15\textwidth}
    \includegraphics [trim=10cm 7cm 10cm 4cm, width=0.95\textwidth]{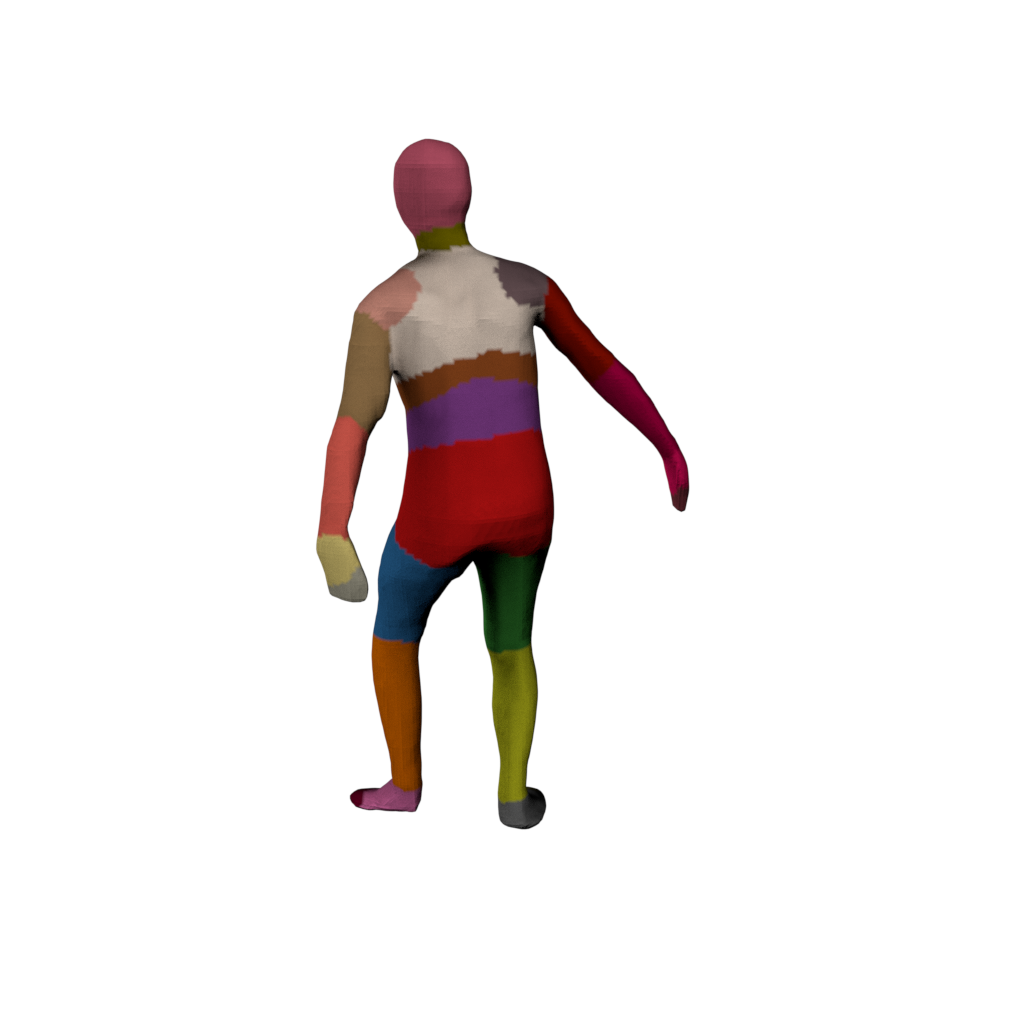}
    \caption{Ours surface}
 \end{subfigure}
\begin{subfigure}[b]{0.15\textwidth}
    \includegraphics [trim=10cm 7cm 10cm 4cm, width=0.95\textwidth]{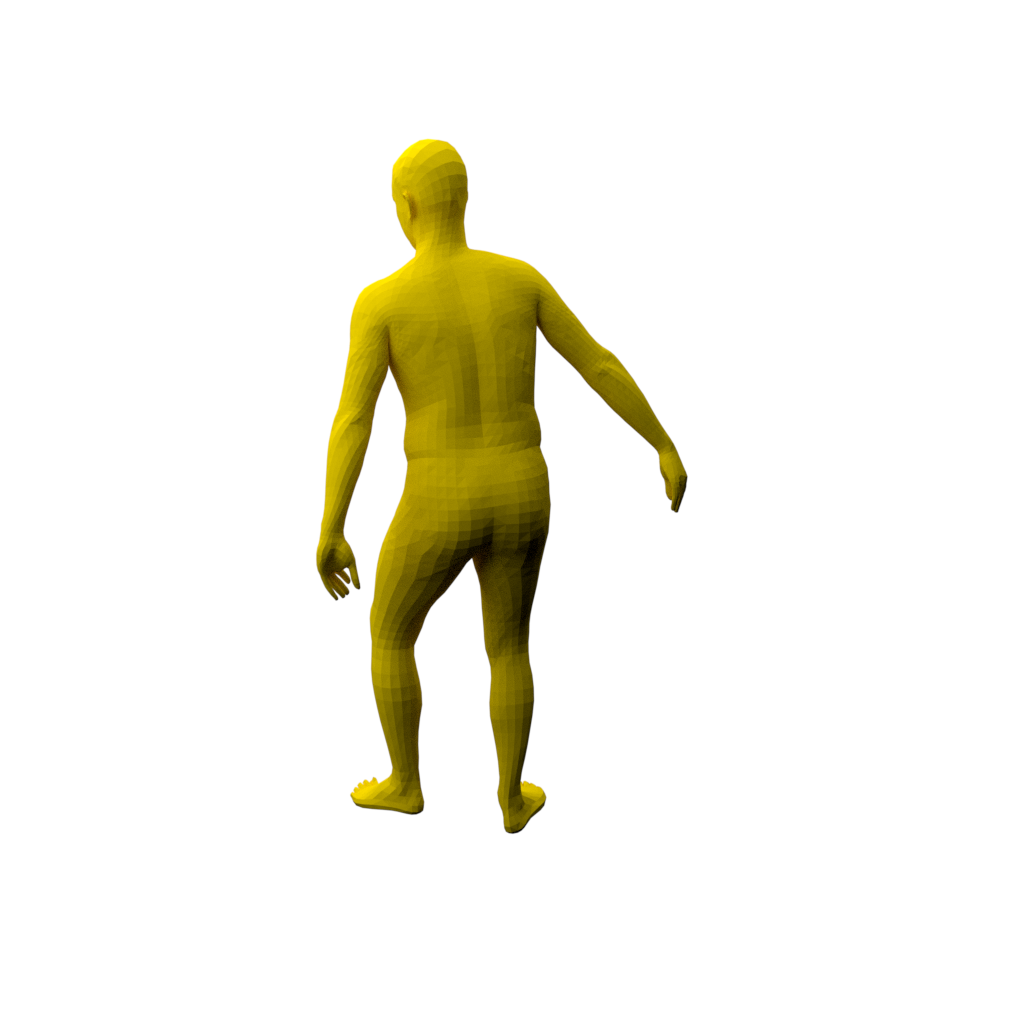}
    \caption{Ours SMPL}
 \end{subfigure}
\begin{subfigure}[b]{0.15\textwidth}
    \includegraphics [trim=10cm 7cm 10cm 4cm, width=0.95\textwidth]{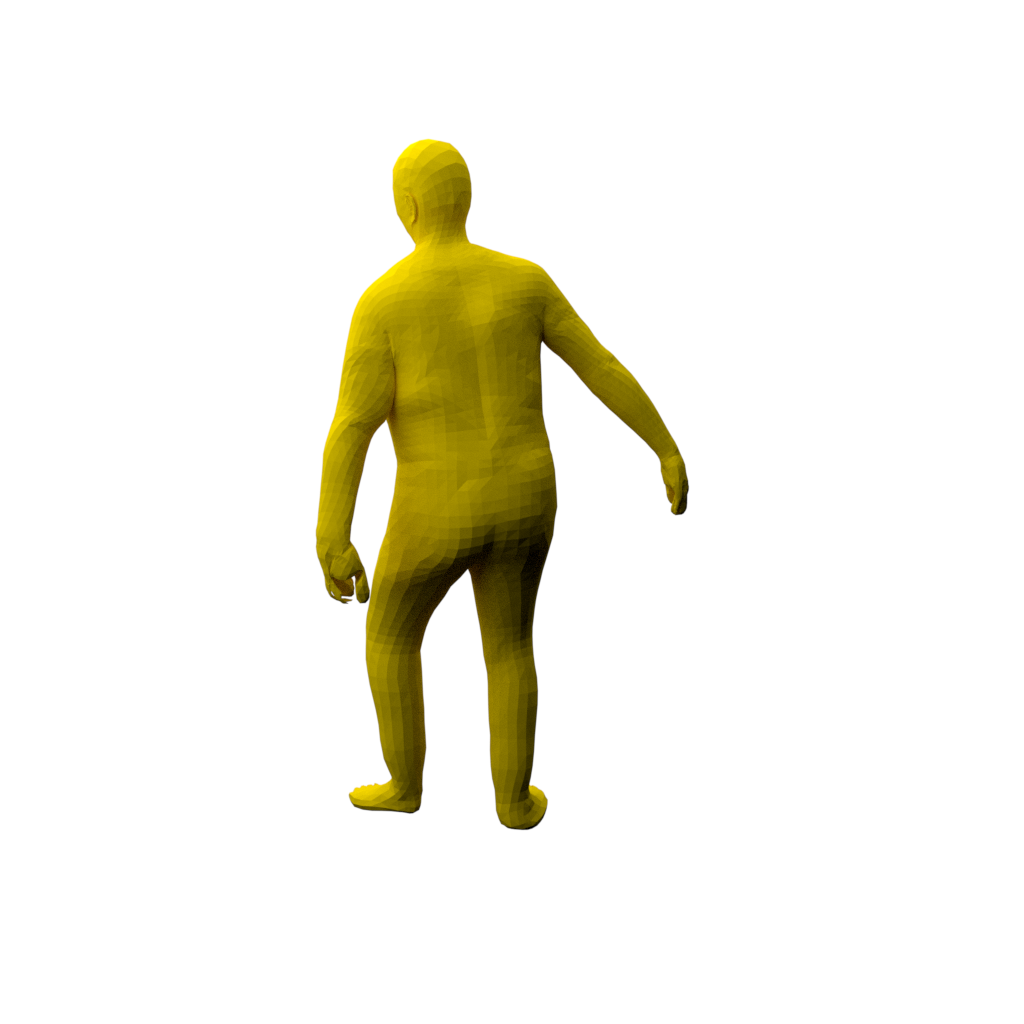}
    \caption{Ours SMPL+D}
 \end{subfigure}
\caption{One common failure case of IPNet~\cite{Bhatnagar_ECCV2020} happens when the global orientation deviates significantly from zero. This often results in catastrophic failures in the subsequent (b) SMPL fit and (c) SMPL+D fit.}
\label{fig:qualitative_results_CAPE1}
\end{figure*}
\begin{figure*}[t]
\centering
 \begin{subfigure}[b]{0.15\textwidth}
    \includegraphics [trim=10cm 7cm 10cm 4cm, width=0.95\textwidth]{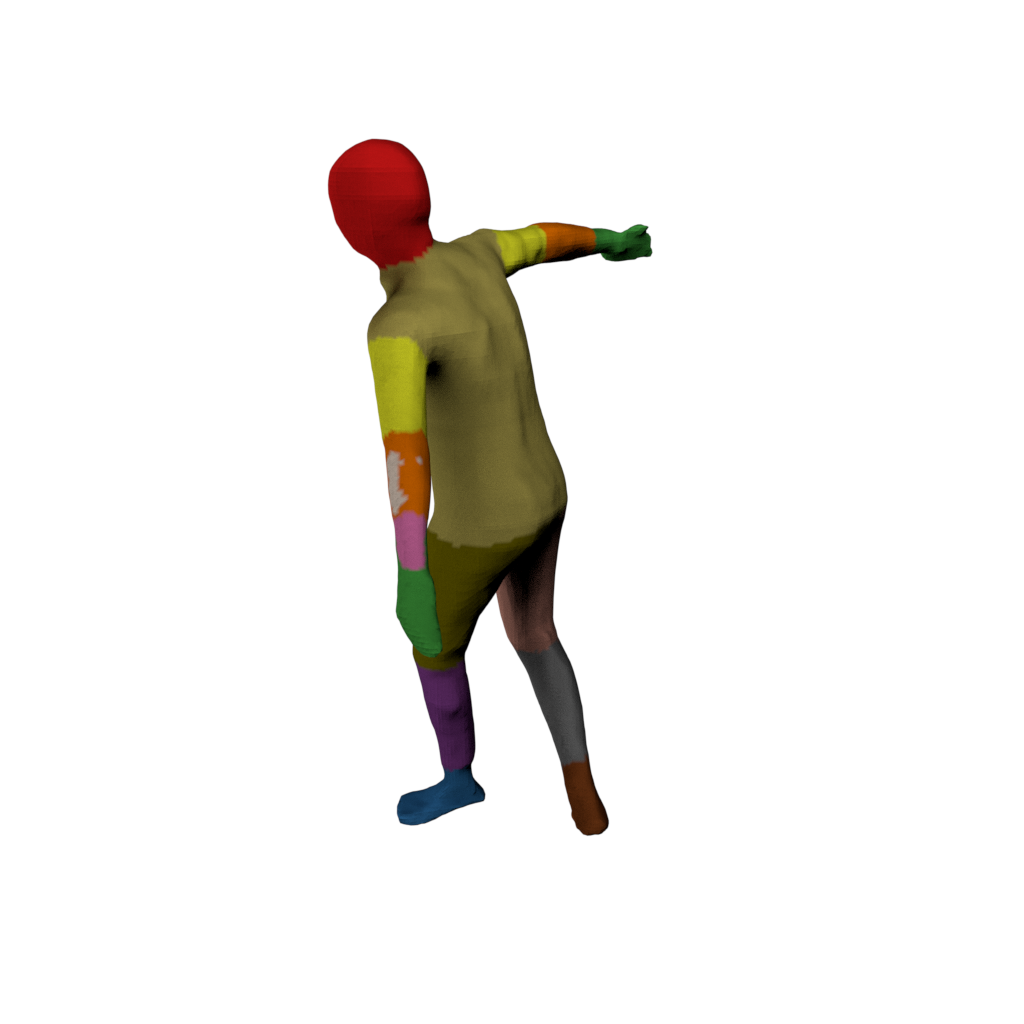}
    \caption{IPNet surface}
\end{subfigure}
 \begin{subfigure}[b]{0.15\textwidth}
    \includegraphics [trim=10cm 7cm 10cm 4cm, width=0.95\textwidth]{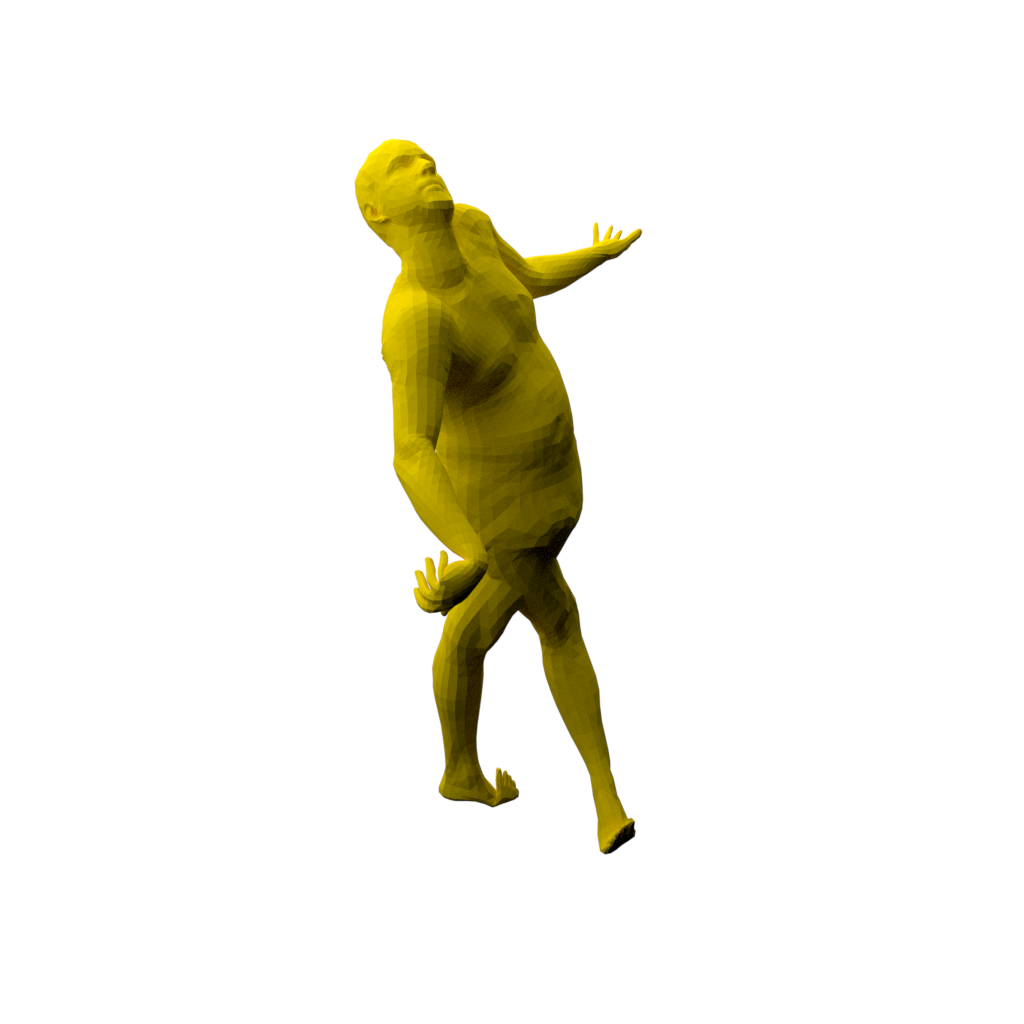}
    \caption{IPNet SMPL}
 \end{subfigure}
 \begin{subfigure}[b]{0.15\textwidth}
    \includegraphics [trim=10cm 7cm 10cm 4cm, width=0.95\textwidth]{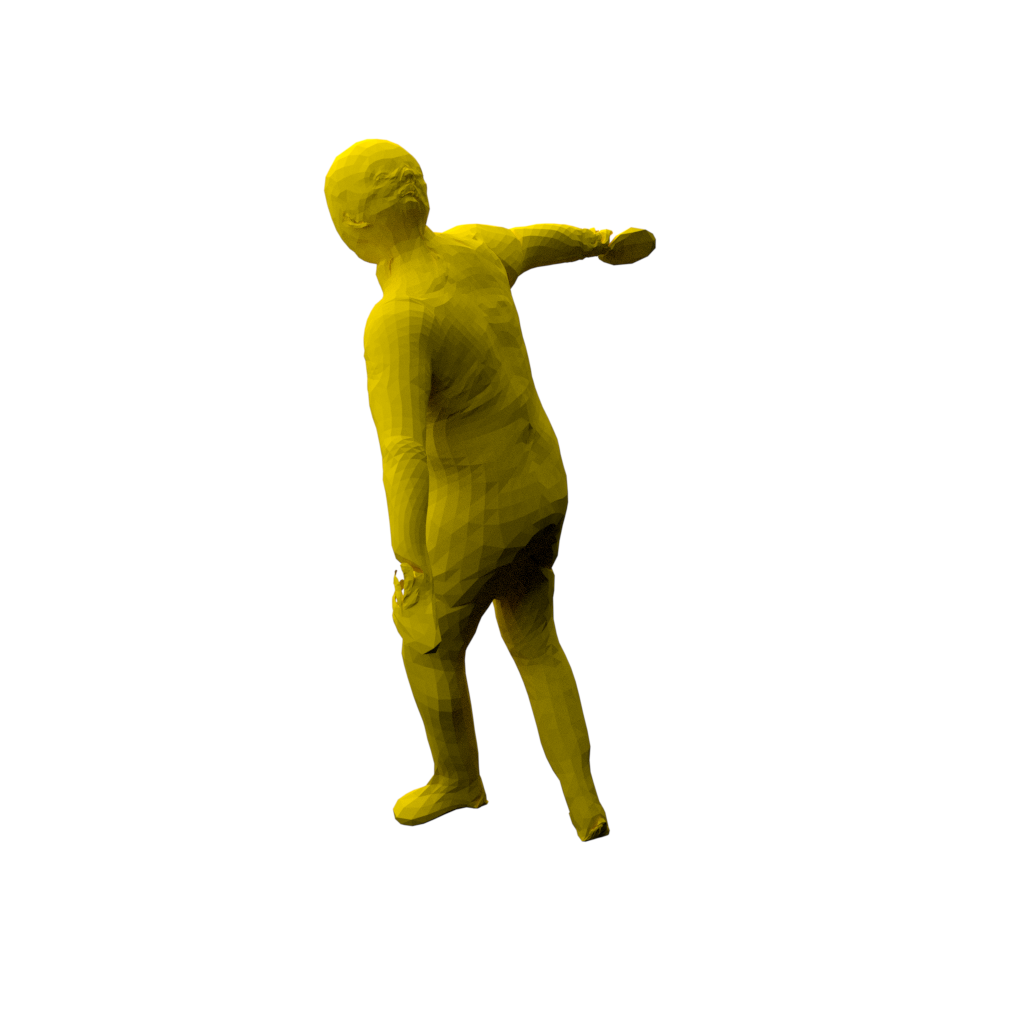}
    \caption{IPNet SMPL+D}
 \end{subfigure}
 \begin{subfigure}[b]{0.15\textwidth}
    \includegraphics [trim=10cm 7cm 10cm 4cm, width=0.95\textwidth]{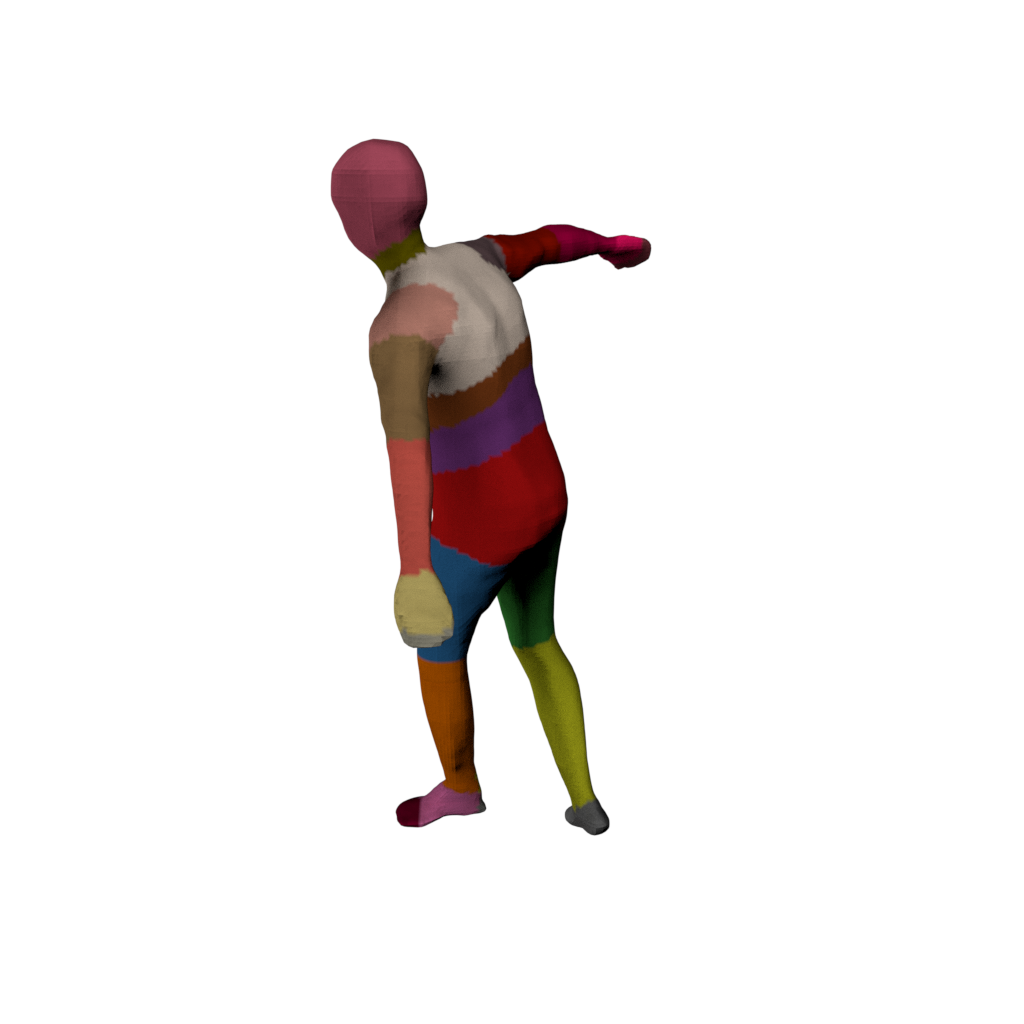}
    \caption{Ours surface}
 \end{subfigure}
\begin{subfigure}[b]{0.15\textwidth}
    \includegraphics [trim=10cm 7cm 10cm 4cm, width=0.95\textwidth]{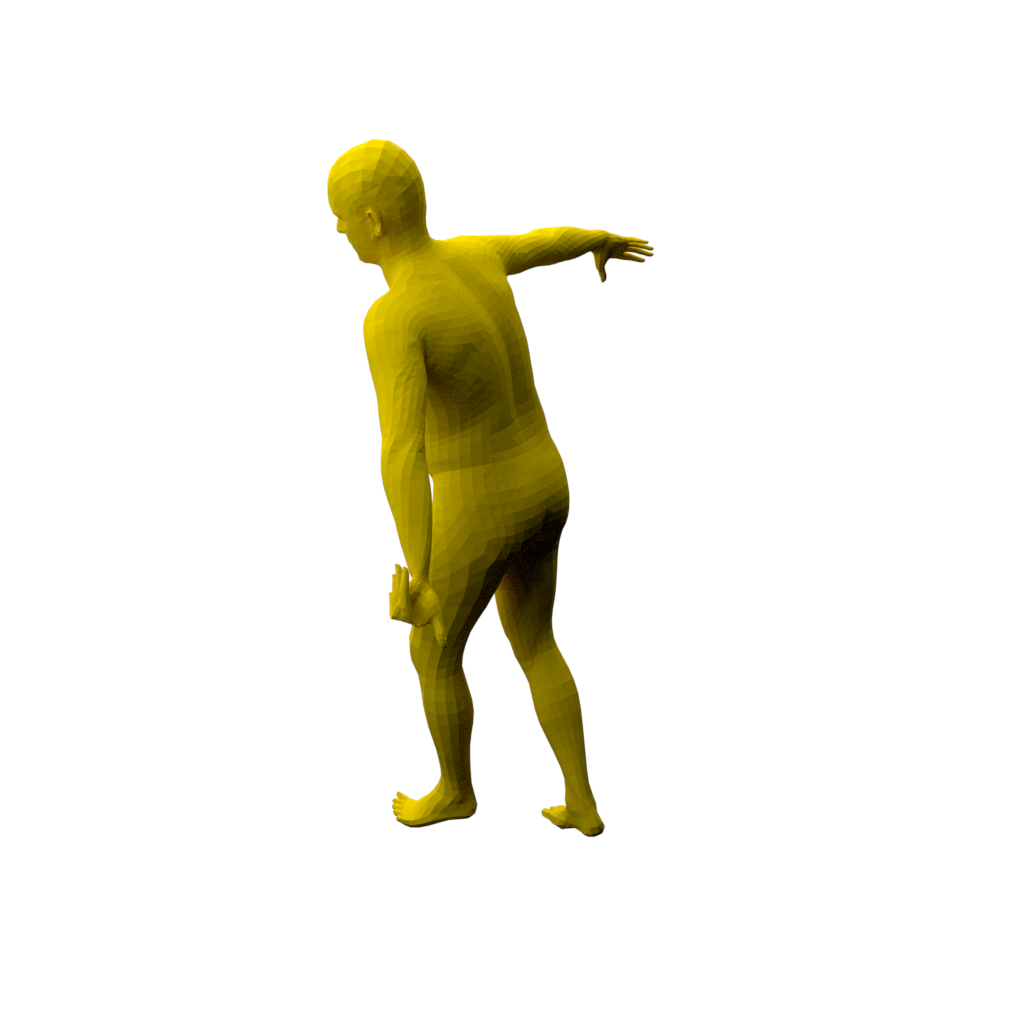}
    \caption{Ours SMPL}
 \end{subfigure}
\begin{subfigure}[b]{0.15\textwidth}
    \includegraphics [trim=10cm 7cm 10cm 4cm, width=0.95\textwidth]{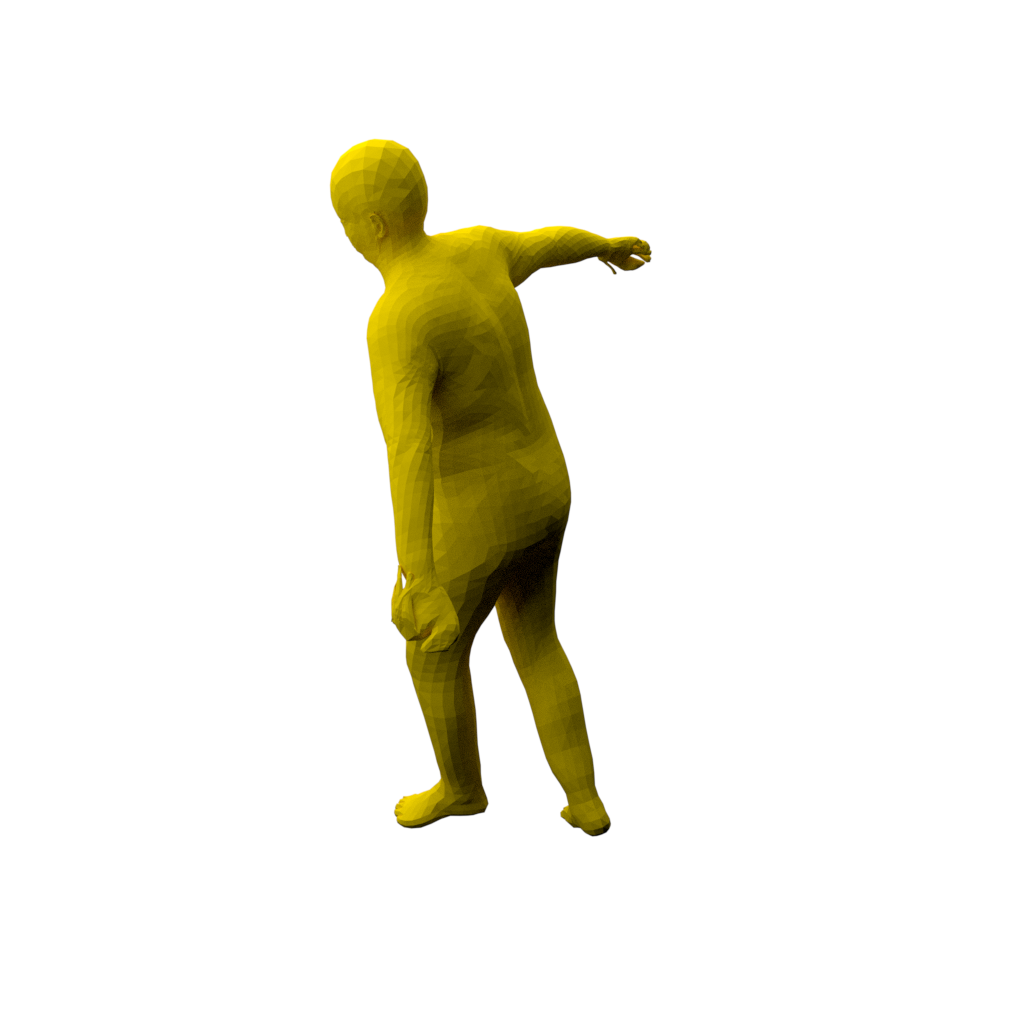}
    \caption{Ours SMPL+D}
 \end{subfigure}
\caption{Another type of failure cases of IPNet~\cite{Bhatnagar_ECCV2020} is caused by poor generalizability. As can be seen in (a), although IPNet has been trained using random augmentation to global orientation, it still struggles to generalize to the case in which the person's two arms are roughly aligned with the z-axis. IPNet misclassifies both arms of the person as the left-arm (same color), this results in catastrophic failure in the subsequent (b) SMPL fit and (c) SMPL+D fit. In (d) our model generalizes better and correctly distinguishes the left-arm and the right-arm under this rare pose, resulting in much more accurate registrations (e+f).}
\label{fig:qualitative_results_CAPE2}
\end{figure*}

We show additional qualitative results on the CAPE dataset in Fig.~\ref{fig:qualitative_results_CAPE1} and Fig.~\ref{fig:qualitative_results_CAPE2}. We observe that IPNet~\cite{Bhatnagar_ECCV2020} often fails when the global orientation deviates too much from zero. As can be seen in Fig.~\ref{fig:qualitative_results_CAPE1}, even if the local poses are relatively simple, IPNet still fails catastrophically; this is most likely due to the fact that, in order to make the optimization stable, the optimization objective often has regularization terms that penalize poses which deviate too much from the mean-pose. This choice is statistically meaningful, but without pose initialization, it will also make it impossible for the optimizer to converge to rare poses that are realistic but deviate too much from the mean-pose. 

Another type of failure cases of IPNet is caused by its network's poor generalizability. Fig.~\ref{fig:qualitative_results_CAPE2} shows that IPNet fails to correctly distinguish between the left-arm and the right-arm under a rare pose. This is because IPNet learns the occupancy functions of the two arms in posed space, and thus these occupancy functions need to memorize all possible locations of arms in posed space. On the other hand, our model learns to canonicalize points before the occupancy classification, thus the occupancy functions of arms only need to memorize a small region in rest-pose space. This results in better generalizability and our model correctly distinguishes the left-arm and the left-arm in this case.

%% file: arxiv_qualitative_buff.tex
\section{Additional Qualitative Results on the BUFF Dataset}
\label{appx:BUFF_qualitative}
\begin{figure*}[t]
\centering
 \begin{subfigure}[b]{0.15\textwidth}
    \includegraphics [trim=9cm 7cm 9cm 5cm, width=0.95\textwidth]{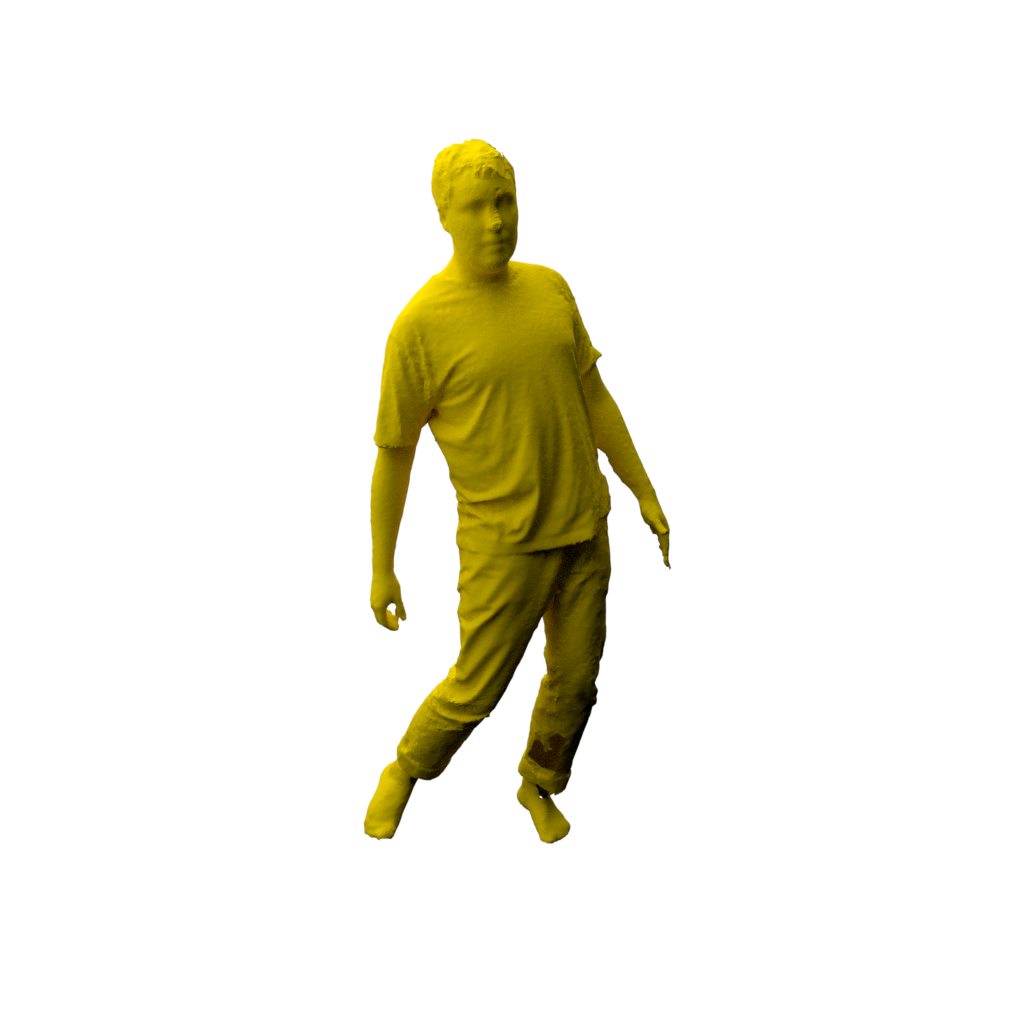}
\end{subfigure}
 \begin{subfigure}[b]{0.15\textwidth}
    \includegraphics [trim=9cm 7cm 9cm 5cm, width=0.95\textwidth]{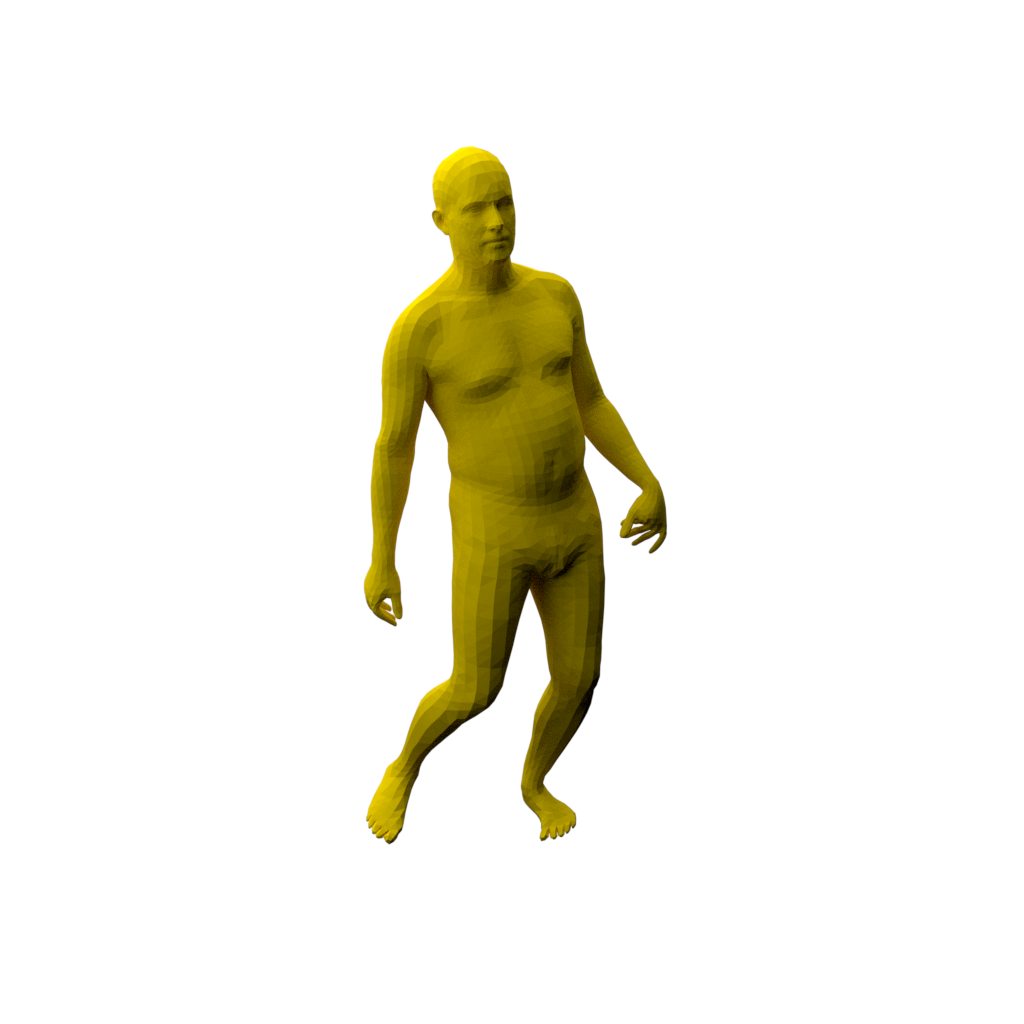}
 \end{subfigure}
 \begin{subfigure}[b]{0.15\textwidth}
    \includegraphics [trim=9cm 7cm 9cm 5cm, width=0.95\textwidth]{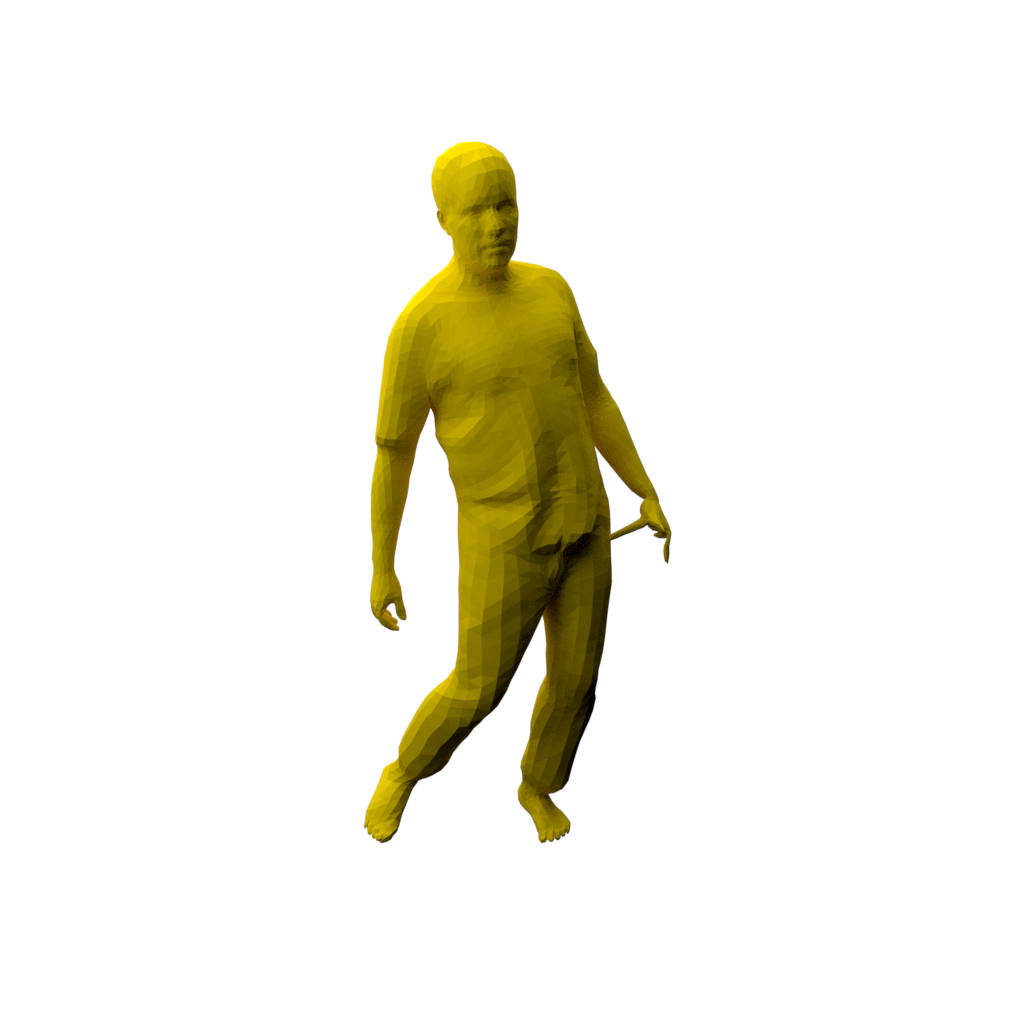}
 \end{subfigure}
 \begin{subfigure}[b]{0.15\textwidth}
    \includegraphics [trim=9cm 7cm 9cm 5cm, width=0.95\textwidth]{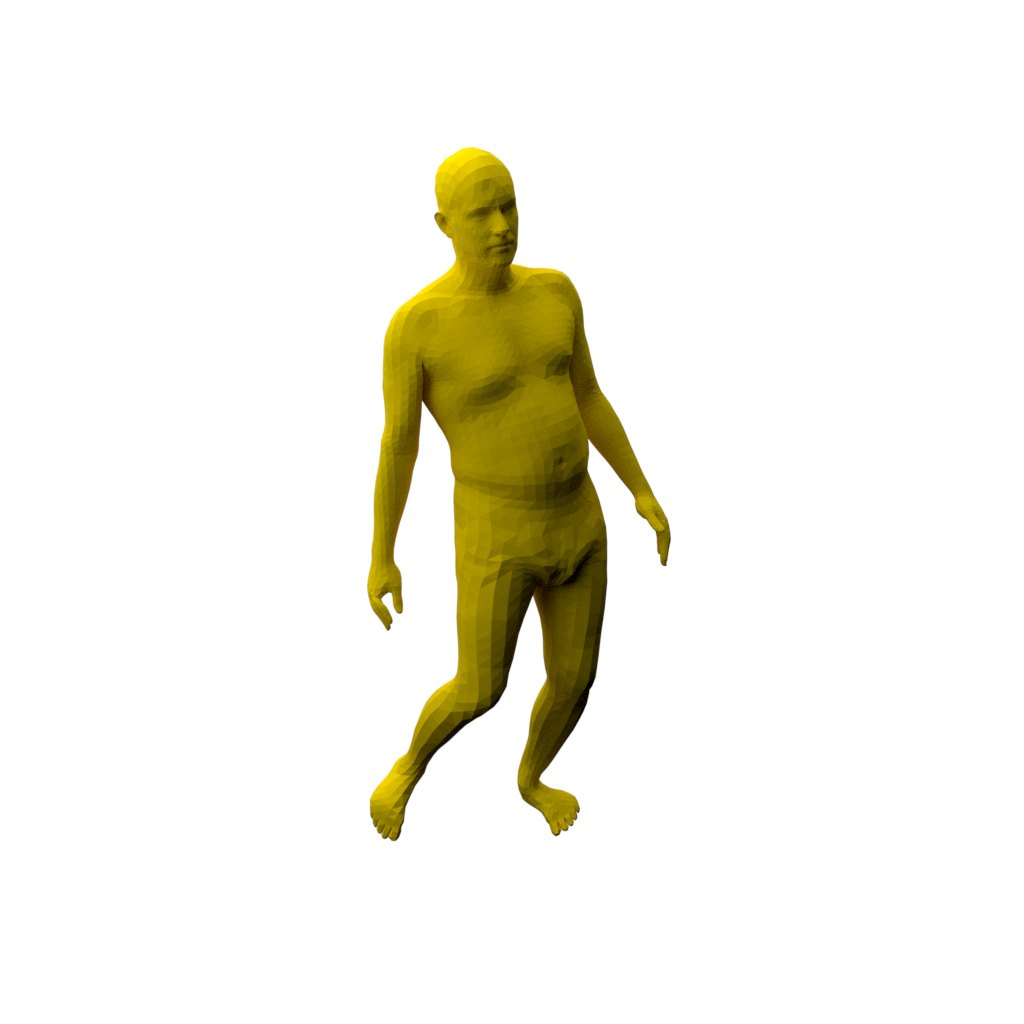}
 \end{subfigure}
 \begin{subfigure}[b]{0.15\textwidth}
    \includegraphics [trim=9cm 7cm 9cm 5cm, width=0.95\textwidth]{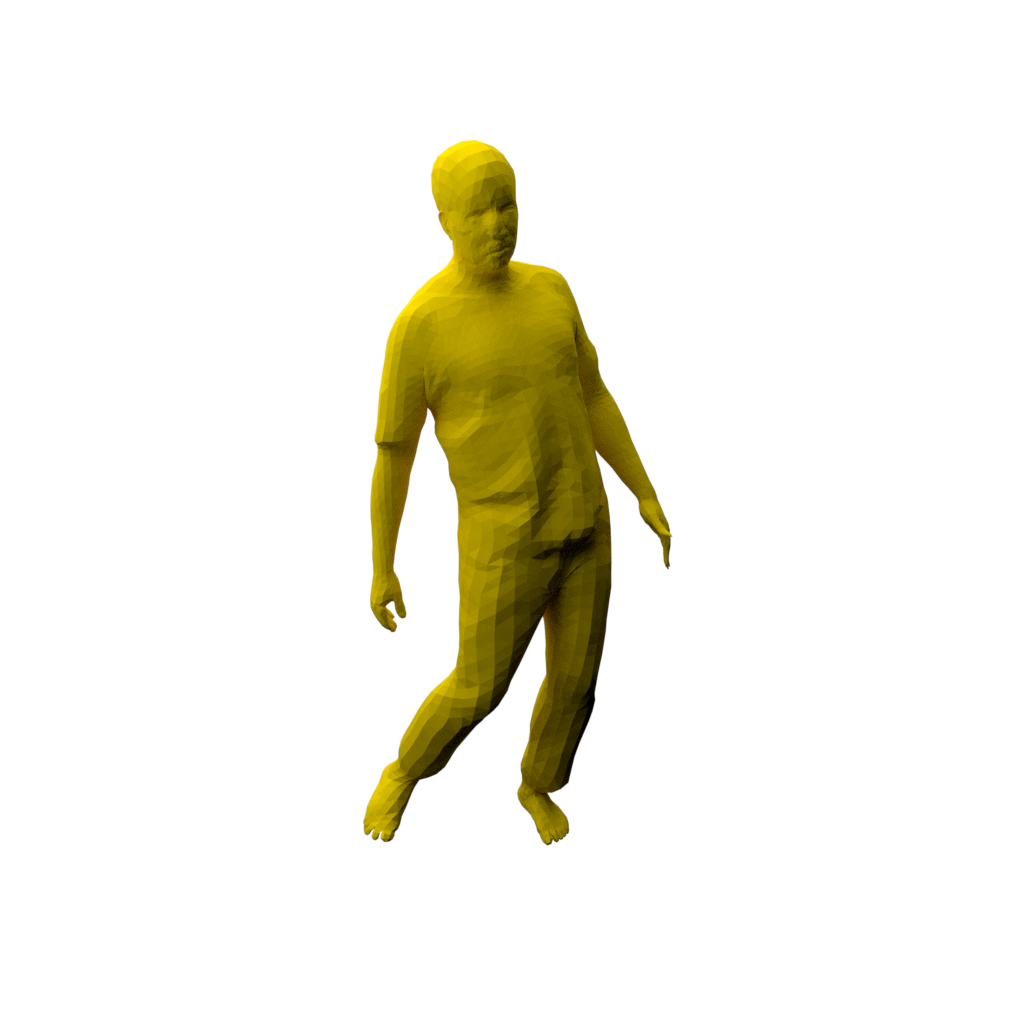}
 \end{subfigure}
 \begin{subfigure}[b]{0.15\textwidth}
    \includegraphics [trim=9cm 7cm 9cm 5cm, width=0.95\textwidth]{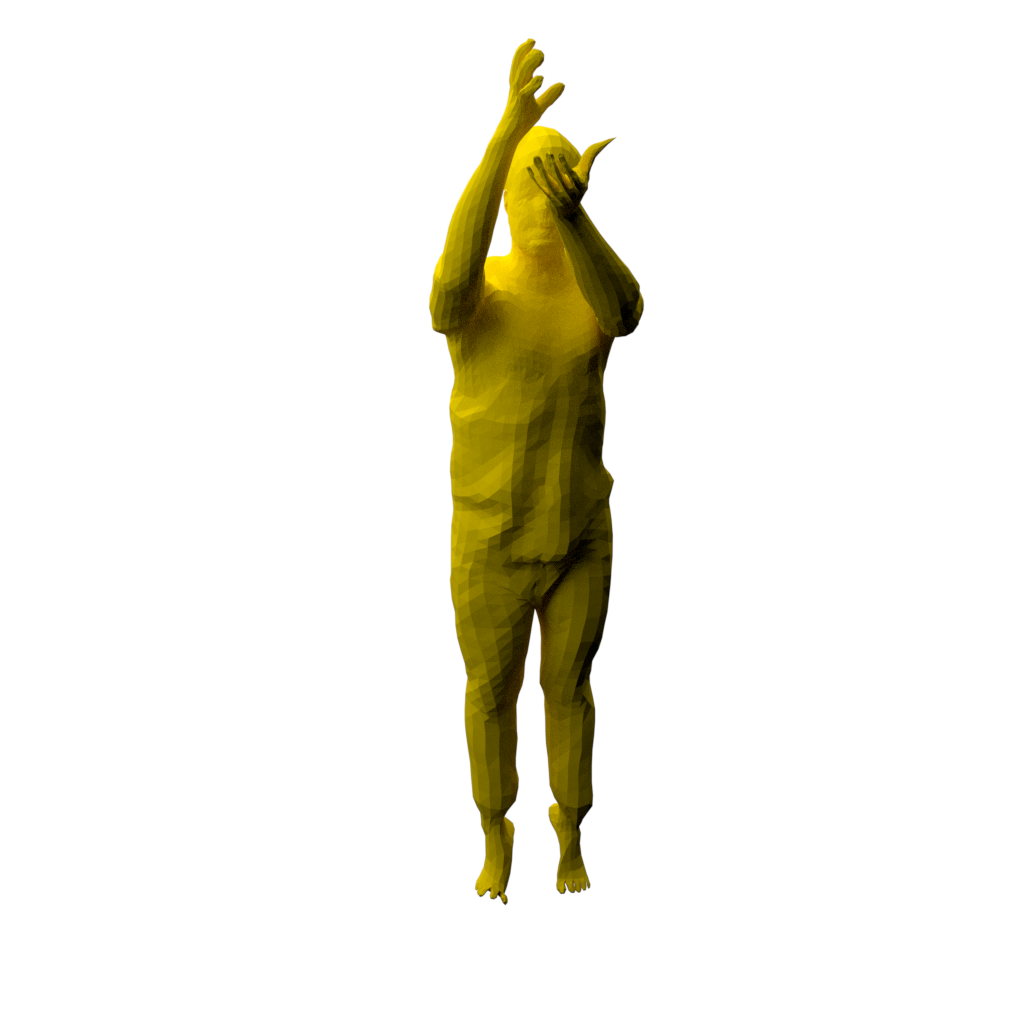}
 \end{subfigure} \\
 \begin{subfigure}[b]{0.15\textwidth}
    \includegraphics [trim=9cm 6cm 9cm 4cm, width=0.95\textwidth]{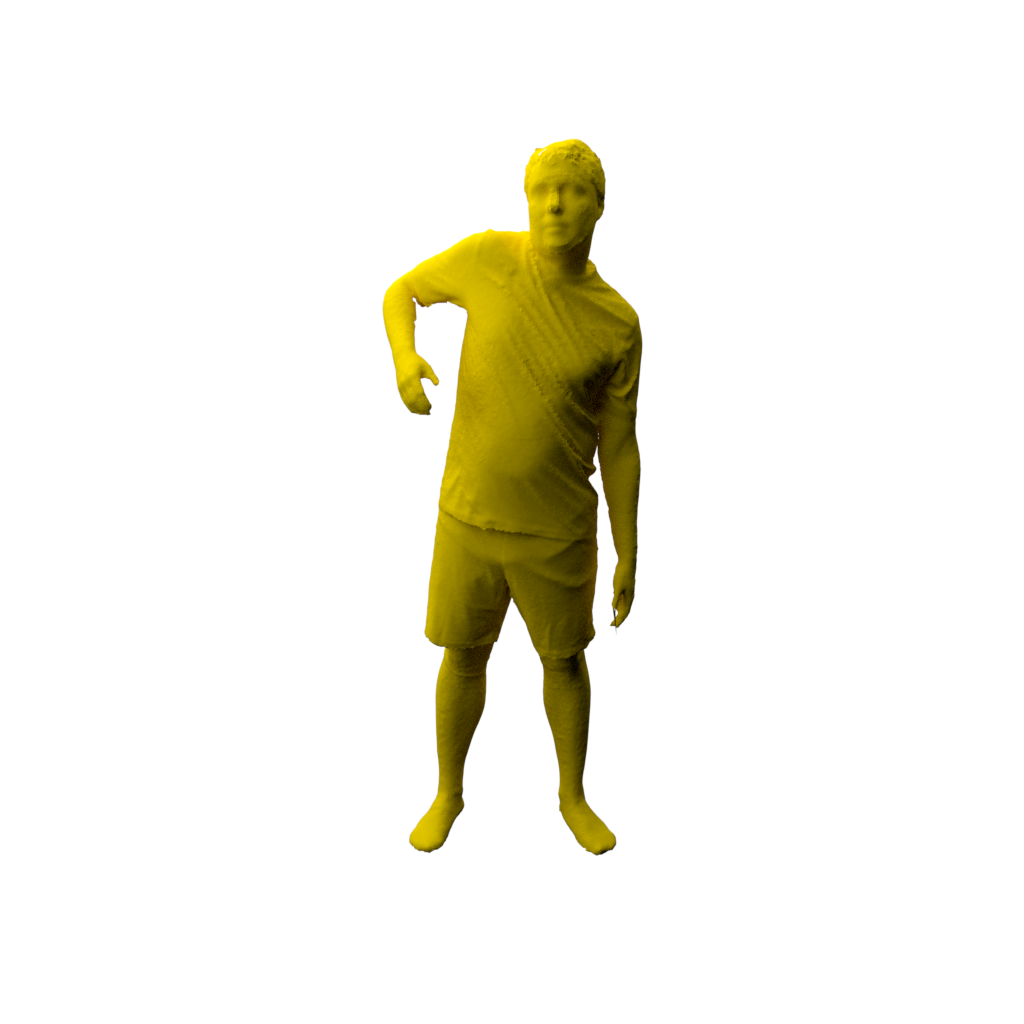}
\end{subfigure}
 \begin{subfigure}[b]{0.15\textwidth}
    \includegraphics [trim=9cm 6cm 9cm 4cm, width=0.95\textwidth]{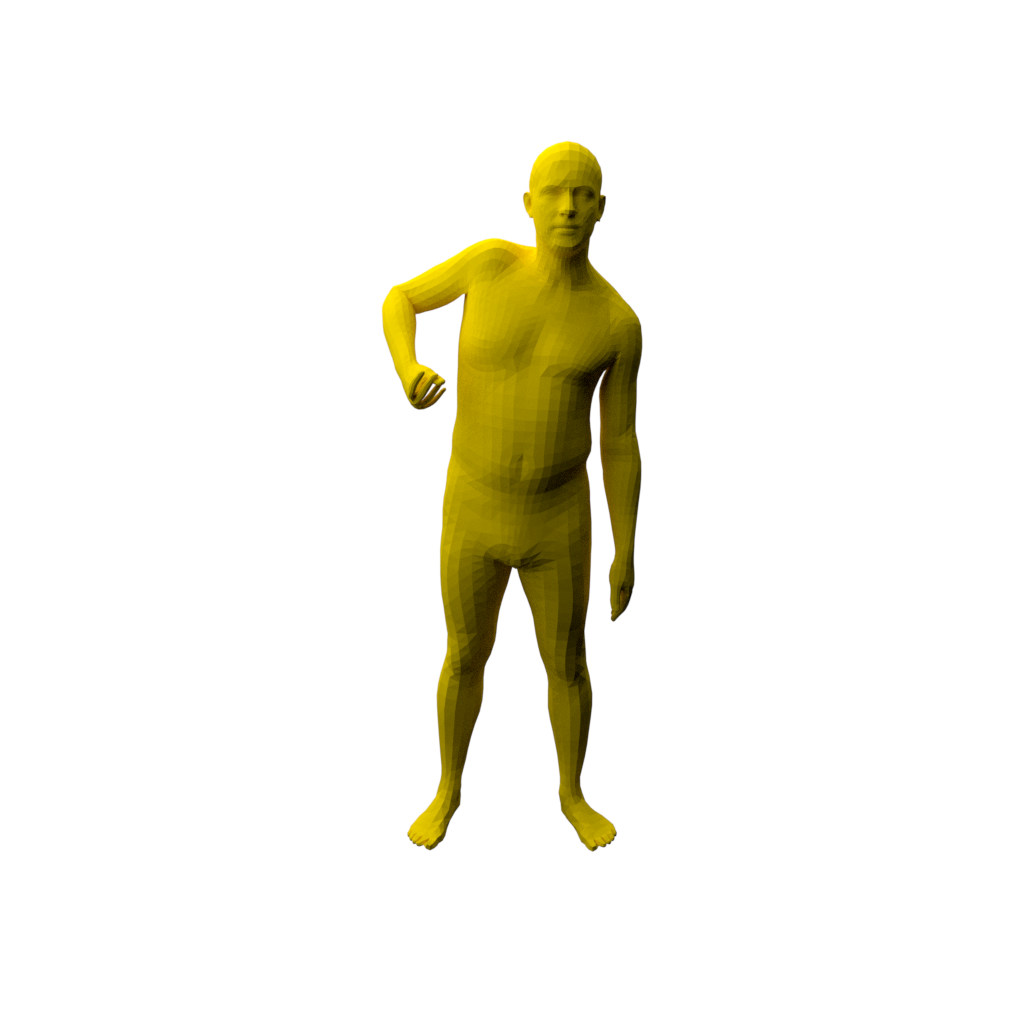}
 \end{subfigure}
 \begin{subfigure}[b]{0.15\textwidth}
    \includegraphics [trim=9cm 6cm 9cm 4cm, width=0.95\textwidth]{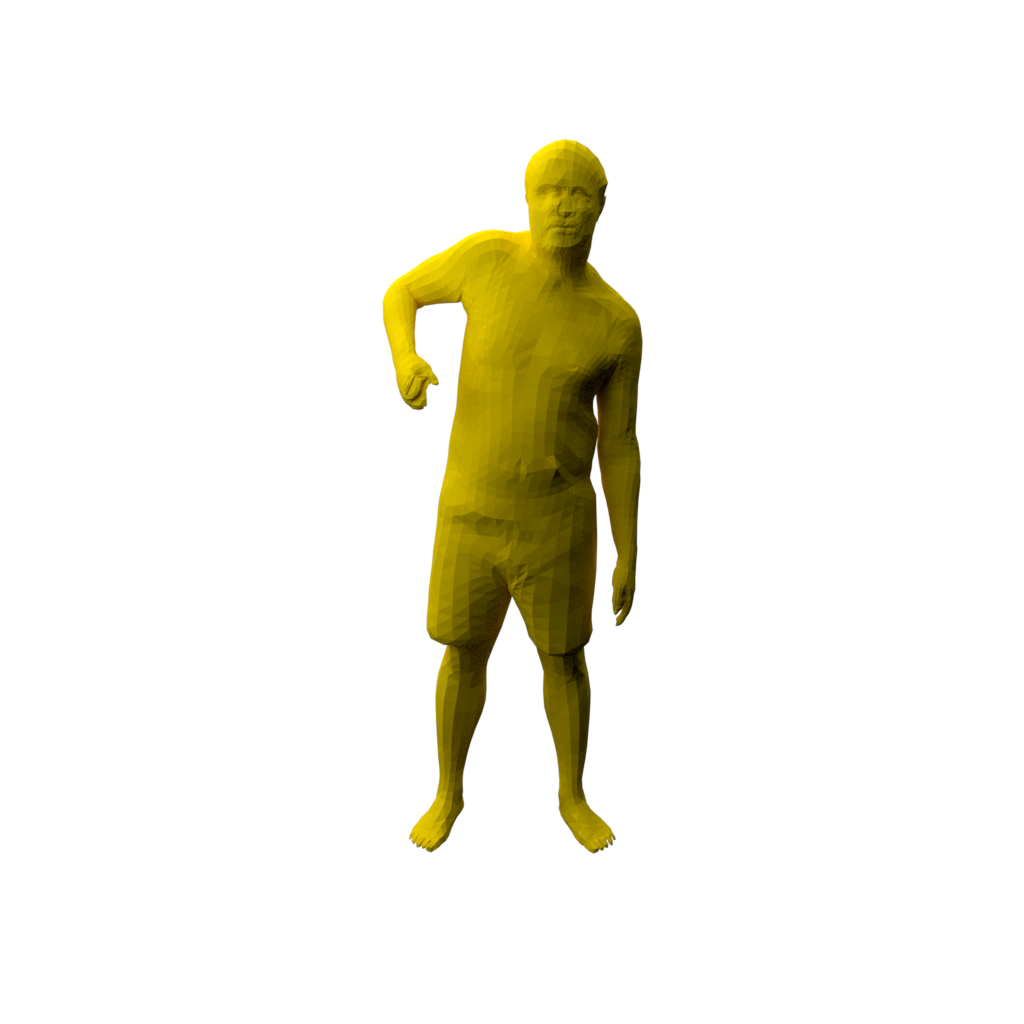}
 \end{subfigure}
 \begin{subfigure}[b]{0.15\textwidth}
    \includegraphics [trim=9cm 6cm 9cm 4cm, width=0.95\textwidth]{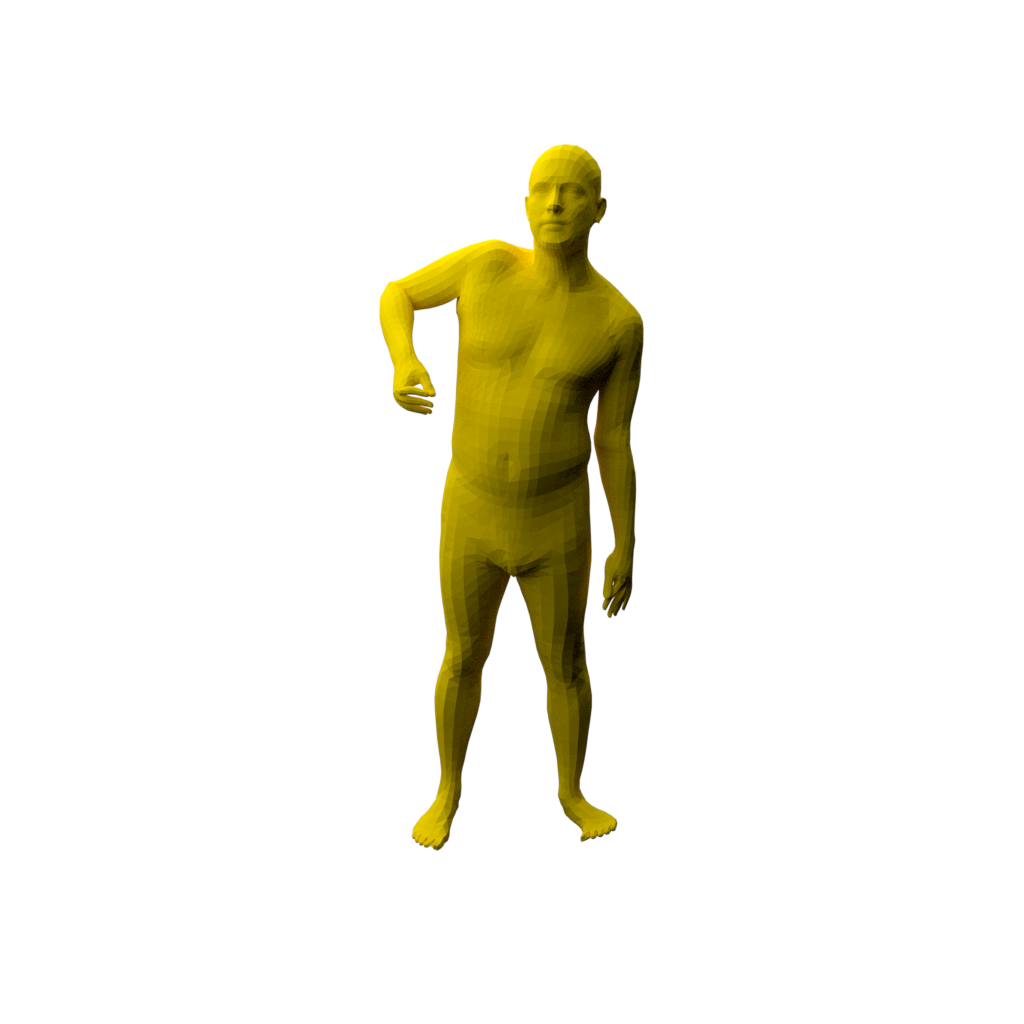}
 \end{subfigure}
 \begin{subfigure}[b]{0.15\textwidth}
    \includegraphics [trim=9cm 6cm 9cm 4cm, width=0.95\textwidth]{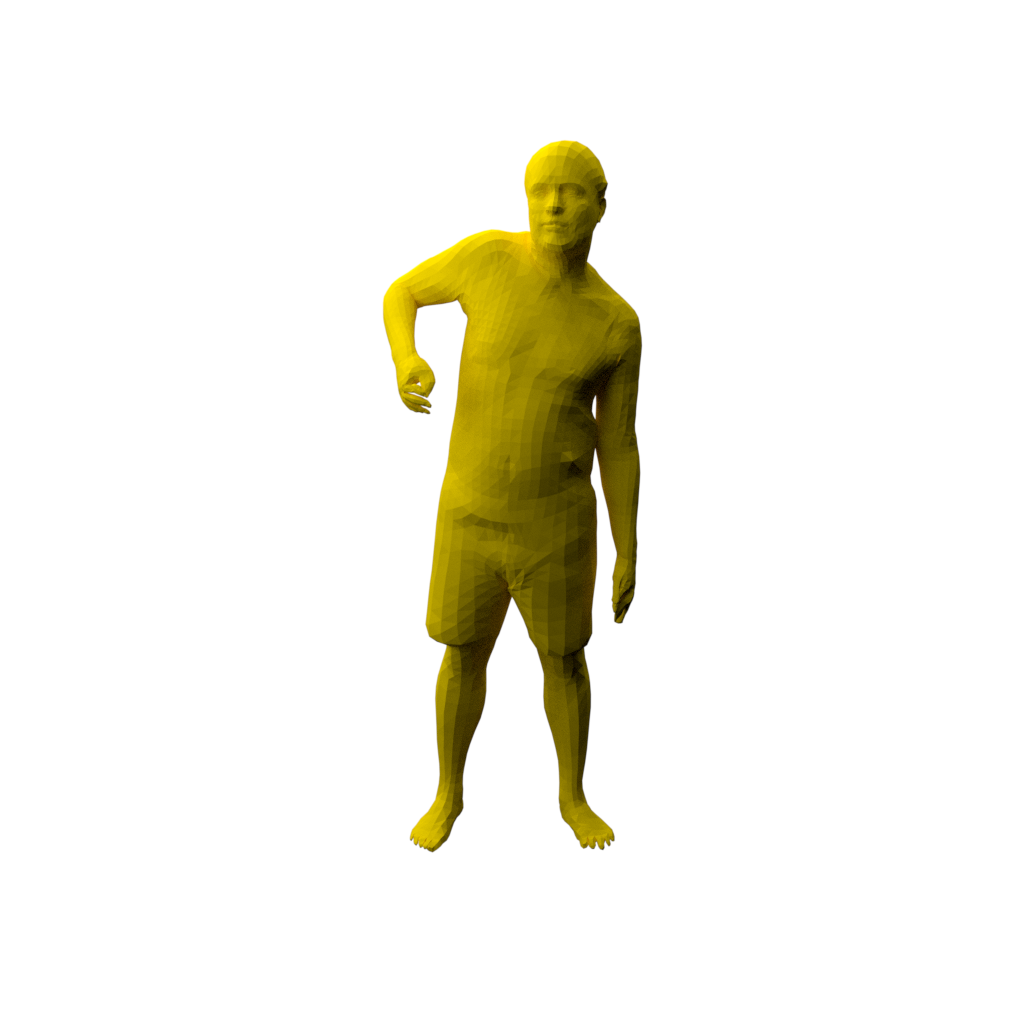}
 \end{subfigure}
 \begin{subfigure}[b]{0.15\textwidth}
    \includegraphics [trim=9cm 6cm 9cm 2cm, width=0.95\textwidth]{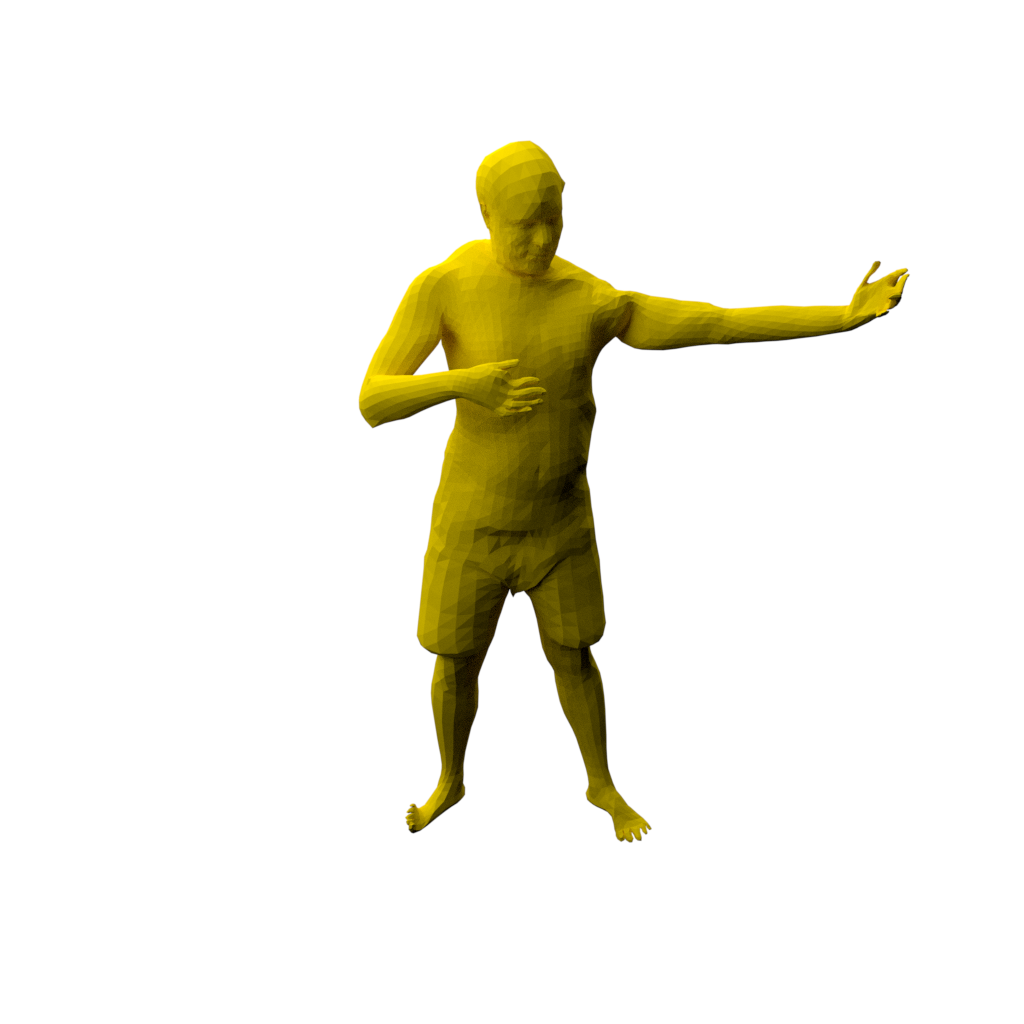}
 \end{subfigure} \\
 \begin{subfigure}[b]{0.15\textwidth}
    \includegraphics [trim=7cm 5cm 8cm 4.5cm, width=0.95\textwidth]{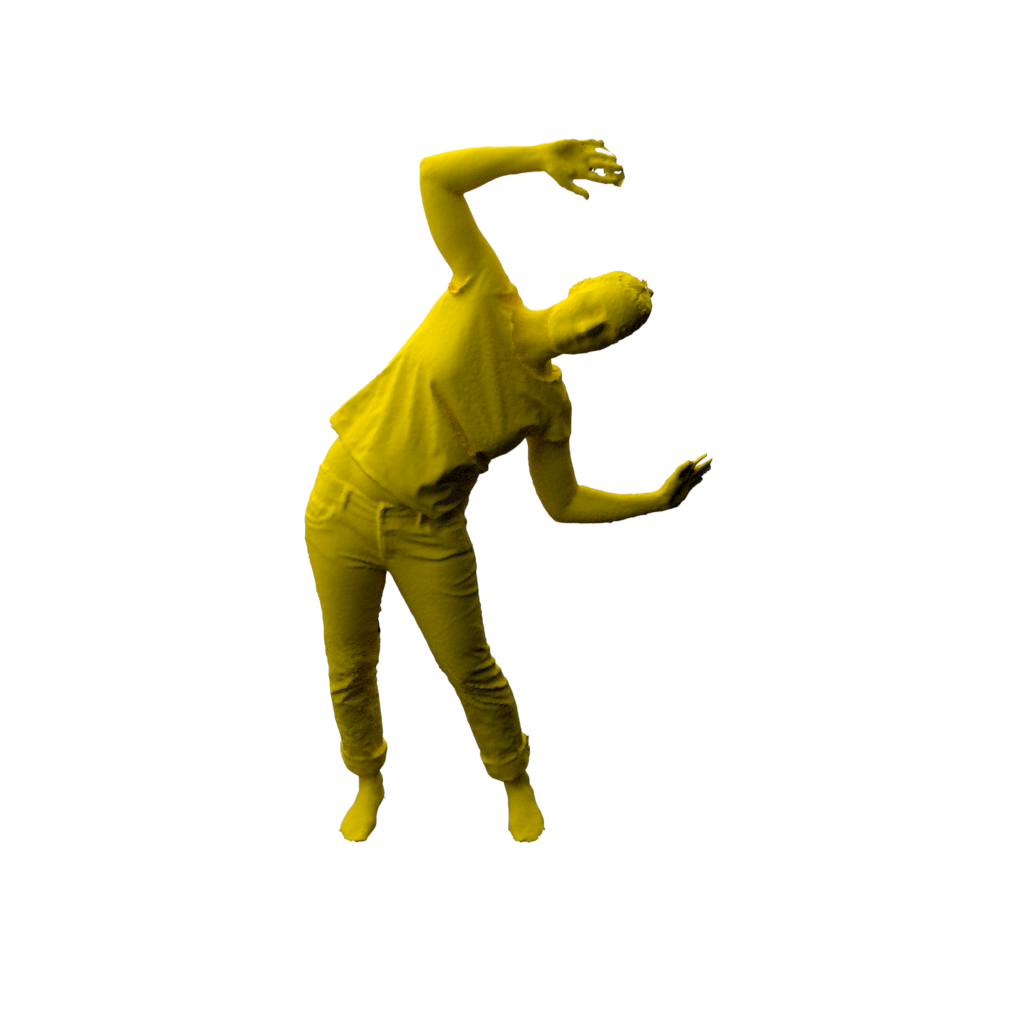}
    \caption{Raw scan}
\end{subfigure}
 \begin{subfigure}[b]{0.15\textwidth}
    \includegraphics [trim=7cm 5cm 8cm 4.5cm, width=0.95\textwidth]{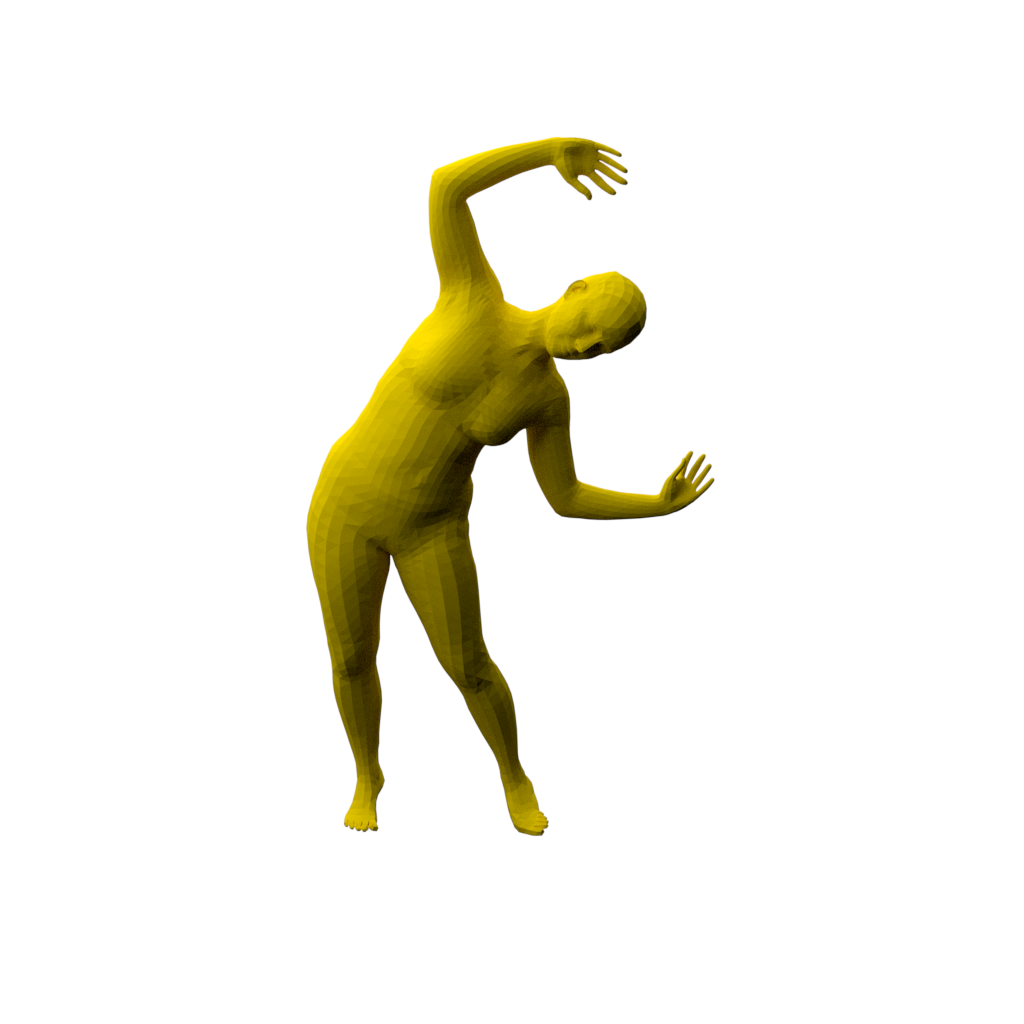}
    \caption{IPNet SMPL}
 \end{subfigure}
 \begin{subfigure}[b]{0.15\textwidth}
    \includegraphics [trim=7cm 5cm 8cm 4.5cm, width=0.95\textwidth]{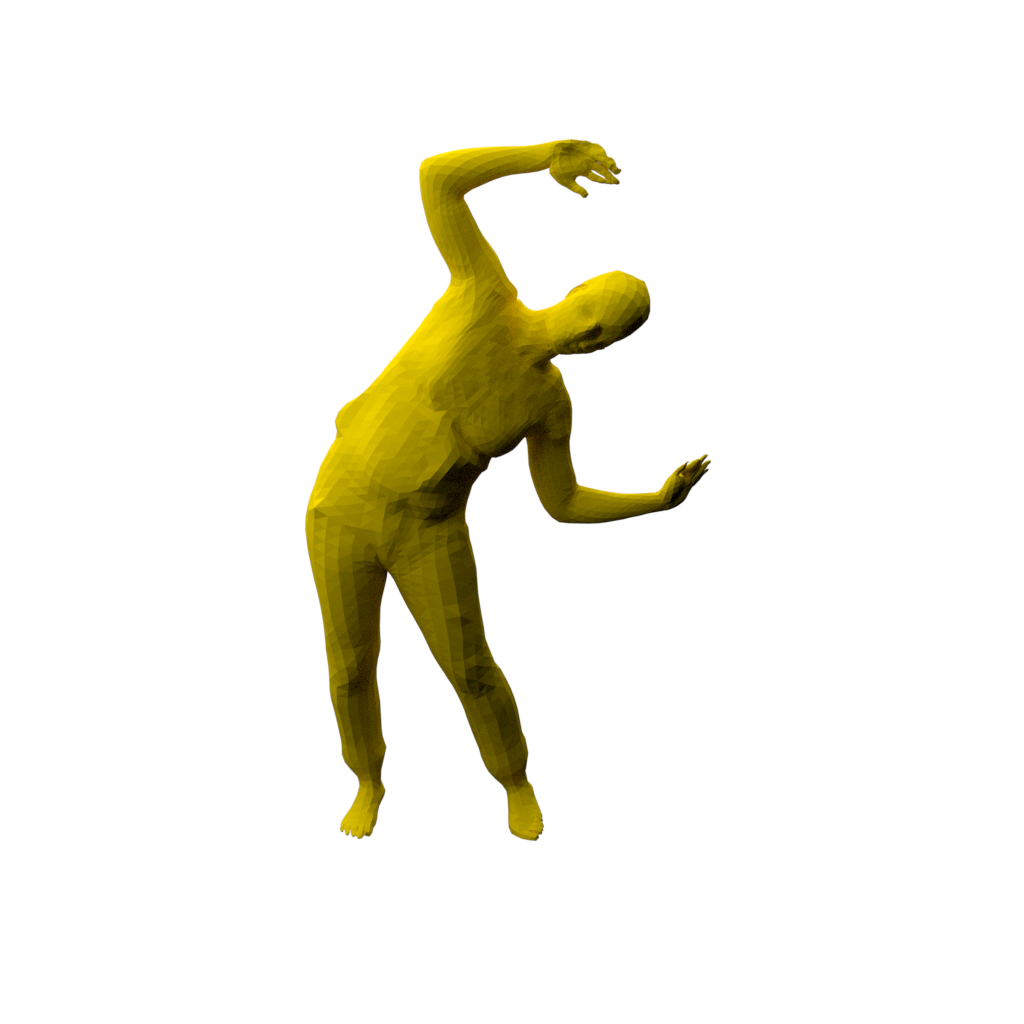}
    \caption{IPNet SMPL+D}
 \end{subfigure}
 \begin{subfigure}[b]{0.15\textwidth}
    \includegraphics [trim=7cm 5cm 8cm 4.5cm, width=0.95\textwidth]{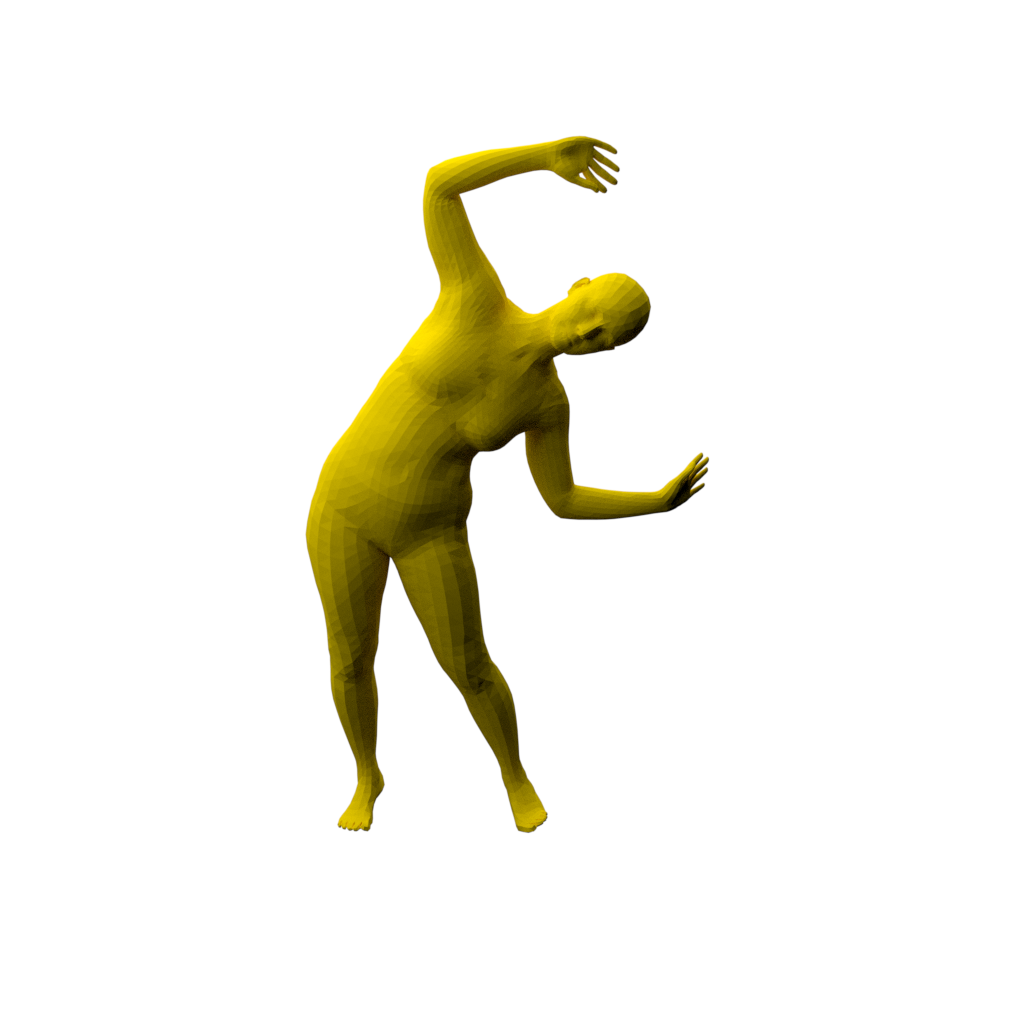}
    \caption{Ours SMPL}
 \end{subfigure}
 \begin{subfigure}[b]{0.15\textwidth}
    \includegraphics [trim=7cm 5cm 8cm 4.5cm, width=0.95\textwidth]{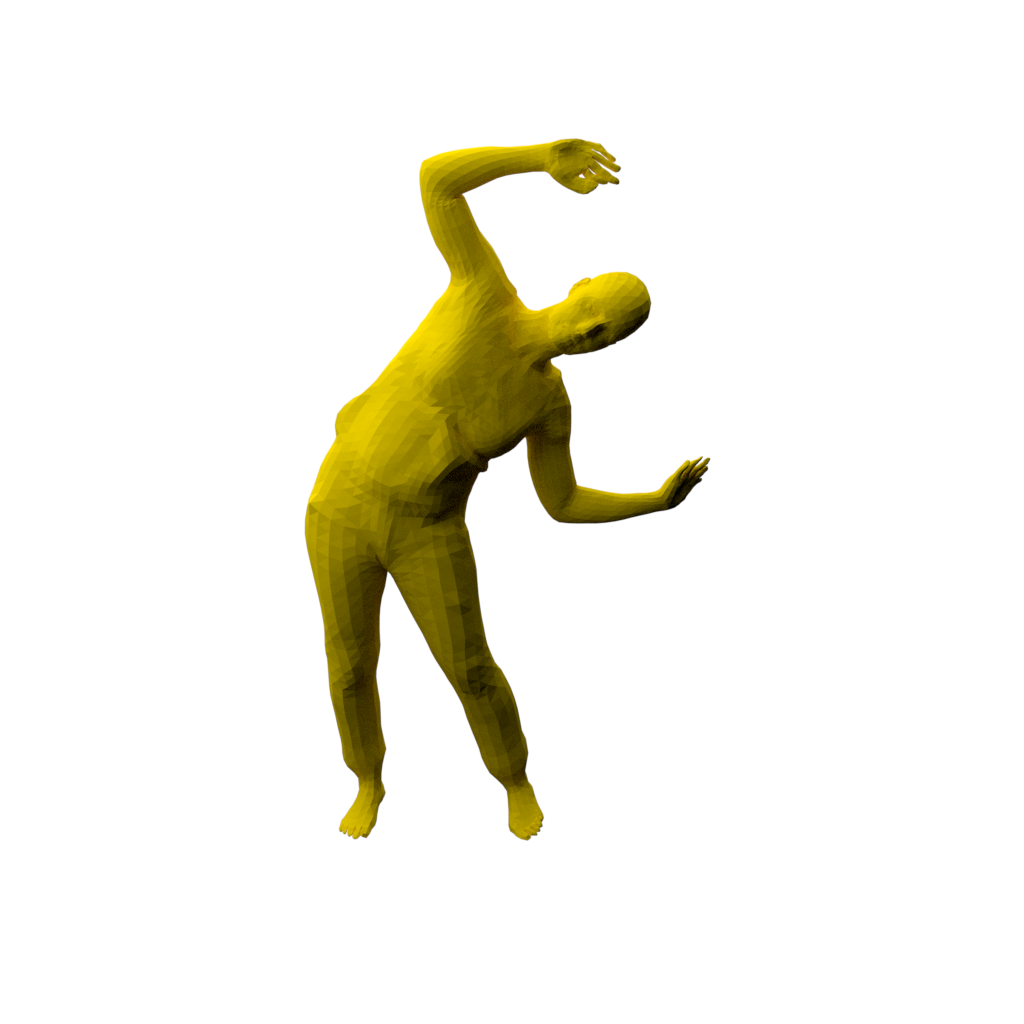}
    \caption{Ours SMPL+D}
 \end{subfigure}
 \begin{subfigure}[b]{0.15\textwidth}
    \includegraphics [trim=7cm 5cm 8cm 4.5cm, width=0.95\textwidth]{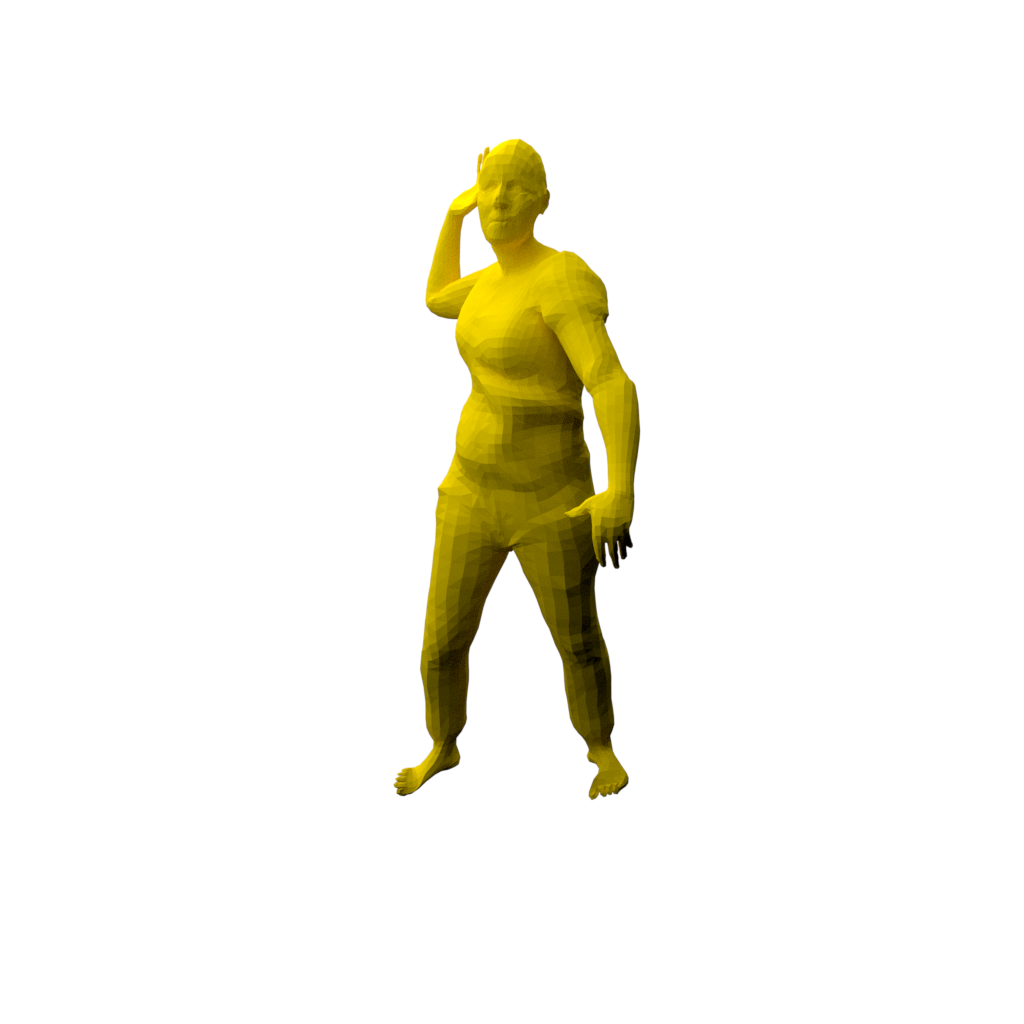}
    \caption{Ours reposed}
 \end{subfigure}
\caption{More qualitative results on the BUFF dataset. Note that our SMPLD/SMPL+D fits perform consistently better than IPNet~\cite{Bhatnagar_ECCV2020}, especially around faces and hands}
\label{fig:qualitative_results_BUFF}
\end{figure*}

We present more qualitative results on the BUFF dataset in Fig.~\ref{fig:qualitative_results_BUFF}. Note that the BUFF dataset \textit{does not} contain ground-truth pose parameters, thus it is not possible to evaluate quantitatively on registration errors as we did on the CAPE dataset. These qualitative results are meant to demonstrate generalization performance of our trained model to real scans, even though the model is only trained on synthetically sampled point clouds from registered dressed people.

%

%% file: arxiv_limitations.tex
\section{Limitations}
\label{appx:limitations}
\begin{figure*}[t]
\captionsetup[subfigure]{labelformat=empty}
\centering
 \begin{subfigure}[b]{0.15\textwidth}
    \includegraphics [trim=10cm 7cm 10cm 5cm, width=0.95\textwidth]{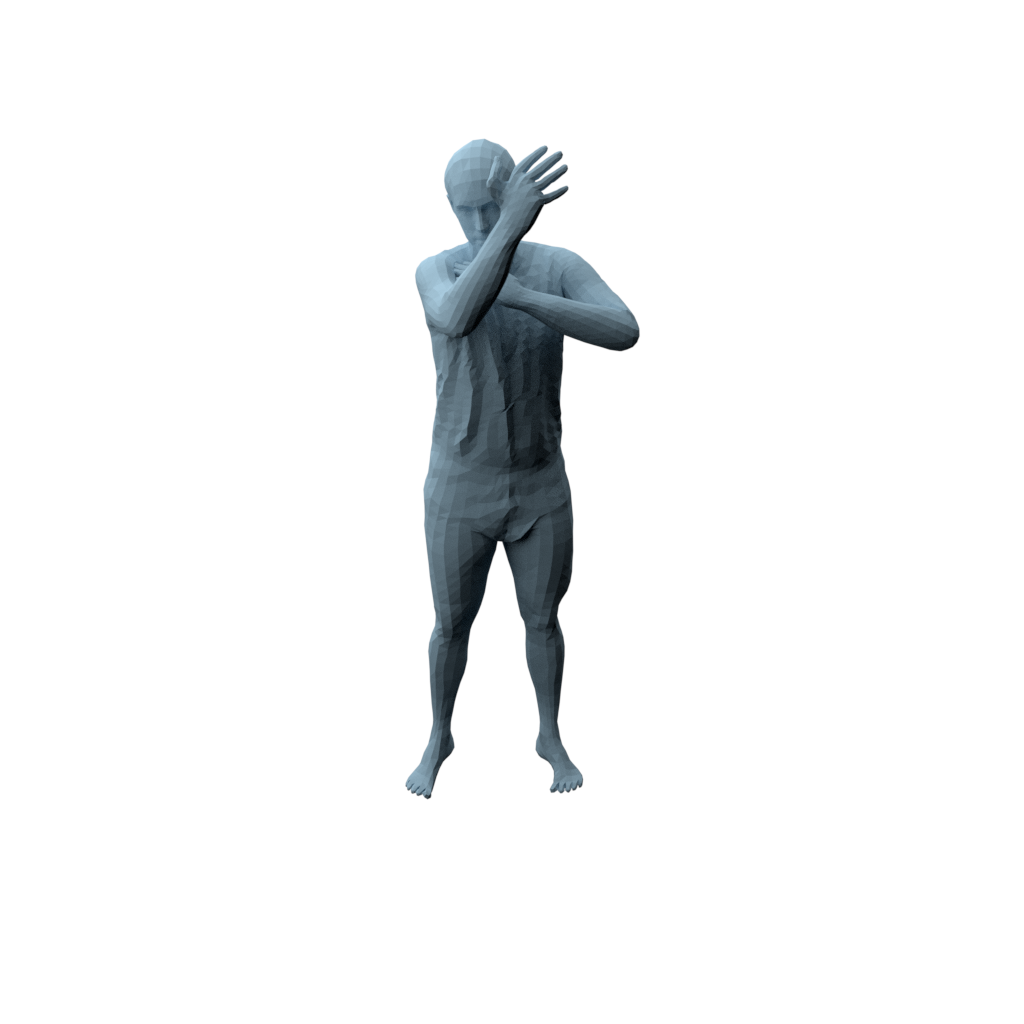}
    \caption{GT}
\end{subfigure}
 \begin{subfigure}[b]{0.15\textwidth}
    \includegraphics [trim=10cm 7cm 10cm 5cm, width=0.95\textwidth]{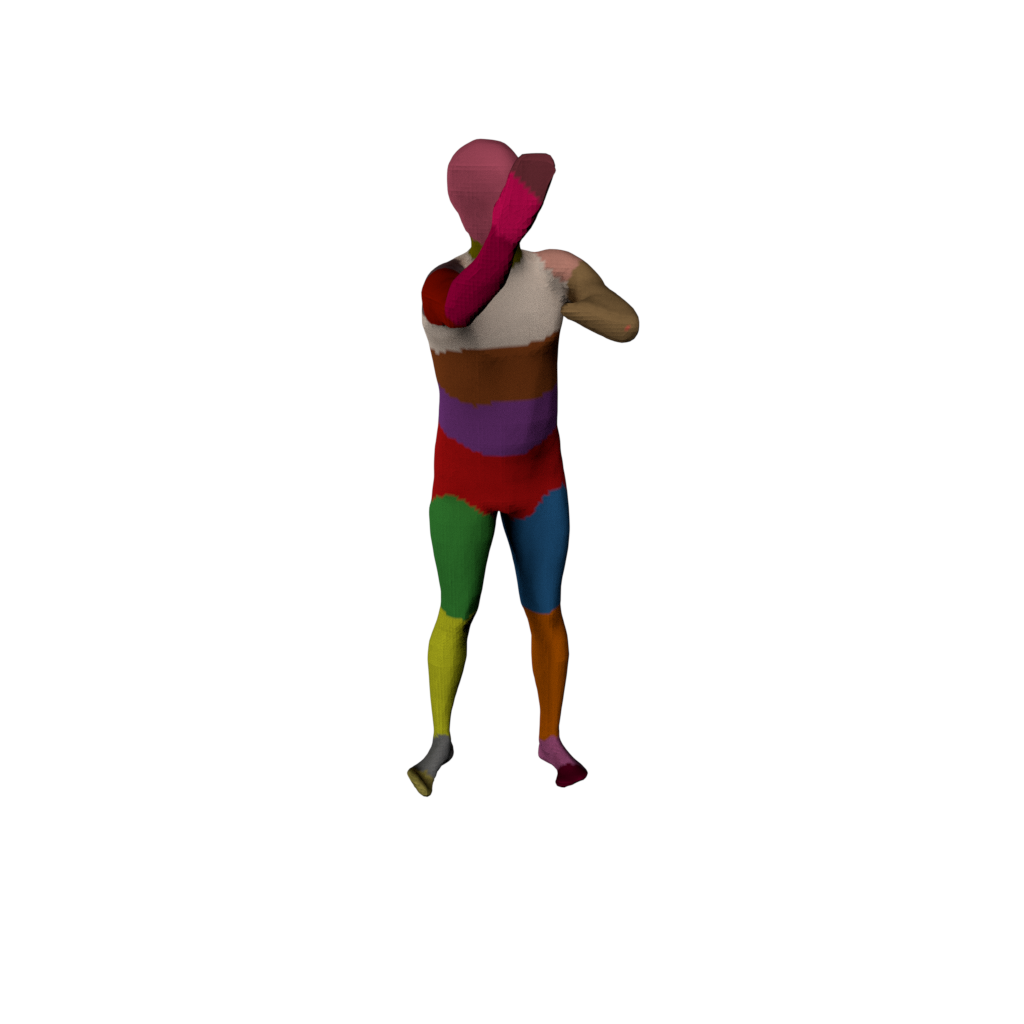}
    \caption{Ours surface}
 \end{subfigure}
 \begin{subfigure}[b]{0.15\textwidth}
    \includegraphics [trim=10cm 7cm 10cm 5cm, width=0.95\textwidth]{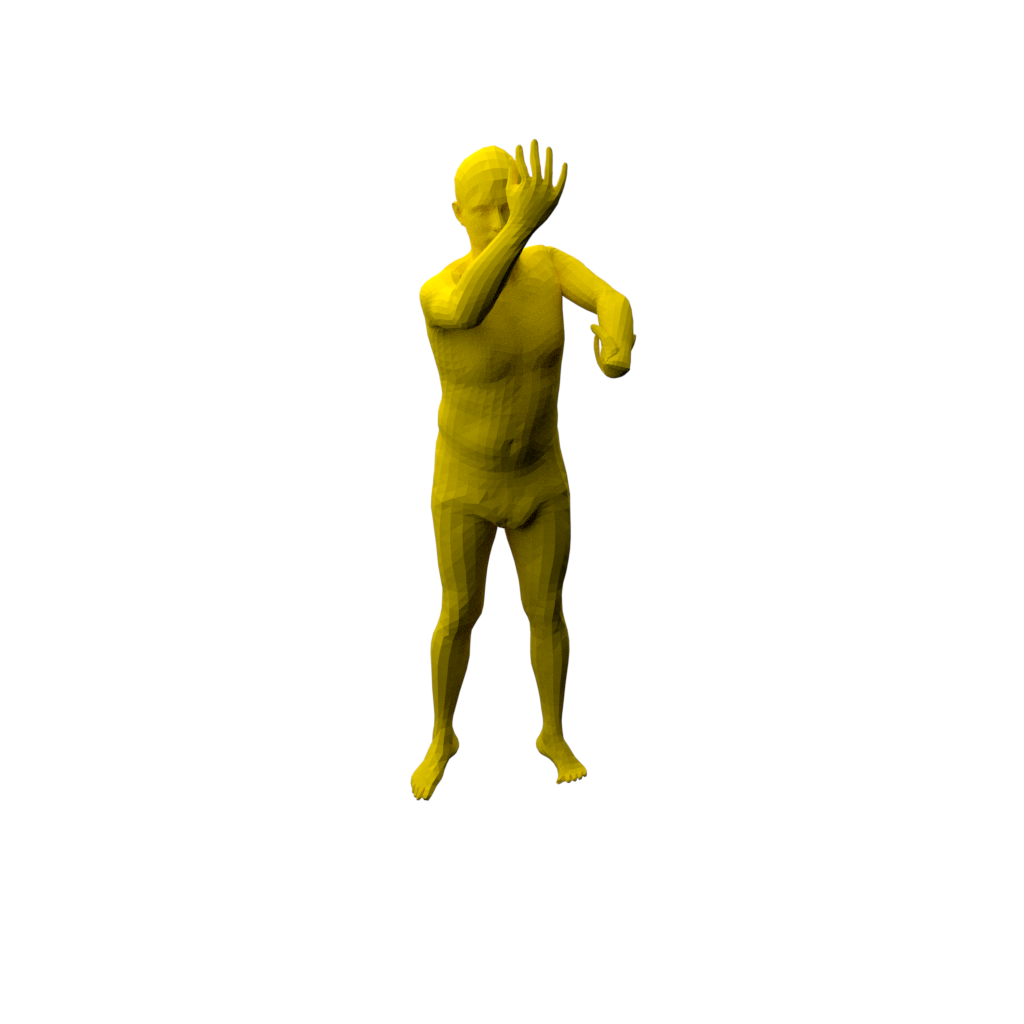}
    \caption{Ours SMPL}
 \end{subfigure}
 \begin{subfigure}[b]{0.15\textwidth}
    \includegraphics [trim=9cm 7cm 9cm 5cm, width=0.95\textwidth]{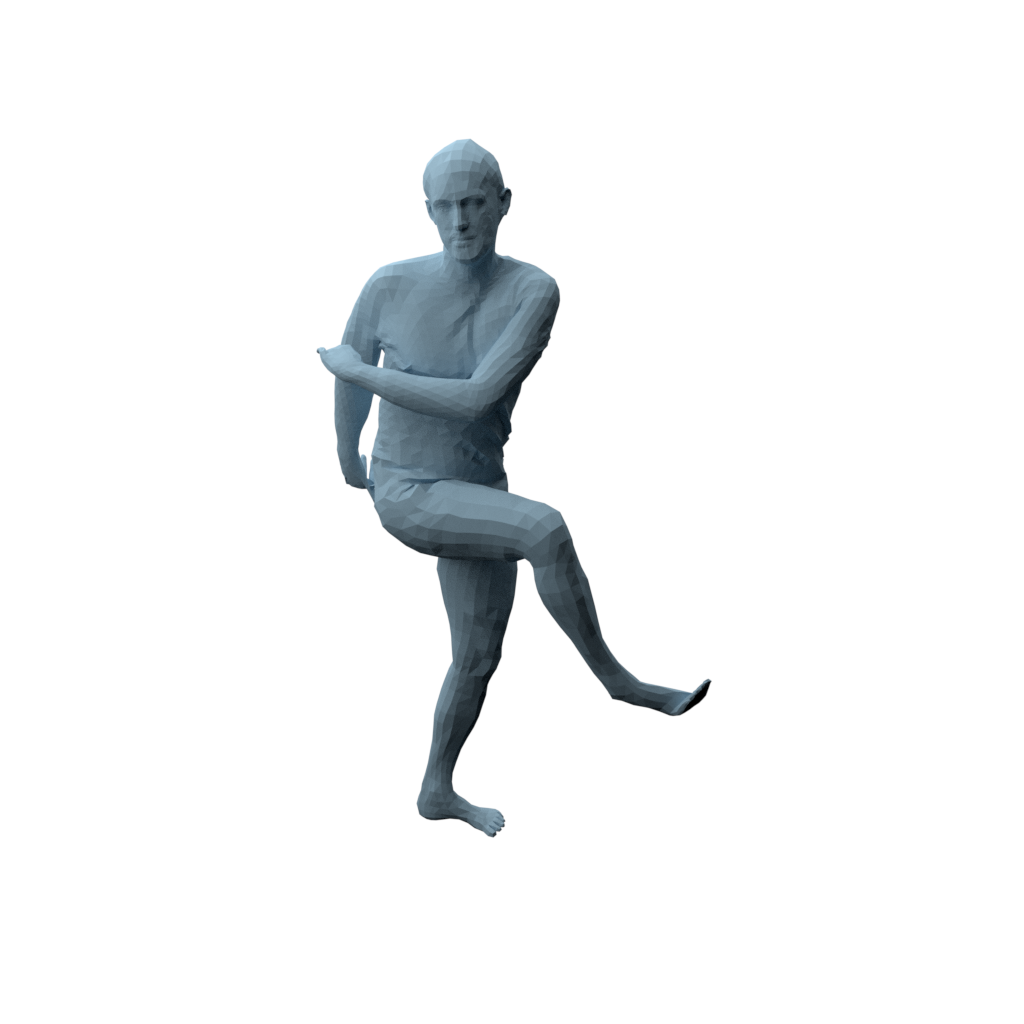}
    \caption{GT}
 \end{subfigure}
 \begin{subfigure}[b]{0.15\textwidth}
    \includegraphics [trim=9cm 7cm 9cm 5cm, width=0.95\textwidth]{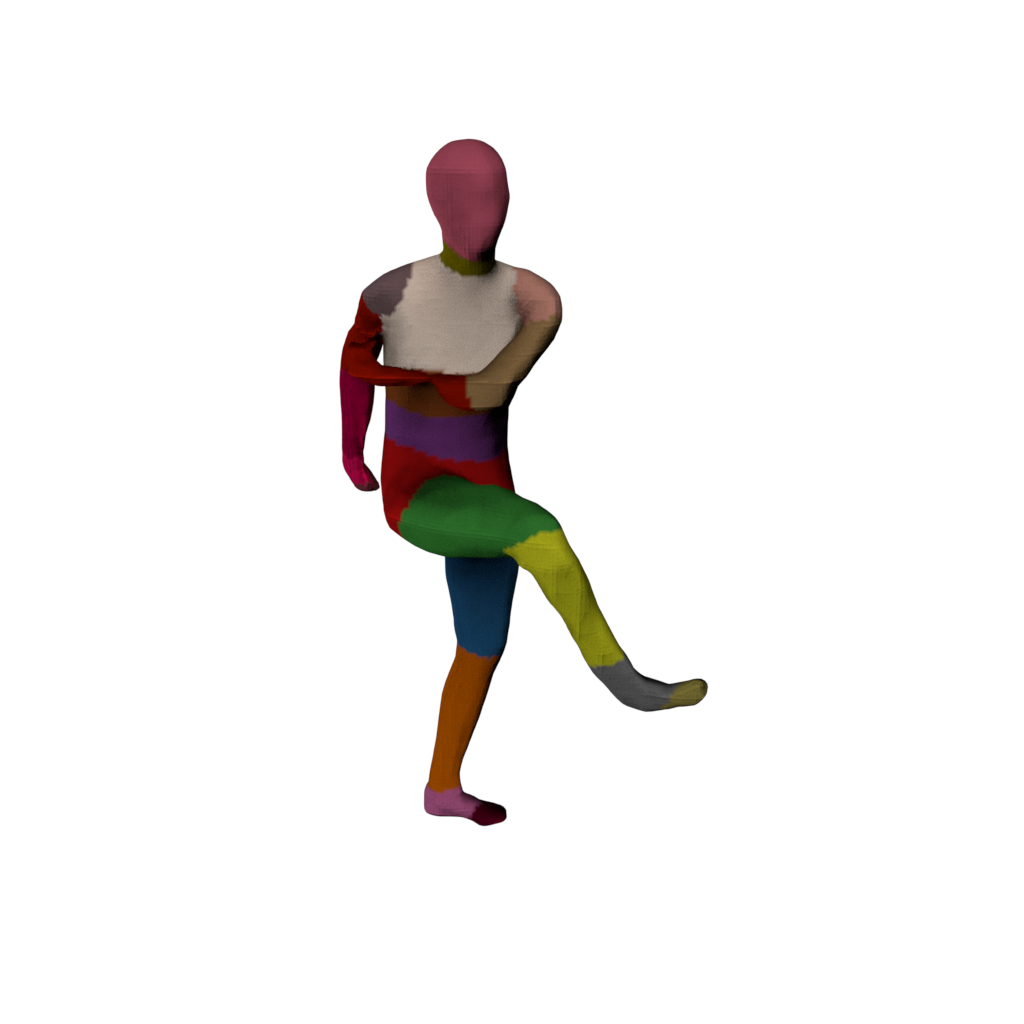}
    \caption{Ours surface}
 \end{subfigure}
 \begin{subfigure}[b]{0.15\textwidth}
    \includegraphics [trim=9cm 7cm 9cm 5cm, width=0.95\textwidth]{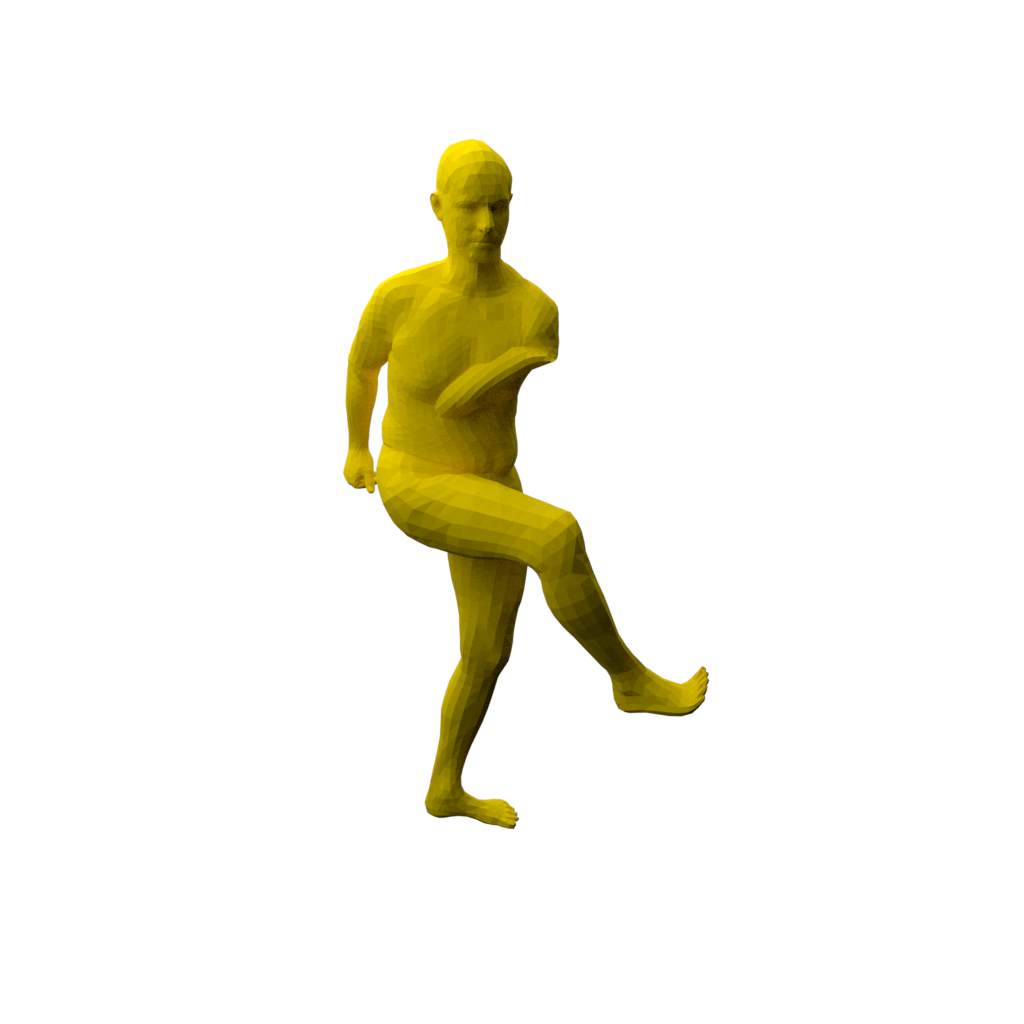}
    \caption{Ours SMPL}
 \end{subfigure}
\caption{Failure cases: we note that implicit surface reconstruction often fails in self-contact or near-self-contact scenarios. This leads to failures in registration.}
\label{fig:failure_cases}
\end{figure*}

Our approach often fails in self-contact or near-self-contact scenarios. We show typical failure cases in Fig~\ref{fig:failure_cases}. Another limitation of our approach is that it requires fully-supervised training on accurate surface registration. This kind of data is very hard to acquire in practice, thus limiting the scalability of our approach. A straightforward improvement would be integrating the self-supervised loop of~\cite{bhatnagar2020loopreg} into our pipeline, or utilizing the weakly supervised approach of~\cite{SCANimate:CVPR:21} to generate training data using registered under-cloth SMPL body.